\documentclass[10pt,letterpaper]{article}
\usepackage[utf8]{inputenc}
\usepackage{authblk}
\usepackage{setspace}
\usepackage[margin=1in]{geometry}
\usepackage{graphicx}
\graphicspath{{./Figures/}}
\usepackage{subcaption}
\usepackage{amsmath}
\usepackage{titlesec}

\usepackage[dvipsnames]{xcolor}
\usepackage{amssymb} 
\usepackage[textfont=it,labelfont=bf]{caption} 
\usepackage{booktabs} 
\usepackage{pgfplotstable} 
\pgfplotsset{compat=1.18} 
\usepackage{float}
\usepackage{tabularx}
\usepackage{tabularray} 
\usepackage{algorithm}
\usepackage{algpseudocode}
\usepackage{multirow}
\usepackage{hyperref}
\usepackage{helvet}
\usepackage{times}
\hypersetup{
    colorlinks,
    linkcolor={blue},
    citecolor={blue},
    urlcolor={blue!80!black},
}

\titlespacing\section{0pt}{12pt plus 3pt minus 1pt}{6pt plus 2pt minus 2pt}
\titlespacing\subsection{0pt}{6pt plus 3pt minus 1pt}{6pt plus 2pt minus 2pt}
\titlespacing\subsubsection{0pt}{3pt plus 2pt minus 2pt}{3pt plus 2pt minus 2pt}
\titlespacing\paragraph{0pt}{3pt plus 2pt minus 2pt}{3pt plus 2pt minus 2pt}
\titleformat{\section}{\large\bfseries\sffamily}{\thesection}{1em}{}
\titleformat{\subsection}{\normalsize\bfseries\sffamily}{\thesubsection}{1em}{}
\titleformat{\subsubsection}{\small\sffamily\bfseries}{\thesubsubsection}{1em}{}
\titleformat{\paragraph}{\small\sffamily\bfseries}{\thesubsubsection}{1em}{}
\newcommand{\filledsquare}[1]{%
  \textcolor{#1}{\blacksquare}%
}
\newcommand{\cref}[2]{\hyperref[#2]{#1~\ref*{#2}}}
\newcommand{\colref}[3]{\hyperref[#2]{#1~\ref*{#2}{#3}}}
\newcommand{\figref}[1]{\cref{Figure}{#1}}

\newcommand{\secref}[1]{\cref{Section}{#1}}

\newcommand{\tabref}[1]{\cref{Table}{#1}}

\newcommand{\figrefs}[2]{\hyperref[#1]{Figure~(\ref*{#1}--\ref*{#2})}}

\usepackage[numbers,sort&compress]{natbib}

\title{\usefont{OT1}{phv}{b}{n}\selectfont\Large{Evaluating Neural Radiance Fields (NeRFs) for 3D Plant Geometry Reconstruction in Field Conditions}}

\author[1]{\small Muhammad Arbab Arshad}
\author[2]{\small Talukder Jubery}
\author[2]{\small James Afful}
\author[2]{\small Anushrut Jignasu}
\author[2]{\small Aditya Balu}
\author[2]{\small \\Baskar Ganapathysubramanian}
\author[1,2*]{\small Soumik Sarkar}
\author[2*]{\small Adarsh Krishnamurthy}

\affil[1]{\small Department of Computer Science, Iowa State University, Ames, USA.}
\affil[2]{\small Department of Mechanical Engineering, Iowa State University, Ames, USA.}
\affil[*]{\small Corresponding Authors: soumik|adarsh@iastate.edu}

\date{}

\begin{document}

\maketitle

\begin{abstract}

We evaluate different Neural Radiance Fields (NeRFs) techniques for the 3D reconstruction of plants in varied environments, from indoor settings to outdoor fields. Traditional methods usually fail to capture the complex geometric details of plants, which is crucial for phenotyping and breeding studies. We evaluate the reconstruction fidelity of NeRFs in three scenarios with increasing complexity and compare the results with the point cloud obtained using LiDAR as ground truth. In the most realistic field scenario, the NeRF models achieve a 74.6\% F1 score after 30 minutes of training on the GPU, highlighting the efficacy of NeRFs for 3D reconstruction in challenging environments. Additionally, we propose an early stopping technique for NeRF training that almost halves the training time while achieving only a reduction of 7.4\% in the average F1 score. This optimization process significantly enhances the speed and efficiency of 3D reconstruction using NeRFs. Our findings demonstrate the potential of NeRFs in detailed and realistic 3D plant reconstruction and suggest practical approaches for enhancing the speed and efficiency of NeRFs in the 3D reconstruction process.

\noindent\textbf{Keywords:} Neural Radiance Fields | 3D Reconstruction | Field Conditions
\end{abstract}


\section{Introduction}

In recent years, reconstructing 3D geometry has emerged as a critical area within plant sciences. As global challenges in food production become increasingly complex~\citep{pereira2017water}, gaining a detailed understanding of plant structures has become essential. This goes beyond mere visual representation; capturing the intricate details of plant geometry provides valuable insights into their growth, responses to environmental stressors, and physiological processes~\citep{kumar2012high, paturkar2021making}. Consequently, there are several efforts for the 3D reconstruction of plants~\citep{feng20233d,cuevas2020segmentation,sarkar2023cyber}.

\begin{figure*}[!htb]
    \centering
    \includegraphics[width=0.95\linewidth]{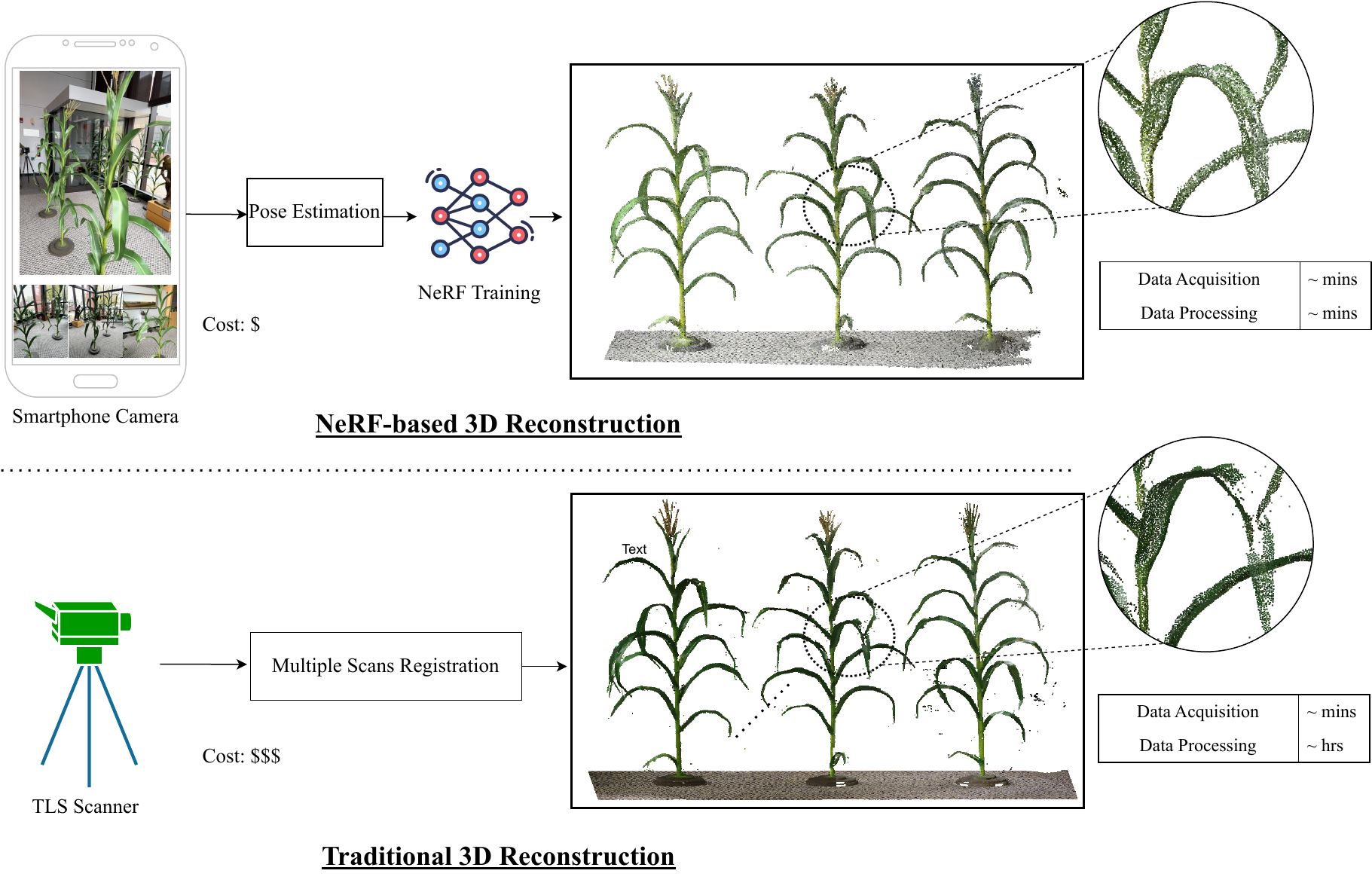}
    \caption{NeRFs are proposed as an alternative to traditional TLS scans for 3D plant reconstruction, offering cost-effective and efficient modeling from images captured at multiple angles using a smartphone camera, in contrast to the higher expense and extensive processing time required by TLS for multi-angle scan registration.}
    \label{fig:teaser}
\end{figure*}

One of the most common approaches for 3D reconstruction is photogrammetry, which relies on the analysis of discrete 2D pixels using techniques such as structure from motion (SfM)~\citep{eltner2020structure} and multi-view stereo (MVS)~\citep{chen2019point}. Another direct approach is utilizing LiDAR scanners (such as FARO 3D LiDAR scanner) to capture a dense 3D point cloud of the plants. This approach has been successfully used for the 3D reconstruction of Maize~\citep{wu2020mvs} and Tomato plants~\citep{wang20223dphenomvs}. Contemporary 3D modeling techniques for plant structures face significant challenges when attempting to capture the minute details inherent in plants~\citep{kumar2012high}. The complexity of plants, from their delicate leaf venation~\citep{lu2009venation} to intricate branching patterns~\citep{evers20113d}, necessitates models that encompass these specific details. Scans from multiple angles are essential to capture every detail, which is challenging since multiple LiDAR scans are time-consuming. Due to the limited poses, this approach does not scale well to capture minute details in large scenes; consequently, some desired details might be missed in the final model. \citet{andujar2018three} have emphasized that, even with advanced sensors, there are gaps in detailed reconstruction. They also point out that while devices such as the MultiSense S7 from Carnegie Robotics combine lasers, depth cameras, and stereo vision to offer reasonable results, the high acquisition costs can be prohibitive. At the same time, while photogrammetry is adept at large-scale reconstructions, it often cannot capture subtle details of plants~\citep{paturkar20193d, wu2020mvs, wang20223dphenomvs}.

In addition to the challenges mentioned above, the dynamic nature of flexible objects such as plants and their environment introduces an added complexity. Plants, unlike static entities, undergo growth, exhibit movement in reaction to environmental stimuli such as wind, and demonstrate both diurnal and seasonal variations. The environmental dynamism, coupled with plant behavior, further complicates modeling efforts. \citet{paturkar20193d} comprehensive investigation underscores that this dynamism inherently complicates the attainment of precise 3D models. Factors such as persistent growth, environmental dynamism, and external perturbations, notably in windy scenarios, jeopardize the consistency of data acquisition during imaging processes~\citep{lu2023bird, paturkar2021effect}. \citet{lienard2016embedded} highlight that errors in post-processing UAV-based 3D reconstructions can lead to severe, irreversible consequences. This complexity necessitates innovative solutions in 3D modeling and data processing.

One of the most recent approaches for 3D reconstruction is Neural Radiance Fields (NeRFs). At its core, NeRFs utilize deep learning to synthesize continuous 3D scenes by modeling the complete volumetric radiance field~\citep{mildenhall2021nerf}. NeRFs enable the rendering of photorealistic scenes from any viewpoint from a neural network trained using a set of 2D images without necessitating explicit 3D geometry or depth maps. NeRFs use implicit representations of the volumetric scene, in contrast to explicit representations such as point clouds in SfM and voxel grids in MVS. The implicit representation utilized by NeRF is resolution invariant, allowing for more detailed and granular modeling without the constraints of resolution-dependent methods. The versatility and rapid adoption of NeRF as a state-of-the-art technique in computer vision and graphics underscore its significance, with applications ranging from virtual reality~\citep{deng2022fov} to architectural reconstructions~\citep{tancik2022block}. Particularly in plant science research, NeRF's ability to capture fine details offers the potential for deep insights into plant structures and has the potential to be a vital tool in plant phenotyping and breeding (see \figref{fig:teaser}).

These factors indicate that the challenges in capturing detailed plant structures remain, even when employing sophisticated sensors. Financial implications further exacerbate these challenges. Traditional 3D modeling techniques often fall short of accurately capturing the complex 3D structures of plants~\citep{nguyen2015structured}. Although direct techniques such as LiDAR scanners provide better accuracy, their exorbitant costs often render them inaccessible to many researchers. \citet{tang2022benefits} delineate that the financial commitment associated with such advanced equipment, combined with the specialized expertise requisite for its operation, limits their adoption within academic and enthusiast domains.

In this paper, we perform a detailed evaluation of NeRF methodologies to assess their applicability and effectiveness for high-resolution 3D reconstruction of plant structures. An essential part of our study involves a comparative analysis of different NeRF implementations to determine the most effective framework for specific plant modeling needs. This includes assessing the methods' fidelity, computational efficiency, and ability to adapt to changes in environmental conditions. Such comparative analysis is crucial for establishing benchmarks for NeRF's current capabilities and identifying future technological improvement opportunities. Building on this foundation, we introduce an early-stopping algorithm to preemptively terminate the training process, significantly reducing computational cost while retaining model precision. We summarize our contributions as follows:
\begin{enumerate}
    \item A dataset collection encompassing a wide range of plant scenarios for reconstruction purposes consisting of images, camera poses, and ground truth TLS scans.  
    \item An evaluation of state-of-the-art NeRF techniques across different 2D and 3D metrics, offering insights for further research.
    \item An early stopping algorithm to efficiently halt the NeRF training when improvements in model fidelity no longer justify computational costs, ensuring optimal resource use.
    \item The development of an end-to-end 3D reconstruction framework using NeRFs designed specifically for the 3D reconstruction of plants.
\end{enumerate}

Our research aims to explore the feasibility of NeRFs for the 3D reconstruction of plants offering an in-depth analysis. A pivotal aspect of our methodology is using low-cost mobile cameras for data acquisition. By utilizing the widespread availability and imaging capabilities of modern smartphones, we can make high-quality image data collection more accessible and cost-effective. This approach, combined with the NeRFs' ability to process various image datasets for 3D reconstruction, can revolutionize plant reconstruction efforts.

The rest of the paper is arranged as follows. In \secref{Sec:Methods}, we outline the dataset collection, NeRF implementations, evaluation methods, and the LPIPS-based early-stopping algorithm. In \secref{Sec:Results}, we analyze results from single and multiple plant scenarios, both indoors and outdoors, using critical performance metrics. In \secref{Sec:Discussion}, we provide a theoretical discussion on the sampling strategies of different NeRF implementations and examine their impact on performance. We finally conclude in \secref{Sec:Conclusion}.

\section{Materials and Methods}
\label{Sec:Methods}
\begin{figure*}[!htb]
    \centering
    \includegraphics[width=0.99\linewidth]{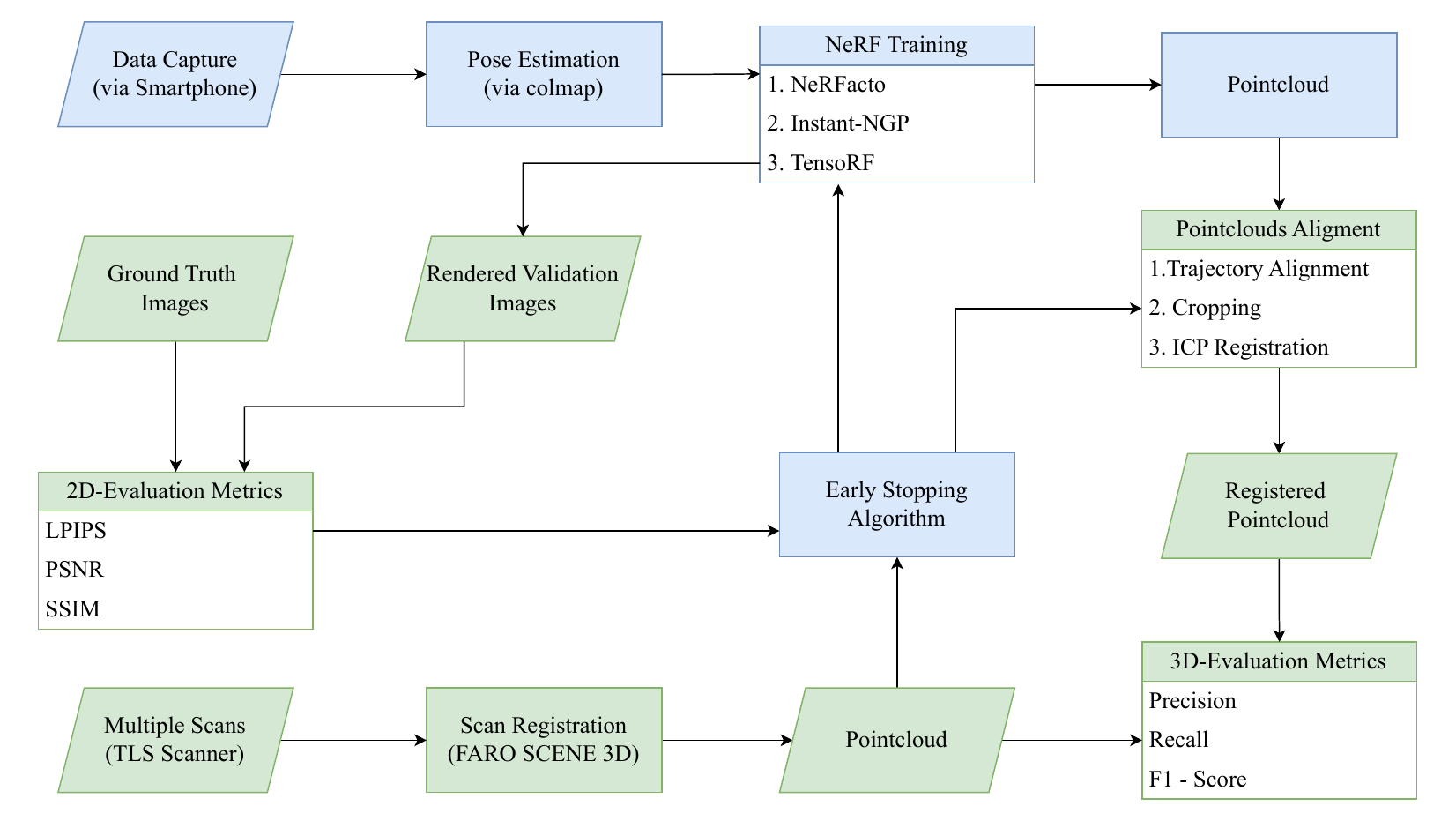}\\
    \vspace{1ex}
    \raisebox{0.5ex}{\textcolor{blue}{\rule{1cm}{1pt}}} Reconstruction \quad
    \raisebox{0.5ex}{\textcolor{green}{\rule{1cm}{1pt}}} Evaluation
    \caption{Workflow for 3D Reconstruction and Evaluation. The different steps of the above workflow is explained in detailed below.}
    \label{fig:workflow}
\end{figure*}

To evaluate 3D plant reconstruction using NeRFs, we propose a comprehensive methodology encompassing data collection, NeRF implementations, evaluation metrics, and an early stopping algorithm. The overall workflow of the different steps of our framework is shown in \figref{fig:workflow}. 

\subsection{Evaluation Scenarios and Data Collection}

\begin{figure*}[!t]
    \centering
    \begin{subfigure}{0.99\linewidth}
        \centering
        \includegraphics[width=0.22\linewidth]{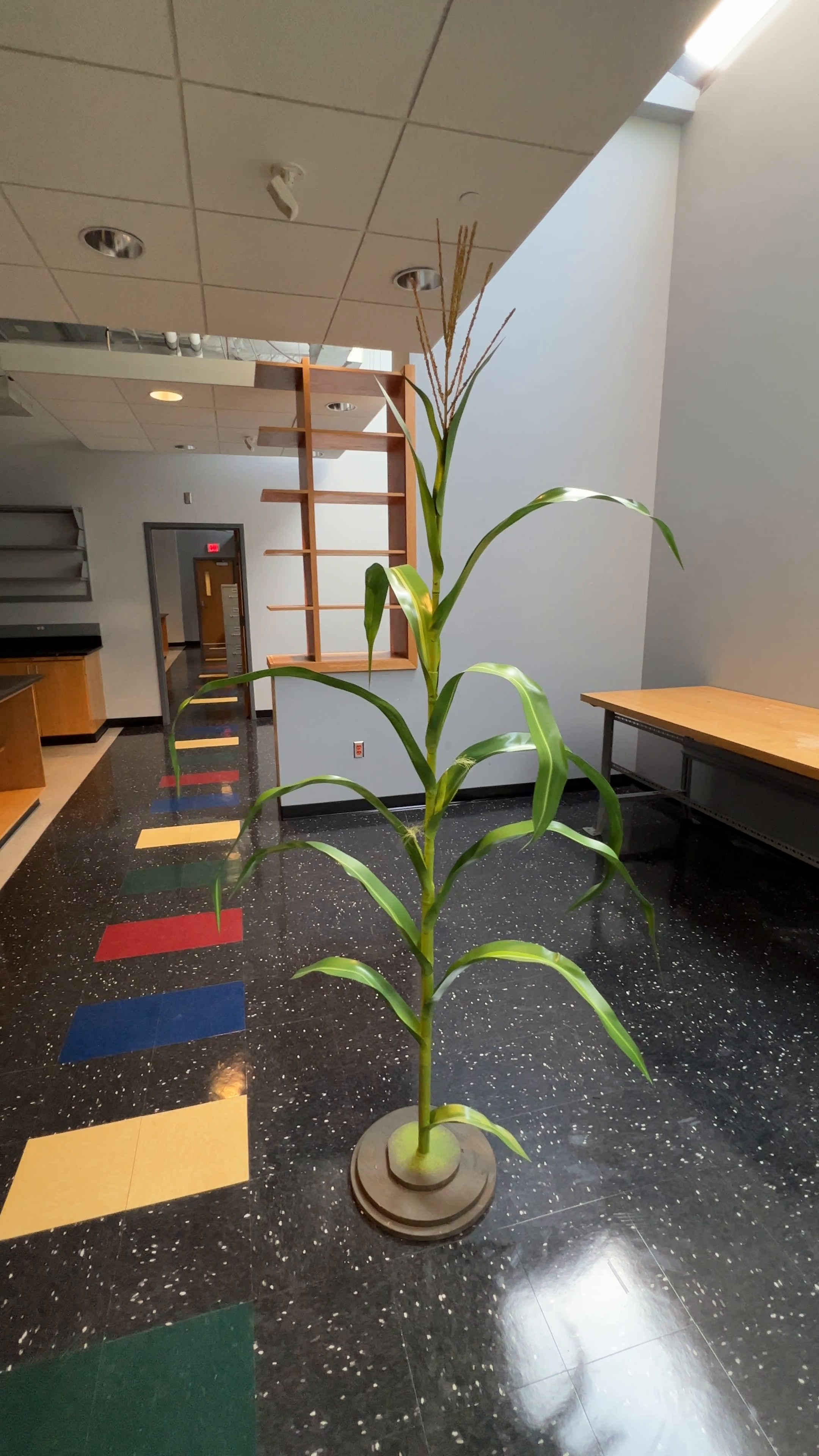}
        \includegraphics[width=0.22\linewidth]{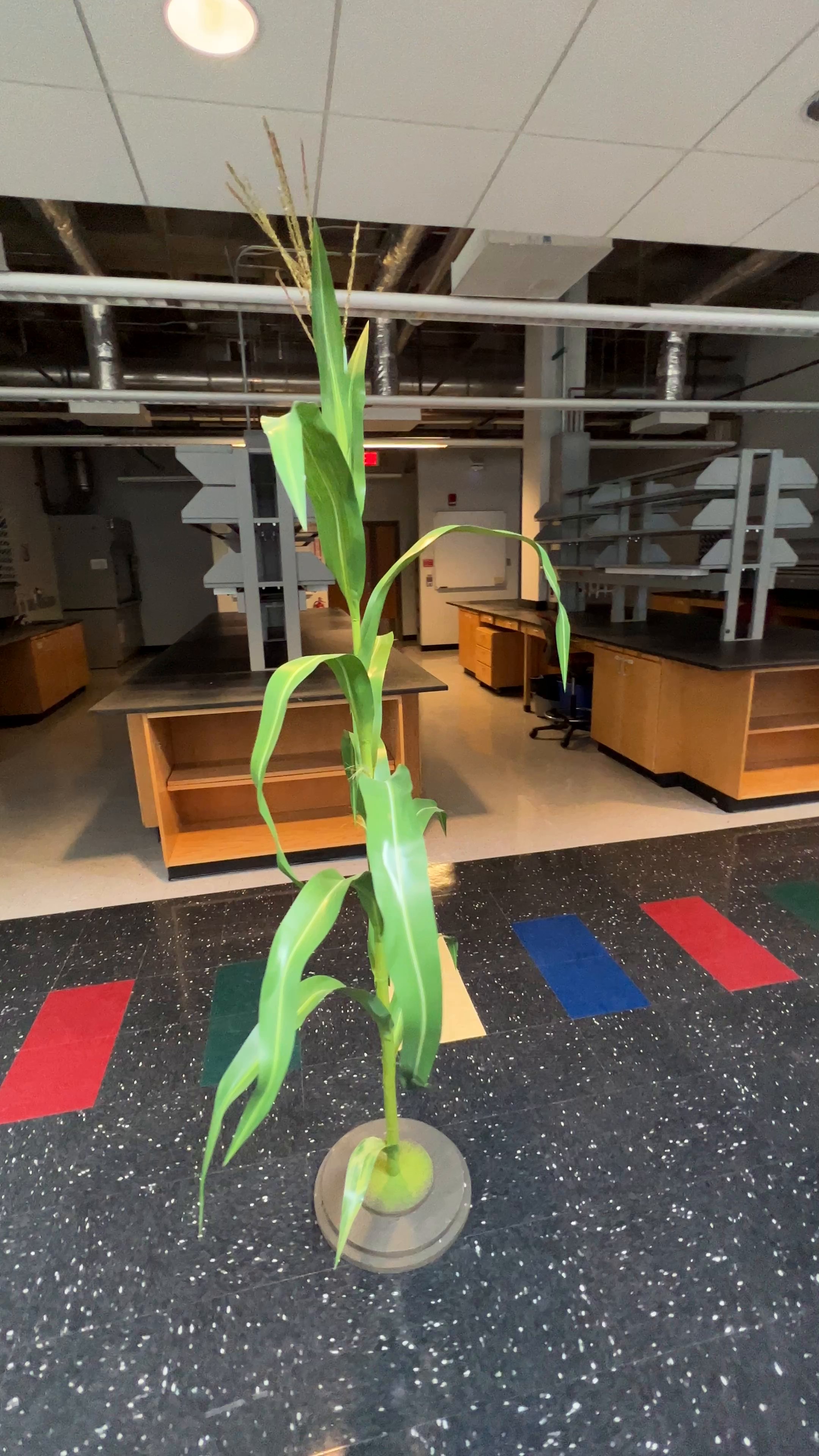}
        \includegraphics[width=0.22\linewidth]{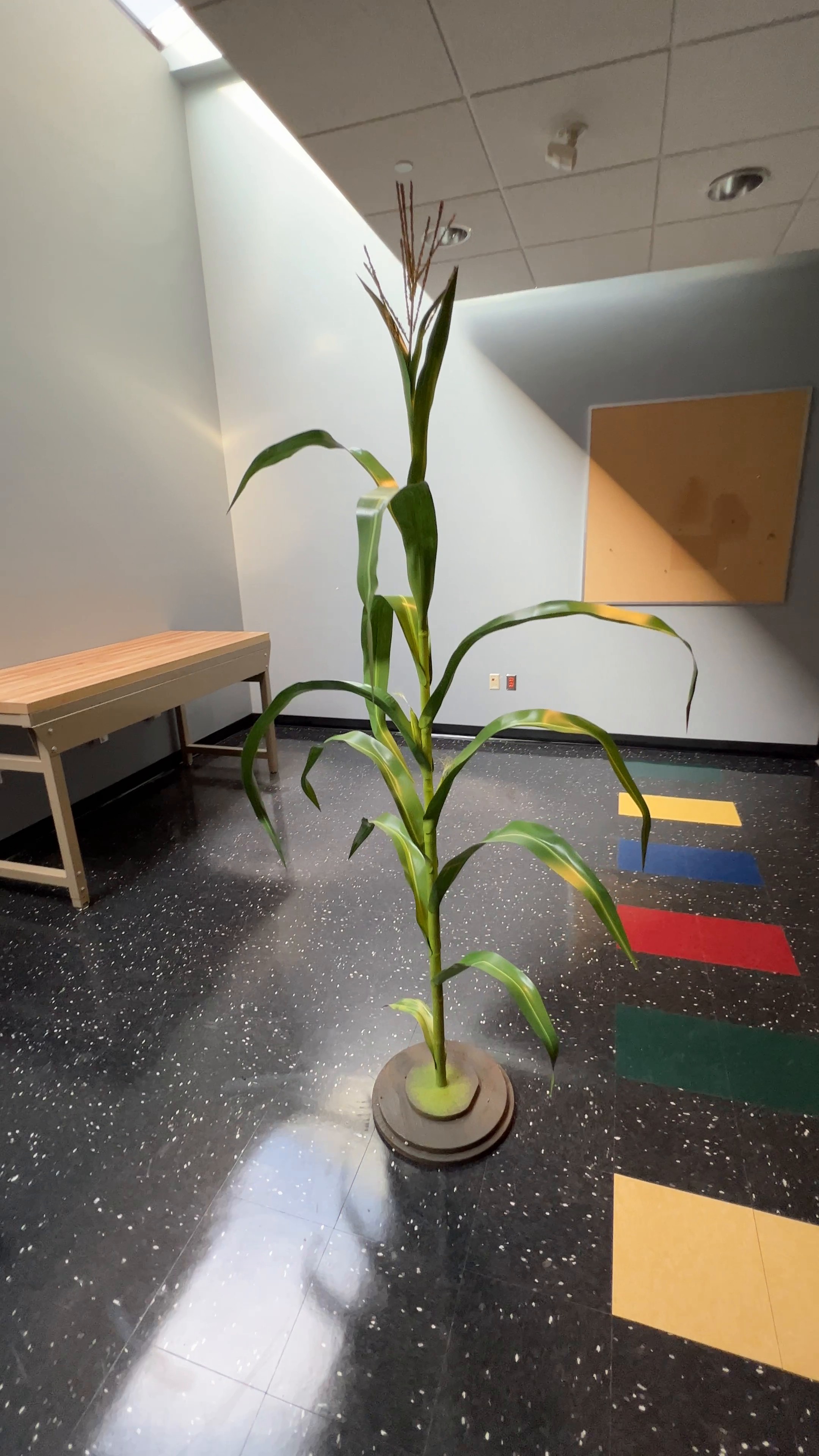}
        \includegraphics[width=0.22\linewidth]{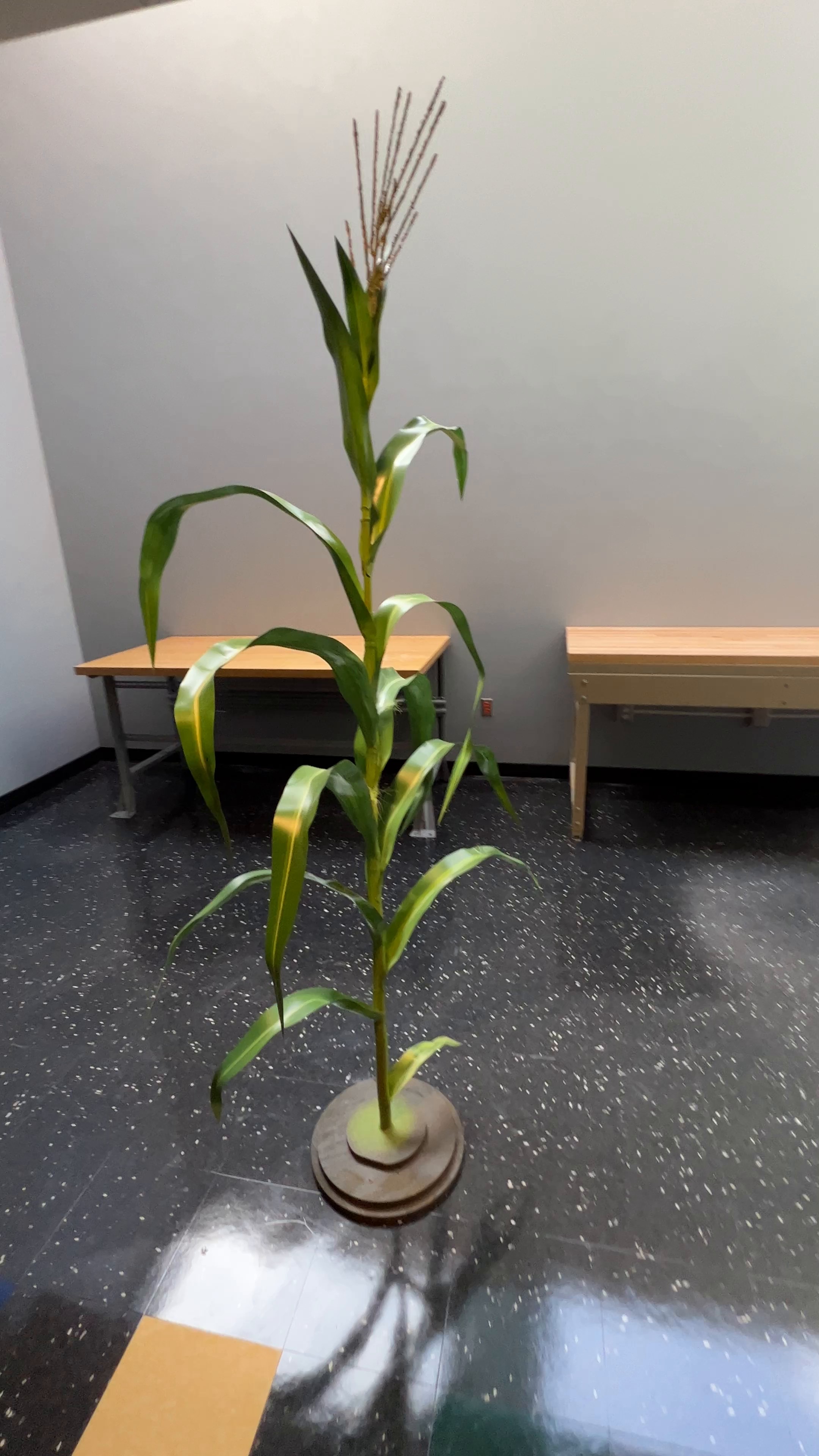}
        \caption{Scenario I}
        \vspace{0.15in}
        \label{fig:data_scenario_1}
    \end{subfigure}
    \begin{subfigure}{0.99\linewidth}
        \centering
        \includegraphics[width=0.22\linewidth]{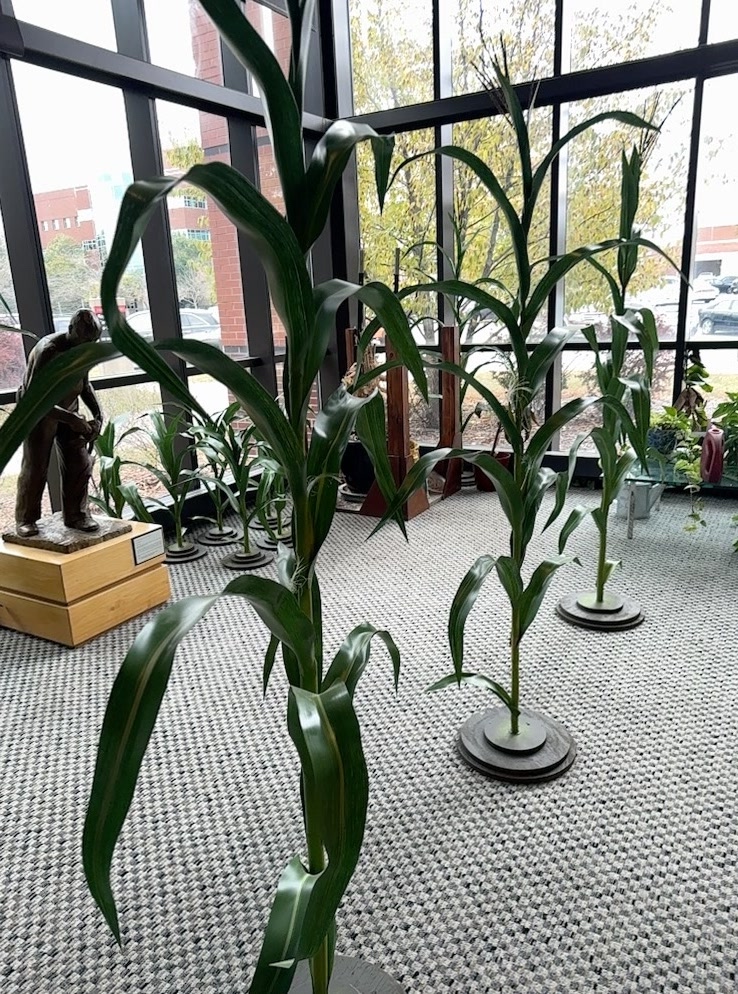}
        \includegraphics[width=0.22\linewidth]{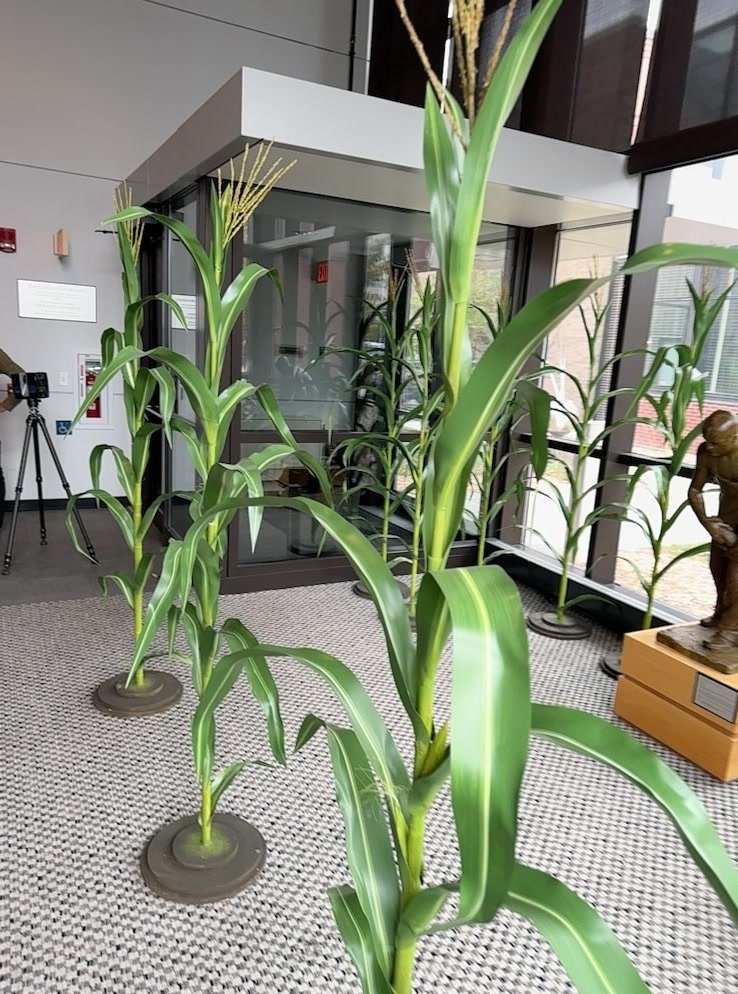}
        \includegraphics[width=0.22\linewidth]{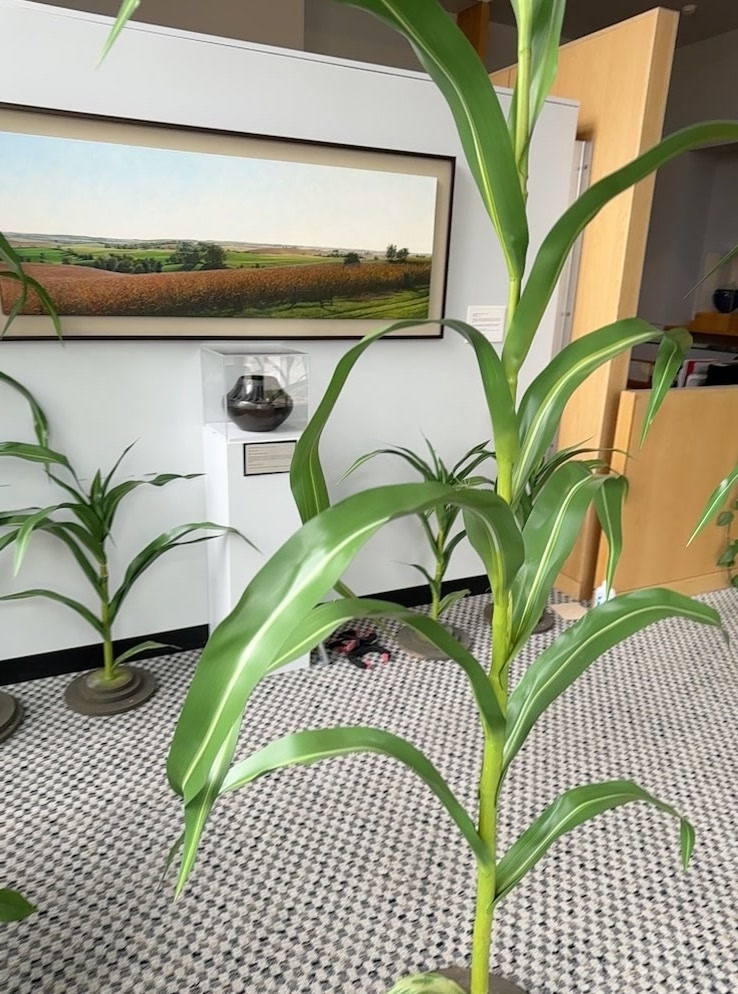}
        \includegraphics[width=0.22\linewidth]{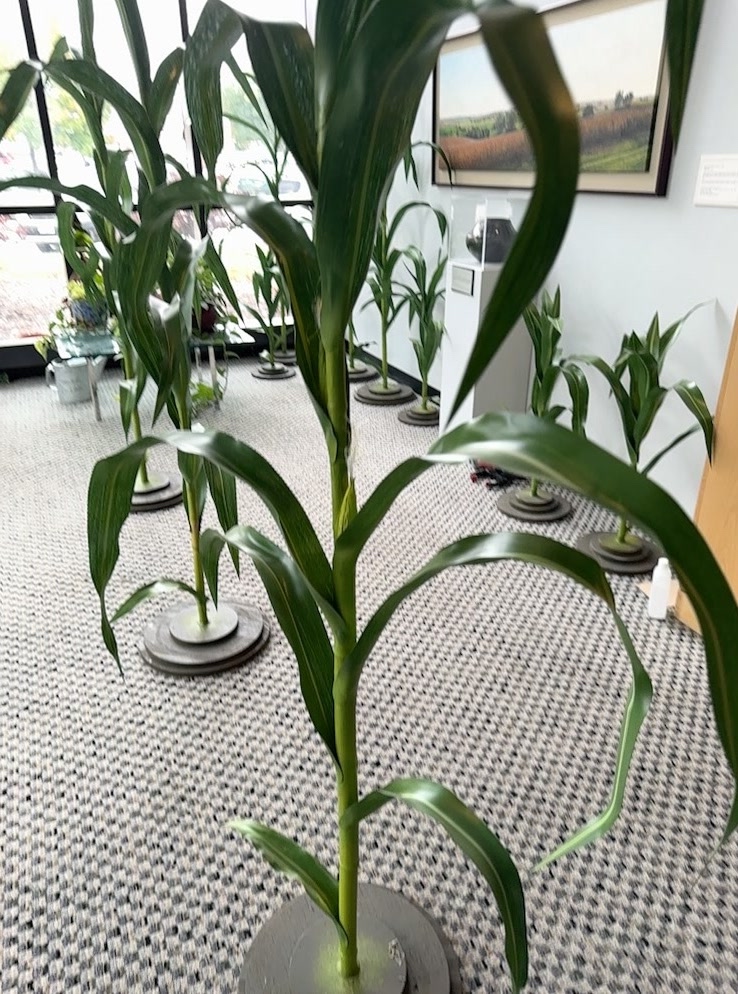}
        \caption{Scenario II}
        \vspace{0.15in}
        \label{fig:data_scenario_2}
    \end{subfigure}
    \begin{subfigure}{0.99\linewidth}
        \centering
        \includegraphics[width=0.22\linewidth]{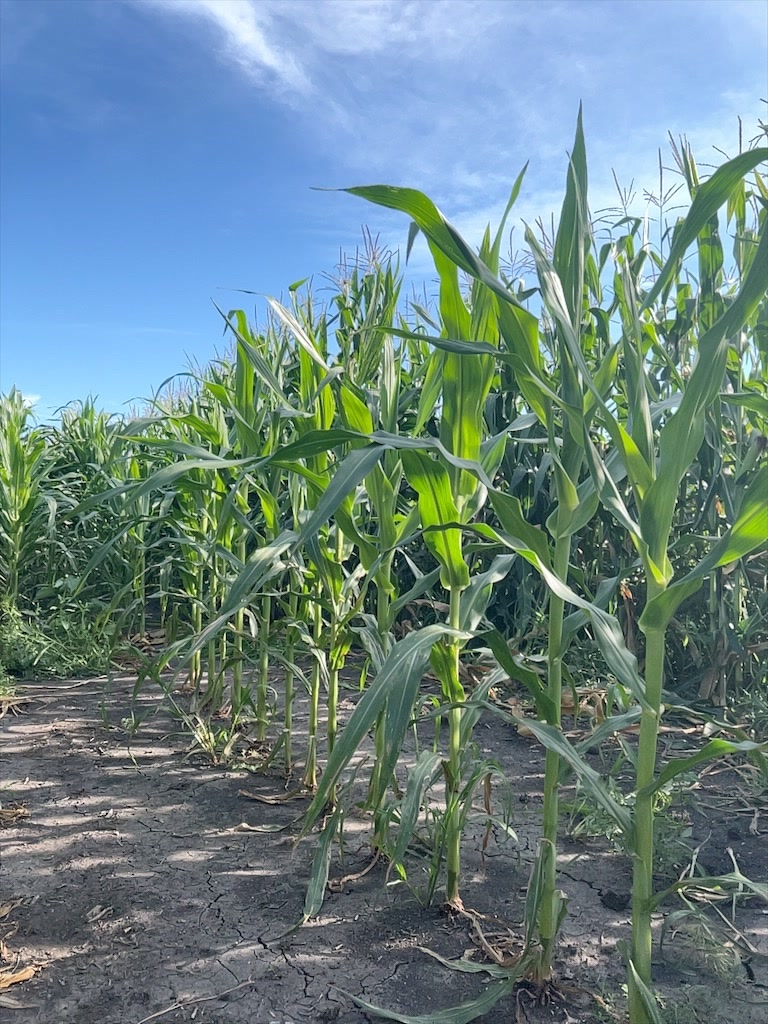}
        \includegraphics[width=0.22\linewidth]{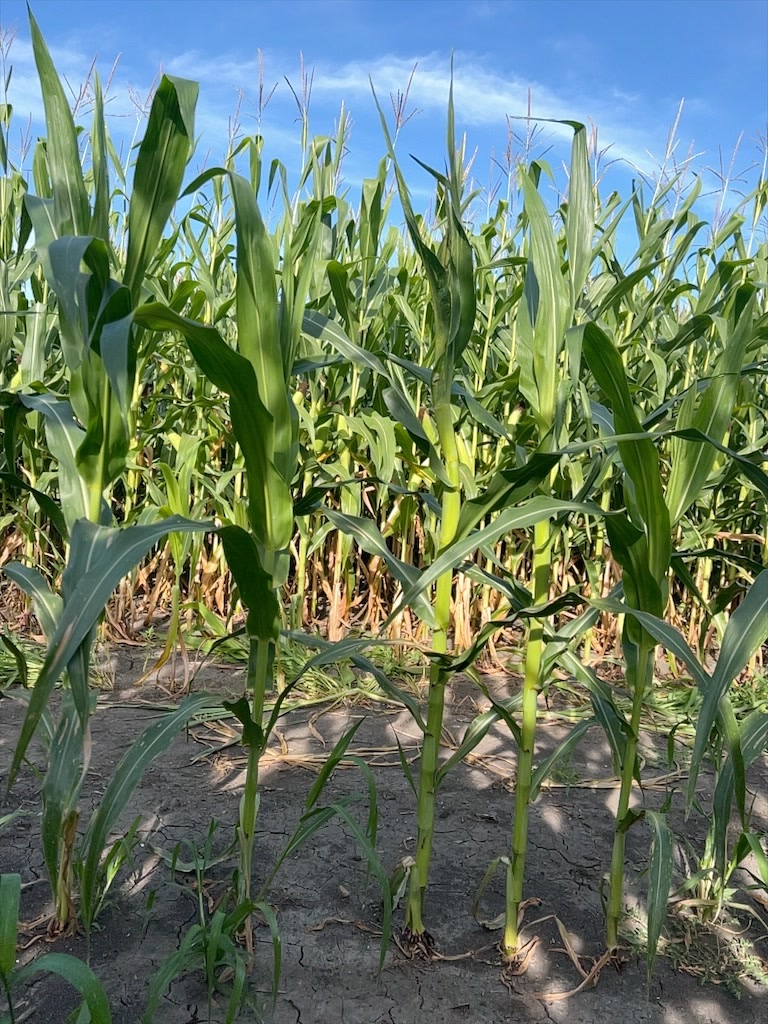}
        \includegraphics[width=0.22\linewidth]{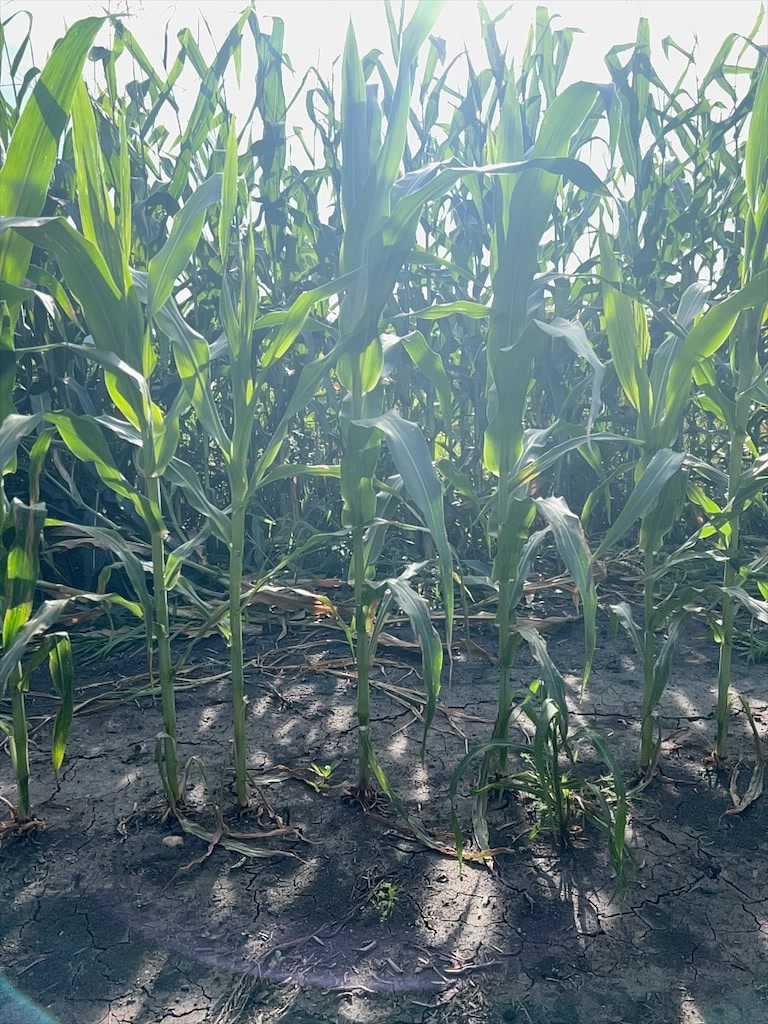}
        \includegraphics[width=0.22\linewidth]{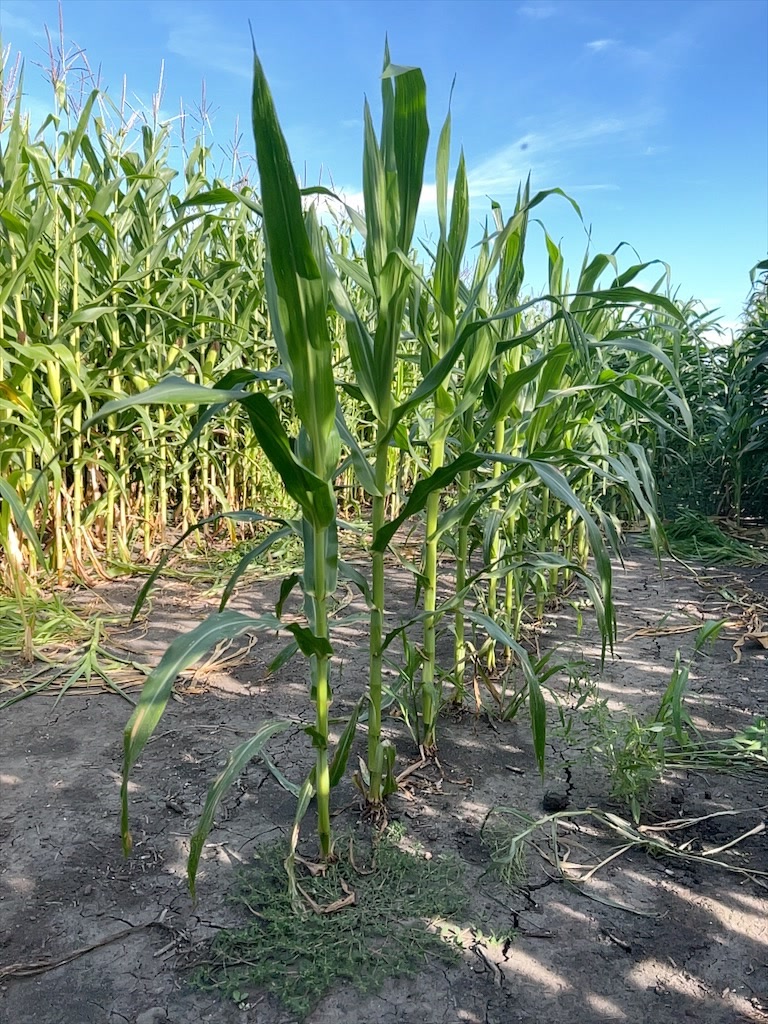}
        \caption{Scenario III}
        \label{fig:data_scenario_3}
    \end{subfigure}
    \caption{Example images input to NeRFs for reconstruction across three different scenarios. (a) Scenario I: Indoor single object, (b) Scenario II: Indoor multiple objects, (c) Scenario III: Outdoor scene.}
    \label{fig:data_scenarios}
\end{figure*}

We evaluate NeRFs, examining three distinct scenarios with ground truth data, from controlled indoor to dynamic outdoor environments, and a final testing scenario. The four scenarios are:
\begin{enumerate}
    \item \textbf{Single Corn Plant Indoor:} This serves as the simplest test case. A solitary corn plant is placed in a controlled indoor environment. The lighting, background, and other environmental factors are kept constant. The objective is to assess the basic capabilities of NeRF in reconstructing an individual plant structure~\citep{jignasu2023plant} (see \figref{fig:data_scenario_1}).    
    \item \textbf{Multiple Corn Plants Indoor:} In this case, more than one corn plant is situated in an indoor setting. The increased complexity due to multiple plants poses a greater challenge for the 3D reconstruction. Inter-plant occlusions and varying plant orientations add an additional layer of complexity (see \figref{fig:data_scenario_2}).
    \item \textbf{Multiple Corn Plants in a Field with Other Plants:} This scenario represents a real-world agricultural field, where corn plants are interspersed with other types of plants. The added complexity due to variable lighting, wind, and other dynamic environmental conditions tests the robustness of the NeRF technology (see \figref{fig:data_scenario_3}). We selected a row plot of corn plants planted at approximately 0.2 m distance, approximately at the V12 stage. The leaves between two neighboring plants are overlapping.
    \item \textbf{In-field Test Data:} For validating the proposed early stopping methodology, a diverse dataset was assembled, featuring scenarios with Soybean, Anthurium Hookeri, a mixture of plants, Cymbidium Floribundum, and Hydrangea Paniculata~(see \figref{fig:lpips_validation}).
\end{enumerate}

Our training dataset for NeRF is sourced from RGB images and LiDAR data captured using a mobile phone, with the RGB images aiding in the 3D reconstruction of the plants and the LiDAR exclusively for pose capture. For all three scenarios, data is captured using an iPhone 13 Pro featuring 4K resolution. The device is held at a constant height while circling the plant to ensure consistent capture angles. The data collection process utilizes the Polycam app~\citep{PolyCam2023}, with approximately 2.5 minutes for scenario 3 (multiple plants in the outdoor setting) and around 1 minute for scenario 1 (single plant in the indoor setting). To establish accurate ground truth, we utilized high-definition terrestrial LiDAR scans using the Faro\textsuperscript{\textregistered} Focus S350 Scanner. The scanner has an angular resolution of 0.011 degrees, equating to a 1.5 mm point spacing over a 10 m scanning range, and the capacity to acquire point clouds of up to 700 million points (MP) at 1 million points per second. Additionally, the scanner includes a built-in RGB camera that captures 360-degree images once the scanning process is complete.

Both in indoor and outdoor settings, we scan the plants from four (for the single plant) to six (for multiple plants) locations around the plant(s) at a height of 1.5 m and a distance of 1.5 m from the plant(s). To reduce the movement of the leaves during scanning, in indoor settings, we ensure that there is no airflow around the plants, and in outdoor settings, we waited for a suitable time when there was negligible wind flow (August 31, 2023, at 8:30 a.m.). Each scan required approximately 2.5 minutes, totaling a capture time of around 18 minutes in outdoor settings, including manually moving the scanner around the plot. The six scans were processed in SCENE\textsuperscript{\textregistered} software to add RGB color data to the point clouds, followed by the registration of the clouds by minimizing cloud-to-cloud distance and top view distance. Afterward, we cropped out the area of interest from the registered point cloud, removed duplicate points, and reduced noise using statistical outlier removal based on global and local point-to-point distance distributions. This process resulted in the point cloud having an average resolution of about 7 mm. This experimental setup enables the NeRF algorithm to work on a range of complexities, from controlled environments to dynamic, real-world conditions.

Camera pose estimation is a crucial second step, typically achieved through a Structure from Motion (SfM) pipeline such as COLMAP~\citep{colmap}. This process is essential for obtaining accurate 3D structures from sequences of images by determining correspondences between feature points and by using sequential matching, especially effective since our dataset comprises video frames.

\subsection{Neural Radiance Fields (NeRFs)}

Neural Radiance Fields (NeRFs) model a scene as a continuous function mapping a 3D position $\mathbf{x} = (x, y, z)$ and a 2D viewing direction $\mathbf{d} = (\theta, \phi)$ to a color $\mathbf{c} = (r, g, b)$ and density $\sigma$. The function is parameterized by a neural network $F_{\theta}$, expressed as:
\begin{equation}
    (\mathbf{c}, \sigma) = F_{\theta}(\mathbf{x}, \mathbf{d})
\end{equation}
Rendering an image involves integrating the color and density along camera rays, a process formalized as:
\begin{equation}
    \mathbf{C}(\mathbf{r}) = \int_{t_n}^{t_f} T(t) \sigma(\mathbf{r}(t)) \mathbf{c}(\mathbf{r}(t), \mathbf{d}) dt
\end{equation}
where $T(t) = \exp\left(-\int_{t_n}^{t} \sigma(\mathbf{r}(s)) ds\right)$ represents the accumulated transmittance along the ray $\mathbf{r}(t) = \mathbf{o} + t\mathbf{d}$, with $\mathbf{o}$ being the ray origin and $[t_n, t_f]$ the near and far bounds. In our workflow, we incorporate some of the state-of-the-art NeRF implementations optimized for their 3D reconstruction capabilities, which are critical to enable large-scale plant phenotyping studies. Specifically, we employ Instant-NGP~\citep{muller2022instant}, TensoRF~\citep{chen2022tensorf}, and NeRFacto~\citep{tancik2023nerfstudio}.

We specifically chose Instant-NGP, TensoRF, and NeRFacto to evaluate for plant reconstruction since these implementations are more efficient and achieve comparable results as a vanilla NeRF approximately 50 times faster. Each of these implementations introduces several new features over the vanilla NeRF implementations. Instant-NGP introduces a small neural network complemented by a multiresolution hash table, optimizing the number of operations required for training and rendering~\citep{muller2022instant}. TensoRF, on the other hand, conceptualizes the radiance field as a 4D tensor and applies tensor decomposition to achieve better rendering quality and faster reconstruction times compared to the traditional NeRF approach~\citep{chen2022tensorf}. NeRFacto combines various techniques such as the Multilayer Perceptron (MLP) adapted from Instant-NGP, and the Proposal Network Sampler from MipNeRF-360~\citep{barron2022mip}. Apart from these three methods, we also tried the vanilla Mip-NeRF~\citep{barron2021mip}. Unfortunately, Mip-NeRF fails to reconstruct more complicated 3D scenes (such as Scenario II) in our testing. Please refer to the Supplementary section where we provide a table for training (over time) of MipNeRF. We briefly describe the three tested NeRF approaches below. 

\noindent\textbf{Instant-NGP:}
Instant-NGP introduces advancements in NeRFs by focusing on three key improvements: enhanced sampling through occupancy grids, a streamlined neural network architecture, and a multi-resolution hash encoding technique. The hallmark of Instant-NGP is its multi-resolution hash encoding. This approach maps input coordinates to trainable feature vectors stored across multiple resolutions. For each input coordinate, the method hashes surrounding voxel vertices, retrieves and interpolates the corresponding feature vectors, and then inputs these interpolated vectors into the neural network. This process enhances the model's ability to learn complex geometries and ensures a smoother function due to the trainable nature of the feature vectors. The overall design of Instant-NGP drastically accelerates NeRF training and rendering, enabling near real-time processing capabilities. These enhancements collectively empower Instant-NGP to achieve speedups of up to 1000$\times$. The method also employs multiscale occupancy grids to efficiently bypass empty space and areas beyond dense media during sampling, thereby reducing the computational load. These occupancy grids are dynamically updated based on the evolving understanding of the scene's geometry, facilitating an increase in sampling efficiency. In parallel, Instant-NGP adopts a compact, fully-fused neural network architecture designed for rapid execution. This network is optimized to operate within a single CUDA kernel, consisting of only four layers with 64 neurons each, resulting in a speed boost—achieving a 5-10 times faster performance than traditional NeRF implementations.

\noindent\textbf{TensoRF:}
TensoRF improves scene representation by modeling the radiance field as a 4D tensor within a 3D voxel grid, where each voxel is enriched with multi-channel features. This model leverages tensor decomposition to efficiently manage the high-dimensional data, utilizing two key techniques: Canonic Polyadic (CP) and Vector-Matrix (VM) decompositions. CP decomposition simplifies the tensor into rank-one components using compact vectors, reducing the model's memory footprint. VM decomposition, alternatively, breaks the tensor into compact vector and matrix factors, striking a balance between memory efficiency and detail capture. These enable TensoRF to reduce memory requirements while enhancing rendering quality and accelerating reconstruction times. CP decomposition leads to faster scene reconstruction with improved rendering quality and a smaller model size compared to conventional NeRF approaches. VM decomposition takes this further, offering even better rendering quality and quicker reconstruction, all within a compact model size.

\noindent\textbf{NeRFacto:}
NeRFacto is an aggregate of techniques optimized for rendering static scenes from real images. The model enhances the NeRF framework by incorporating pose refinement and advanced sampling strategies to improve the fidelity of the scene reconstruction. Pose refinement is critical when initial camera poses are imprecise, which is often the case with mobile capture technologies. NeRFacto refines these poses, thus mitigating artifacts and enhancing detail. The model employs a Piecewise sampler for initial scene sampling, allocating samples to optimize the coverage of both near and distant objects. This is further refined using a Proposal sampler, which focuses on areas that contribute most to the scene's appearance and is informed by a density function derived from a small, fused MLP with hash encoding. Such a design ensures efficient sampling and better reconstruction. Further explanation and contrast with Instant-NGP is given in the discussion section. The implementations for aforementioned algorithms are taken from the open source project NeRFStudio~\citep{tancik2023nerfstudio}.

\begin{table*}[!t]
    \centering
    \caption{Recent works comparing the performance of different NeRF techniques for 3D reconstruction applications.
             $^{\dagger}$Used original implementation. $^{\ddagger}$Used implementation in NeRFStudio or SDFStudio.}
    \setlength\extrarowheight{3pt}
    \begin{tabular}{|p{0.185\linewidth}|c|c|c|c|p{0.20\linewidth}|}
    \hline
    \textbf{Paper} & \textbf{Instant-NGP} & \textbf{NeRFacto} & \textbf{TensoRF} & \textbf{NeRF} & \textbf{Additional} \textbf{Methods} \\
    \hline \hline 
    \citet{azzarelli2023towards} & \checkmark$^{\ddagger}$  & \checkmark$^{\ddagger}$ & $\times$ & $\times$ & Mip-NeRF \\
    \hline
    \citet{radl2023analyzing} & $\times$ & \checkmark$^{\ddagger}$ & $\times$ & \checkmark$^{\ddagger}$ & Mip-NeRF \\
    \hline
    \citet{li2023steernerf} & \checkmark$^{\ddagger}$ & $\times$ & $\times$ & \checkmark$^{\dagger}$ & NSVF, PlenOctree, KiloNeRF, DIVeR \\
    \hline
    \citet{remondino2023critical} & \checkmark$^{\dagger}$ & \checkmark$^{\ddagger}$ & \checkmark$^{\dagger}$ & $\times$ &  MonoSDF, VolSDF, NeuS, UniSurf \\ 
    \hline
    \citet{balloni2023few} & \checkmark$^{\dagger}$ & $\times$ & $\times$ & $\times$ & - \\
    \hline
    \textit{Ours} & \checkmark$^{\ddagger}$ & \checkmark$^{\ddagger}$ & \checkmark$^{\ddagger}$ & $\times$ & - \\
    \hline
    \end{tabular}
    \label{tab:literature}
\end{table*}

There have been several recent works that have compared NeRF approaches for 3D reconstruction. \tabref{tab:literature} summarizes some recent work evaluating different NeRF methodologies. Some of these recent research works also employ additional methods to improve reconstruction fidelity. For example SteerNeRF~\citep{li2023steernerf} utilizes neural sparse voxel fields (NSVF)~\citep{liu2020neural}, KiloNeRF~\citep{reiser2021kilonerf}, PlenOctree~\citep{yu2021plenoctrees}, and DIVeR~\citep{wu2022diver}, to obtain a smooth rendering from different viewpoints. NSVF introduces a fast, high-quality, viewpoint-free rendering method using a sparse voxel octree for efficient scene representation. KiloNeRF accelerates NeRF's rendering by three orders of magnitude using thousands of tiny MLPs, maintaining visual quality with efficient training. PlenOctree uses an Octree data structure to store the Plenoptic function. DIVeR improves upon NeRF by using deterministic estimates for volume rendering, allowing for realistic 3D rendering from few images. Similar to our work, \citet{azzarelli2023towards} propose a framework for evaluating NeRF methods using Instant-NGP, NeRFacto, and Mip-NeRF, focusing on neural rendering isolation and parametric evaluation. \citet{radl2023analyzing} analyze trained vanilla NeRFs, Instant-NGP, NeRFActo, and Mip-NeRF, showing accelerated computations by transforming activation features, reducing computations by 50\%. 

\citet{remondino2023critical} analyze image-based 3D reconstruction comparing different NeRFs (including Instant-NGP, NeRFacto, TensoRF, MonoSDF~\citep{yu2022monosdf}, VolSDF~\citep{yariv2021volume}, NeUS~\citep{wang2021neus}, UniSurf~\citep{oechsle2021unisurf}) with traditional photogrammetry, highlighting their applicability and performance differences for reconstructing heritage scenes and monuments. \citet{balloni2023few} does the same but with using only Instant-NGP. Each of these different NeRF implementations have some advancements over vanilla NeRF. MonoSDF demonstrates that incorporating monocular geometry cues improves the quality of neural implicit surface reconstruction. VolSDF improves the volume rendering of signed distance fields (SDF) using a new density representation. NeuS introduces a bias-free volume rendering method for neural surface reconstruction, outperforming existing techniques in handling complex structures and self-occlusions. UniSurf combines implicit surface models and radiance fields, enhancing 3D reconstruction and novel view synthesis without input masks.

\pagebreak

\subsection{3D Registration}

We reconstruct the scene and capture point clouds using a FARO scan for ground truth. 3D registration or alignment is crucial to perform a one-to-one comparison between the NeRF-based reconstruction and ground truth. Our alignment and evaluation methodology is adapted from~\citet{knapitsch2017tanks}. In their work, they evaluate different pipelines and use COLMAP as a 'arbitrary reference' frame. However, in our case, all the NeRFs use COLMAP in their pipeline, so the reference and reconstruction frames become the same. The steps used for registration are:

\begin{itemize}
\item \noindent\textbf{Preliminary Camera Trajectory Alignment}: The NeRF-reconstructed point cloud is manually aligned with the ground truth using point-based alignment. Four corresponding points are selected in both point clouds to compute an initial transformation matrix. This matrix aligns the camera poses, providing initial scale and orientation estimates. This initial coarse-grained alignment step paves the way for more detailed alignment procedures.
    
\item \noindent\textbf{Cropping}: Each ground-truth model has a manually-defined bounding volume, outlining the evaluation region for reconstruction.
    
\item \noindent\textbf{Iterative Closest Point (ICP) Registration}: Drawing inspiration from the iterative refinement process detailed by \citet{besl1992method} and further refined by \citet{zhang1994iterative}, we adopt a three-stage approach~\citep{knapitsch2017tanks} for our initial registration framework. The process begins with a specified voxel size and an associated threshold for the initial registration. In the next iteration, the transformation result from the previous step is used as a starting point, with the voxel size reduced by half to achieve finer detail in the registration. The third stage aims to refine the alignment further by returning to the original voxel size and adjusting the threshold to facilitate convergence at each stage. This multi-scale strategy is designed to capture both coarse and fine details, thereby improving the accuracy and precision of the model alignment. However, in our adaptation for plant structure reconstruction, we diverged from \citet{knapitsch2017tanks} by maintaining the iterative process within a single stage rather than expanding across multiple stages. We found that increasing the iteration count tenfold, rather than the number of stages, prevented the registration process from collapsing~\citep{billings2015iterative}.
\end{itemize}

\subsection{Evaluation Metrics}
To assess the similarity between the ground truth (obtained from TLS) and the reconstructed 3D pointcloud, the following metrics are employed:

\begin{enumerate}
    \item 
    \textbf{Precision/Accuracy.} Given a reconstructed point set $\mathcal{R}$ and a ground truth set $\mathcal{G}$, the precision metric $P(d)$ assesses the proximity of points in $\mathcal{R}$ to $\mathcal{G}$ within a distance threshold $d$. Mathematically, it is formulated as:
   \begin{equation}
    P(d) = \frac{100}{|\mathcal{R}|} \sum_{\mathbf{r} \in \mathcal{R}} \mathbb{I}\left( \min_{\mathbf{g} \in \mathcal{G}} \|\mathbf{r} - \mathbf{g}\| < d \right),
   \end{equation}
    where $\mathbb{I}(\cdot)$ is an indicator function. Precision values ranges from 0 to 100, with higher values indicating better performance. 

    \item 
    \textbf{Recall/Completeness.} Conversely, the recall metric $R(d)$ quantifies how well the reconstruction $\mathcal{R}$ encompasses the points in the ground truth $\mathcal{G}$ for a given distance threshold $d$. It is defined as:
    \begin{equation}
    R(d) = \frac{100}{|\mathcal{G}|} \sum_{\mathbf{g} \in \mathcal{G}} \mathbb{I}\left( \min_{\mathbf{r} \in \mathcal{R}} \|\mathbf{g} - \mathbf{r}\| < d \right).
    \end{equation}
    Its value ranges from 0 to 100, with higher values indicating better performance. Both the above two metrics are extensively utilized in recent studies.~\citep{balloni2023few, mazzacca2023nerf}.

    \item 
    \textbf{F-score.} The F-score, denoted as \( F(d) \), serves as a harmonic summary measure that encapsulates both the precision \( P(d) \) and recall \( R(d) \) for a given distance threshold \( d \). It is specifically designed to penalize extreme imbalances between \( P(d) \) and \( R(d) \). Mathematically, it can be expressed as:
   \begin{equation}
    F(d) = \frac{2 \times P(d) \times R(d)}{P(d) + R(d)}.
    \end{equation}
    The harmonic nature of the F-score ensures that if either \( P(d) \) or \( R(d) \) approaches zero, the F-score will also tend towards zero, providing a more robust summary statistic than the arithmetic mean. F-score ranges from 0 to 100, with higher values indicating better performance. The details about value of $d$ cutoff is given later in discussion about precision-recall curves. 
    
    For quantifying the quality of the NeRF-rendered 2D image compared to the validation image (left out from NeRF training), the following metrics are used: 

    \item 
    \textbf{Learned Perceptual Image Patch Similarity (LPIPS)~\citep{zhang2018unreasonable}:}
    To quantify the perceptual differences between two image patches, \( x \) and \( x_0 \), the Learned Perceptual Image Patch Similarity (LPIPS) framework employs activations from a neural network \( F \). Features are extracted from \( L \) layers and normalized across the channel dimension. For each layer \( l \), the normalized features are represented by \( \hat{y}^l \) and \( \hat{y}^{l0} \), which exist in the space \( \mathbb{R}^{H_l \times W_l \times C_l} \). These are then weighted channel-wise by a vector \( w_l \in \mathbb{R}^{C_l} \). The perceptual distance is computed using the \( \ell_2 \) norm, both spatially and across channels, as expressed in the equation:
    \begin{equation}
    d(x, x_0) = \sum_{l} \frac{1}{H_l W_l} \sum_{h,w} \left\| w_l \odot (\hat{y}^l_{hw} - \hat{y}^{l0}_{hw}) \right\|_2^2
    \end{equation}
    This distance metric, \( d(x, x_0) \), provides a scalar value indicating the perceptual dissimilarity between the patches. The vector \( w_l \) weights the contribution of each channel to the distance metric. By setting \( w_l \) to \( 1/\sqrt{C_l} \), the computation effectively measures the cosine distance, highlighting the directional alignment of the feature vectors instead of their magnitude. Its value ranges from 0 to 1, with lower values indicating better performance.

    \item 
    \textbf{Peak Signal-to-Noise Ratio (PSNR):}~\citep{hore2010image} The PSNR between two images, one being the reference and the other the reconstructed image, is defined as:
    \begin{equation}
    \text{PSNR} = 10 \cdot \log_{10}\left(\frac{\text{MAX}_I^2}{\text{MSE}}\right),
    \end{equation}
    where $\text{MAX}_I$ is the maximum possible pixel value of the image, and $\text{MSE}$ is the Mean Squared Error between the reference and the reconstructed image. The MSE is given by:
    \begin{equation}
    \text{MSE} = \frac{1}{mn} \sum_{i=1}^{m} \sum_{j=1}^{n} \left( I(i,j) - K(i,j) \right)^2,
    \end{equation}
    where $I$ is the reference image, $K$ is the reconstructed image, and $m$ and $n$ are the dimensions of the images. A higher value of PSNR indicate better performance. 

    \item 
    \textbf{Structural Similarity Index (SSIM):}~\citep{wang2004image} The SSIM index is a method for predicting the perceived quality of digital television and cinematic pictures, as well as other kinds of digital images and videos. SSIM is designed to improve on traditional methods like PSNR and MSE, which have proven to be inconsistent with human eye perception. The SSIM index between two images $x$ and $y$ is defined as:
    \begin{equation}
    \text{SSIM}(x, y) = \frac{(2\mu_x\mu_y + C_1)(2\sigma_{xy} + C_2)}{(\mu_x^2 + \mu_y^2 + C_1)(\sigma_x^2 + \sigma_y^2 + C_2)},
    \end{equation}
    where $\mu_x$ is the average of $x$, $\mu_y$ is the average of $y$, $\sigma_x^2$ is the variance of $x$, $\sigma_y^2$ is the variance of $y$, $\sigma_{xy}$ is the covariance of $x$ and $y$, and $C_1$ and $C_2$ are constants to stabilize the division with weak denominator. These last three metrics do not need the 3D ground truth and are widely used in literature~\citep{xu2022point,zhang2021ners} for evaluation. SSIM ranges from -1 to 1, with higher values indicating better performance.
\end{enumerate}

\paragraph{Precision-Recall curves:}
Precision-recall curves are utilized to methodically evaluate how distance threshold $d$ changes influence precision $P(d)$ and recall $R(d)$ metrics, demonstrating the trade-off between these measurements under varying threshold conditions. To set the value of $d$ for the final assessment, we opt for a conservative estimate before the plateauing of precision-recall curves. For indoor scenarios, assuming a hypothetical grid size of 128x128x128 for reference, we establish $d$ at 0.005. In this scenario, the voxel size is calculated as $1/128 \approx 0.0078125$, which makes the threshold of 0.005 smaller than the voxel size. This indicates a requirement for points to be closer than the dimensions of a single voxel to be identified as distinct, highlighting a prioritization of detail sensitivity within a hypothetically coarser grid. Such a setting is especially pertinent for capturing the complex geometries of indoor plants, where precision in detail is crucial. Due to the size and complexity of the scene, a threshold of 0.01 is selected for outdoor plant reconstructions.

\subsection{Early Stopping of NeRF Training using LPIPS}
In training NeRFs for plant scene reconstruction, the F1 score is essential for validating the accuracy of the reconstructed point cloud against the ground truth. The inherent challenge during the training phase of NeRFs is the absence of ground truth, paradoxically the output we aim to correspond. Moreover, the training process for NeRFs is notoriously compute-intensive. The cumulative costs become challenging when scaled to multiple scenes or across extensive agricultural fields.

\begin{figure}[!h]
    \centering
    \begin{minipage}[c]{0.03\linewidth}
        \rotatebox{90}{\fontfamily{phv}\selectfont\footnotesize{F1 Score}}
        \vspace{5ex}
    \end{minipage}%
    \begin{minipage}[c]{0.96\linewidth}
        \begin{subfigure}{0.3\linewidth}
            \centering
            \includegraphics[width=\linewidth]{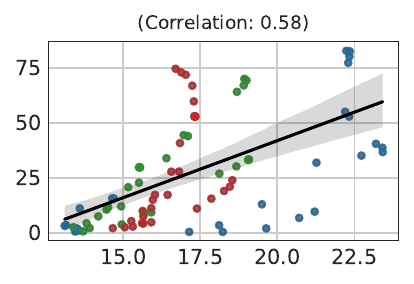}
            \caption{PSNR}
        \end{subfigure}
        \hfill
        \begin{subfigure}{0.3\linewidth}
            \centering
            \includegraphics[width=\linewidth]{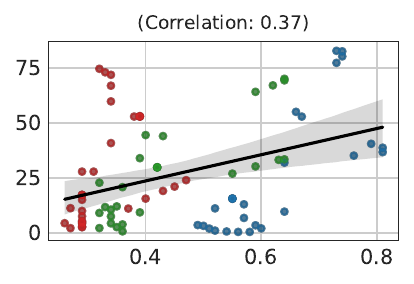}
            \caption{SSIM}
        \end{subfigure}
        \hfill
        \begin{subfigure}{0.3\linewidth}
            \centering
            \includegraphics[width=\linewidth]{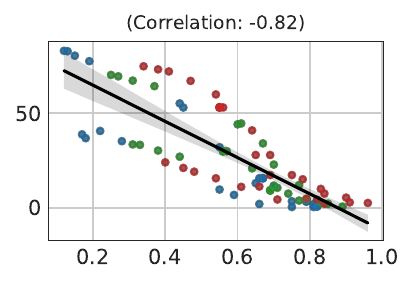}
            \caption{LPIPS}
        \end{subfigure}
    \vspace{2ex}
    \end{minipage}
    \raisebox{0.5ex}{\textcolor{blue}{\rule{1cm}{1pt}}} Scenario I \quad
    \raisebox{0.5ex}{\textcolor{green}{\rule{1cm}{1pt}}} Scenario II \quad
    \raisebox{0.5ex}{\textcolor{red}{\rule{1cm}{1pt}}} Scenario III
    \caption{Correlation analysis between different metrics with F1 Score via Pearson coefficients: (a)~PSNR, (b)~SSIM, and (c)~LPIPS.}
    \label{fig:correlation_scatter}
\end{figure}

\figref{fig:correlation_scatter} shows the scatter plots of PSNR, SSIM, and LPIPS scores against the F1 score, alongside their respective Pearson correlation coefficients. This visualization offers an immediate visual assessment of the relationships between these metrics, and allows for a nuanced understanding of how accurately each metric predicts the true F1 score. The exceptionally strong negative correlation between LPIPS and F1 score (-0.82) reinforces the notion that LPIPS effectively captures the perceptual similarity between the reconstructed and ground truth point clouds, making it a reliable proxy for F1 score, the ultimate measure of reconstruction fidelity.

\begin{algorithm}[t!]
\caption{Plateau Detection Algorithm with LPIPS}
\begin{algorithmic}[1]
\Procedure{DetectPlateau}{$\mathcal{S}, \mathcal{G}, \theta, C$} \Comment{Inputs: Sets of images per iteration $\mathcal{S}$, Sets of GT images $\mathcal{G}$, threshold $\theta$, consistency length $C$}
    \State Initialize an empty list $\mathcal{M}$ to store average LPIPS values over training iterations.
    \For{each set $\mathcal{I}$ and corresponding GT set $\mathcal{G}_I$ in $\mathcal{S}$ and $\mathcal{G}$}
        \State Initialize $\text{sumLPIPS} = 0$.
        \For{each image $I$ and corresponding GT image $G$ in $\mathcal{I}$ and $\mathcal{G}_I$}
            \State $\text{sumLPIPS}$ += $\text{LPIPS}(I, G)$.
        \EndFor
        \State Compute average LPIPS for the set: $\text{avgLPIPS} = \text{sumLPIPS} / |\mathcal{I}|$.
        \State Append $\text{avgLPIPS}$ to $\mathcal{M}$.
    \EndFor
    \If{$|\mathcal{M}| < C$}
        \State \Return $0$ \Comment{Insufficient data for plateau detection}
    \EndIf
    \For{$i = 1$ to $|\mathcal{M}| - 1$}
        \State Let $\text{consistent} = \text{True}$.
        \For{$j = \max(0, i - C + 1)$ to $i$}
            \If{$|\mathcal{M}[j] - \mathcal{M}[j - 1]| \geq \theta$}
                \State Set $\text{consistent} = \text{False}$ and \textbf{break}.
            \EndIf
        \EndFor
        \If{$\text{consistent}$}
            \State \Return $i$ \Comment{Plateau point detected, Output: $P$}
        \EndIf
    \EndFor
    \State \Return $|\mathcal{M}| - 1$ \Comment{No plateau detected, Output: $P$}
\EndProcedure
\end{algorithmic}
\end{algorithm}

The significant negative correlation between LPIPS and the F1 score~(-0.82), PSNR~(-0.81), and SSIM~(-0.69) underscore the impact of LPIPS on the quality of 3D reconstruction (see supplementary material for detailed correlation matrix). The high magnitude of these coefficients, particularly the -0.82 with the F1 score, indicates that LPIPS is a robust predictor of reconstruction accuracy: as the perceptual similarity measure improves (meaning LPIPS decreases), the fidelity of the reconstructed point cloud to the ground truth improves correspondingly. This observation not only suggests the utility of LPIPS as a stand-in metric when the ground truth is unavailable but also highlights its potential as a more influential factor than traditional metrics such as PSNR (0.58) and SSIM (0.37) in determining the overall quality of NeRF-generated reconstructions.

Given this strong correlation, LPIPS emerges as a promising surrogate metric for early stopping during NeRF training. By monitoring LPIPS, one can infer the likely F1 score and make informed decisions about halting the training process. This method could decrease computational costs and time, as one need not await the completion of full training to predict its efficacy in terms of F1 score.

\textbf{Algorithm for Plateau Detection:}
The plateau detection algorithm identifies a stabilization point in a series of metric values, such as LPIPS. The updated algorithm computes the average Learned Perceptual Image Patch Similarity (LPIPS) for each set of images in \(\mathcal{S}\) against their corresponding ground truth images in \(\mathcal{G}\). It then assesses the sequence of these average LPIPS values to identify a plateau, using a specified threshold \(\theta\) and a consistency length \(C\). The detection of the plateau point \(P\) is crucial for indicating an optimal stopping point in the training process. To validate the efficacy of the early stopping algorithm, we applied it to a diverse dataset comprising five plant types captured in both indoor and outdoor settings. The threshold ($\theta$) was set to 0.005, and the consistency length (C) was fixed at 6. The granularity of interpolation was set to 1000, spanning a total of 60000 training iterations. These hyperparameters were chosen based on empirical observations to ensure a balance between computational efficiency and reconstruction accuracy.

\section{Results}
\label{Sec:Results}

\begin{figure*}[!t]
    \centering
    \begin{subfigure}{0.32\linewidth}
        \centering
        \includegraphics[width=0.99\linewidth]{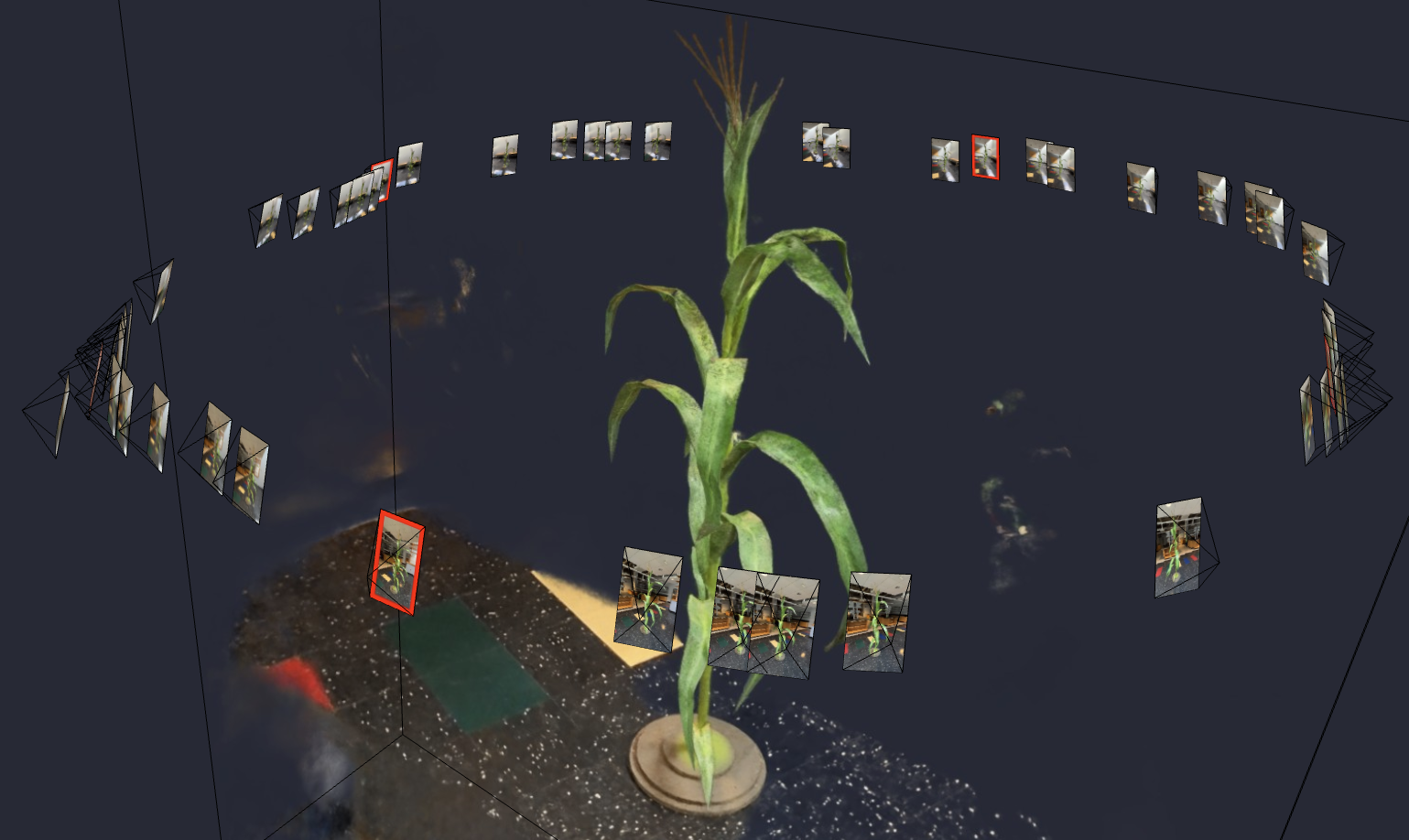}
        \caption{Scenario-I.}
        \label{fig:scenario1}
    \end{subfigure}
    \begin{subfigure}{0.32\linewidth}
        \centering
        \includegraphics[width=0.99\linewidth]{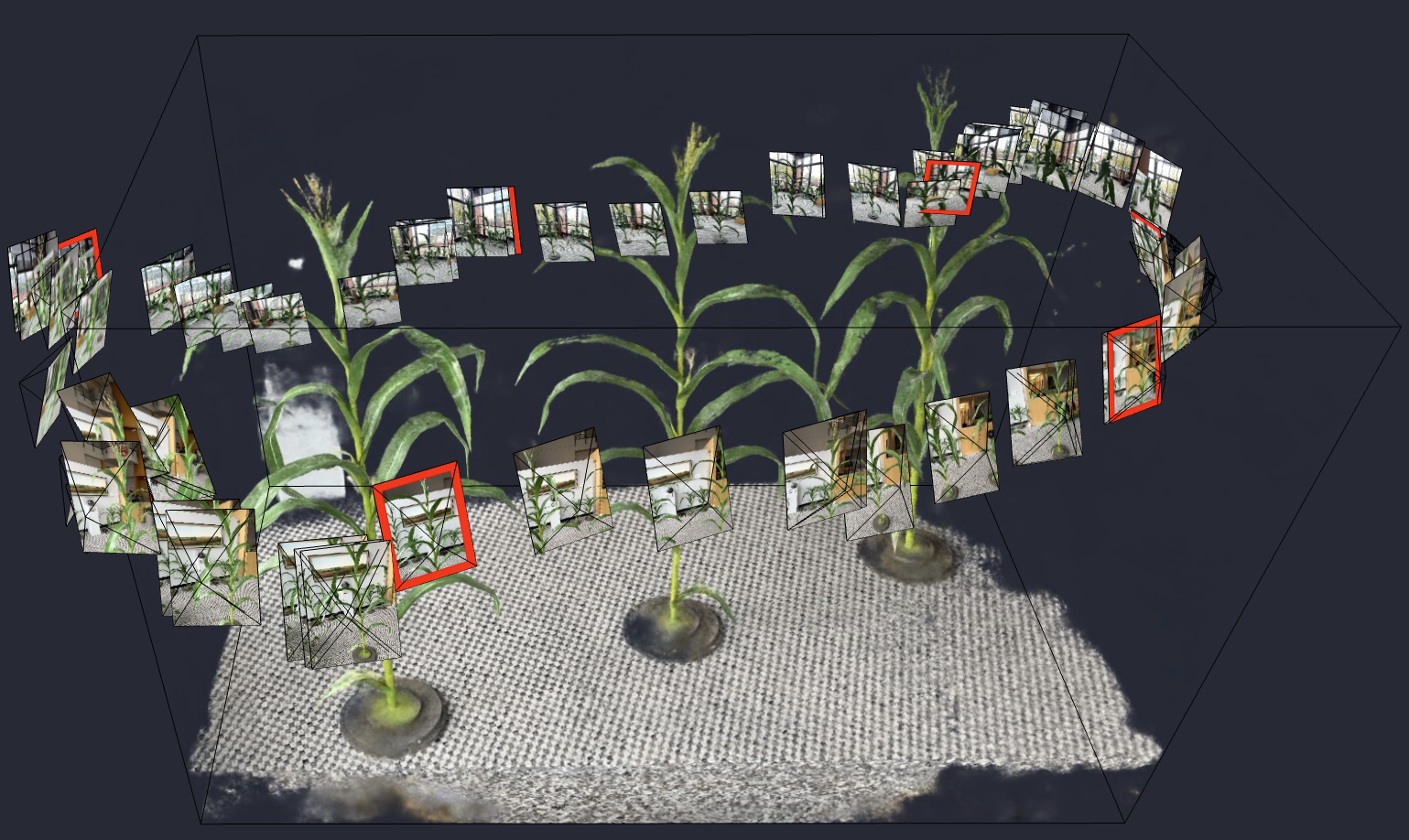}
        \caption{Scenario-II.}
        \label{fig:scenario2}
    \end{subfigure}
    \begin{subfigure}{0.32\linewidth}
        \centering
        \includegraphics[width=0.99\linewidth]{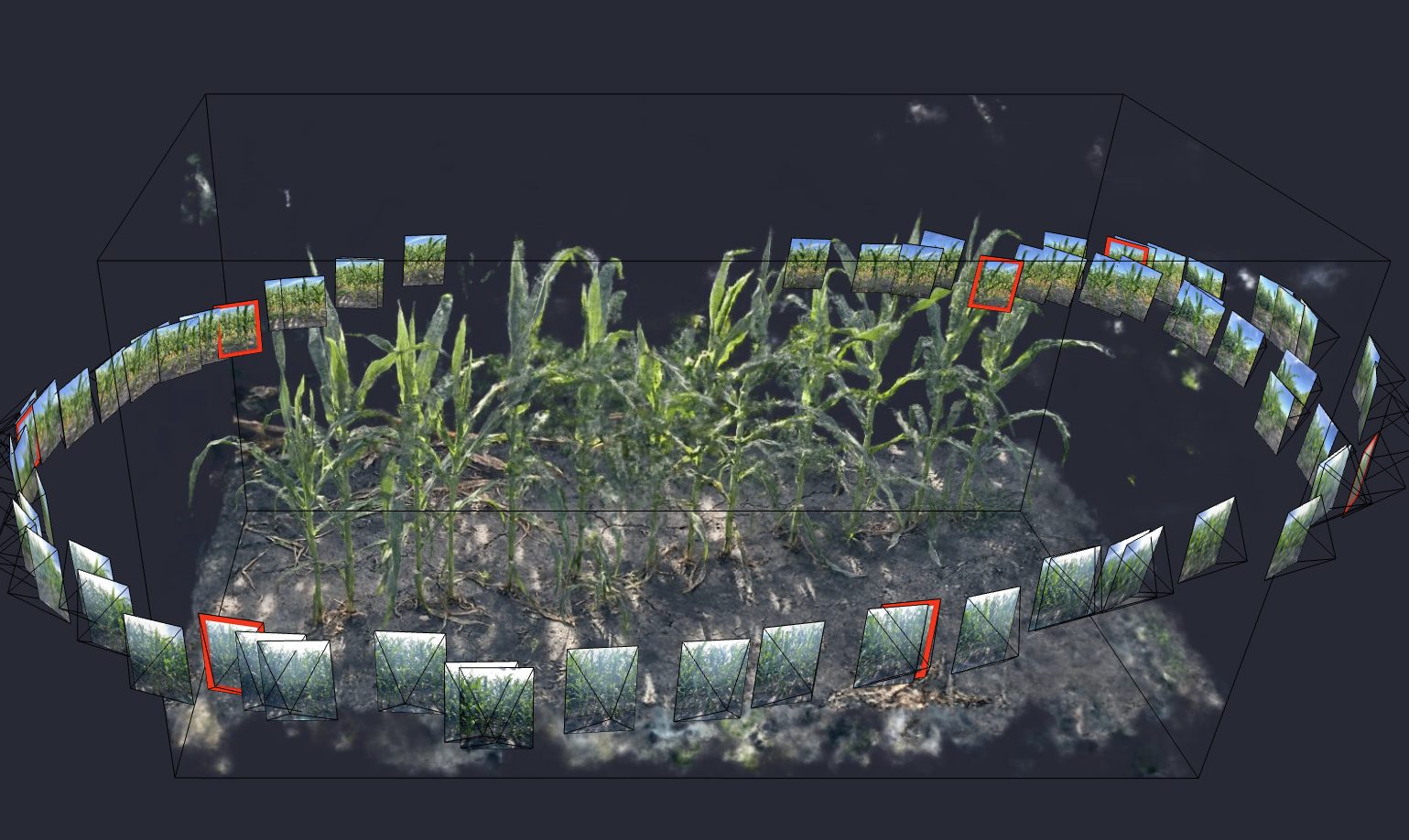}
        \caption{Scenario-III.}
        \label{fig:scenario3}
    \end{subfigure}
    \caption{Camera pose estimations across three different scenarios. (a) Scenario I, (b) Scenario II, (c) Scenario III.}
    \label{fig:scenarios}
\end{figure*}

We evaluated the performance of NeRF models across various scenarios, from controlled indoor environments to complex outdoor field conditions, using key performance metrics to assess their efficacy in 3D plant reconstruction. The NeRFs were trained on an NVIDIA A100 GPU with 80GB GPU RAM attached to an AMD EPYC 7543 32-core CPU with 503GB CPU RAM. Post-training, the models are converted into point clouds with approximately a million points each. Estimated camera poses from COLMAP are visualized in \figref{fig:scenarios}, and a summary of the performance metrics of each of the three scenarios is given in \tabref{tab:table_scenario_all_short}. 3D evaluation metrics are presented in this section; for a more granular analysis of 2D image metrics, please refer to Supplement. Visually, the performance of each model could be assessed using Precision and Recall as shown in \figref{fig:precision_recall}. The Precision-Recall curves of the different Scenarios for different threshold values are shown in \figref{fig:all_curves_scenario}.

\textbf{Visualization Color Code:} The color-coded visualizations employed provide an intuitive understanding of spatial relationships within the 3D reconstructed plant structures. The interpretation of colors is as follows:
\begin{itemize}
\item \textbf{Grey}: (Correct) Represents points within a predefined distance threshold relative to the reference point cloud. This color indicates accurate points in precision and recall evaluations, where precision assesses the reconstruction against the ground truth, and recall evaluates the ground truth against the reconstruction.
\item \textbf{Red}: (Missing) Depicts points in the point cloud being tested that are beyond the distance threshold but within 3 standard deviations from the nearest point in the reference point cloud. These points are considered inaccuracies, showing missing details in the reconstruction when assessing precision and highlighting missing elements in the ground truth during recall analysis.
\item \textbf{Black}: (Outlier) Highlights points in the point cloud being tested that are more than 3 standard deviations away from any point in the reference point cloud. These points are extreme outliers and represent significant errors in the reconstruction relative to the ground truth for precision evaluations, and similarly significant discrepancies in the ground truth relative to the reconstruction for recall.
\end{itemize}

\begin{table*}[!t]
    \centering
    \caption{Performance metrics of NeRFs reconstruction techniques across scenarios I, II and III.}
    \begin{filecontents*}{Data/scenario-all-short.csv}
Scenario,Model Name,Precision,Recall,F1 Score,PSNR,SSIM,LPIPS,Time Taken (s)
,Instant-NGP,24.66,90.62,38.77,23.41,0.81,0.17,756
I,TensoRF,9.58,43.34,15.69,14.69,0.55,0.66,1973
,NeRFacto,73.57,94.72,82.81,22.24,0.73,0.12,1938
,,,,,,,,
,Instant-NGP,23.45,58.57,33.49,19.08,0.64,0.31,1886
II,TensoRF,20.5,55.34,29.91,15.54,0.42,0.56,2607
,NeRFacto,64.47,76.8,70.1,18.93,0.64,0.25,1226
,,,,,,,,
,Instant-NGP,15.06,59.55,24.04,18.54,0.47,0.4,1466
III,TensoRF,40.95,75.62,53.13,17.32,0.39,0.55,1965
,NeRFacto,68.29,82.32,74.65,16.7,0.32,0.34,1499
\end{filecontents*}
\pgfplotstabletypeset[
        col sep=comma,
        string type,
        every head row/.style={
            before row={\toprule},
            after row={\midrule}
        },
        every last row/.style={after row={\bottomrule}},
        columns/Scenario/.style={column name=\#, column type=l},
        columns/Model Name/.style={column name=Model, column type=l},
        columns/Precision/.style={column name=Precision $\uparrow$, column type=r},
        columns/Recall/.style={column name=Recall $\uparrow$, column type=r},
        columns/F1 Score/.style={column name=F1 $\uparrow$, column type=r},
        columns/PSNR/.style={column name=PSNR $\uparrow$, column type=r},
        columns/SSIM/.style={column name=SSIM $\uparrow$, column type=r},
        columns/LPIPS/.style={column name=LPIPS $\downarrow$, column type=r},
        columns/Time Taken (s)/.style={column name=Time (s) $\downarrow$, column type=r}
    ]{Data/scenario-all-short.csv}
    \label{tab:table_scenario_all_short}
\end{table*}

\subsection{Scenario I - Single Plants Indoors}

We first look at the results of reconstructing a single plant in an indoor environment. Detailed evolution of each metric over training iterations is given in the Supplement.

\begin{figure*}[!t]
    \centering
    \footnotesize
    \begin{tblr}
    {
      colspec = {X[{0.015\linewidth},c,t]X[{0.29\linewidth},c,t]X[{0.29\linewidth},c,t]X[{0.29\linewidth},c,t]},
      stretch = 0,
      rowsep = 3pt,
    }
    {\rotatebox{90}{Precision}} &
    \includegraphics[width=0.3\linewidth]{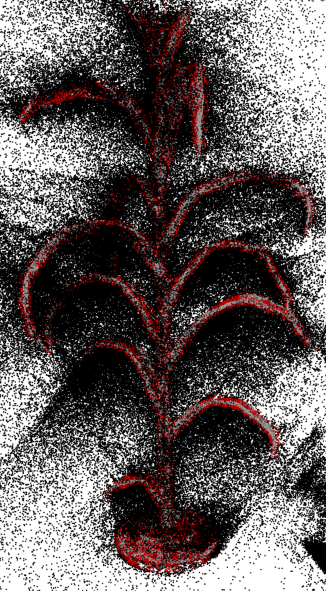} &
    \includegraphics[width=0.3\linewidth]{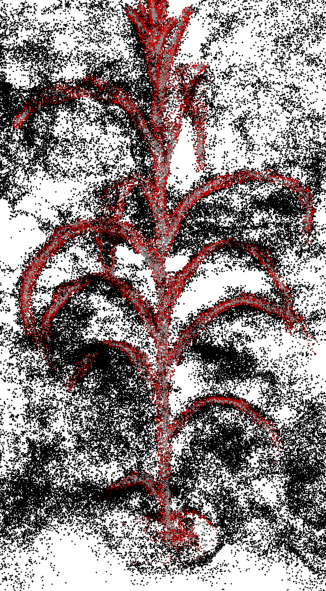} &
    \includegraphics[width=0.3\linewidth]{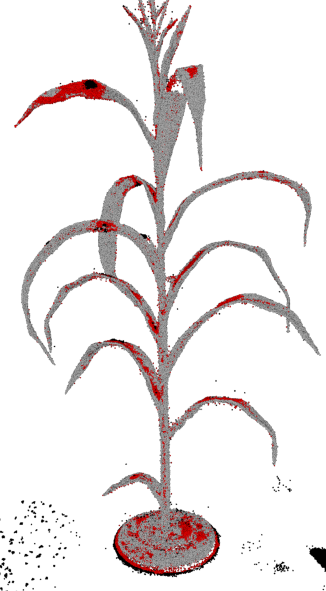}\\
    \rotatebox{90}{{Recall}} &
    \includegraphics[width=0.3\linewidth]{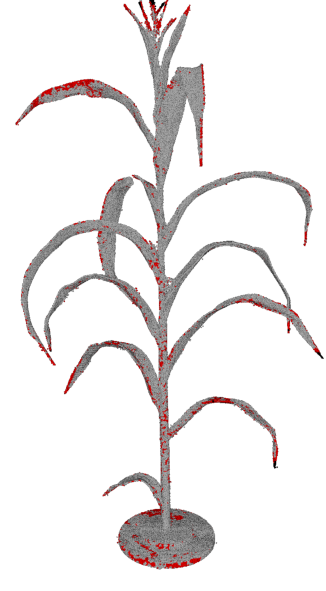} &
    \includegraphics[width=0.3\linewidth]{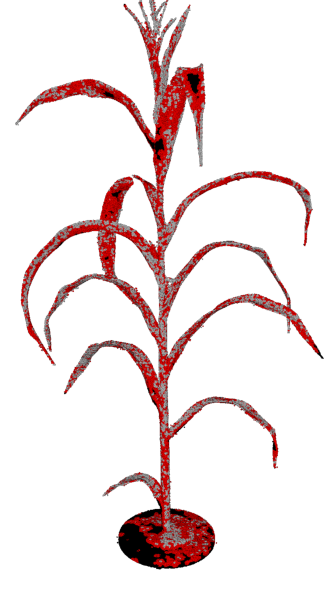} &
    \includegraphics[width=0.3\linewidth]{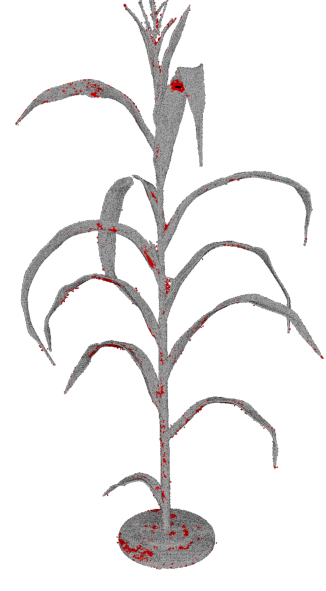} \\
    & & Scenario I &  \\
    \rotatebox{90}{{Precision}} &
    \includegraphics[width=0.99\linewidth]{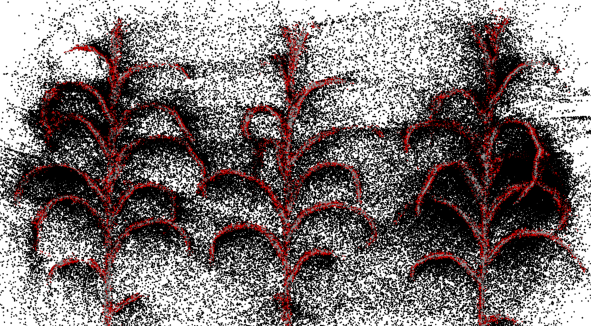} &
    \includegraphics[width=0.99\linewidth]{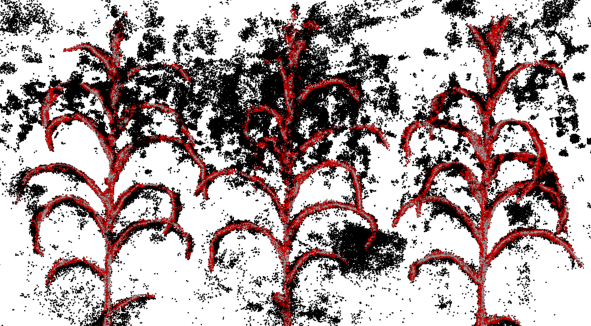} &
    \includegraphics[width=0.99\linewidth]{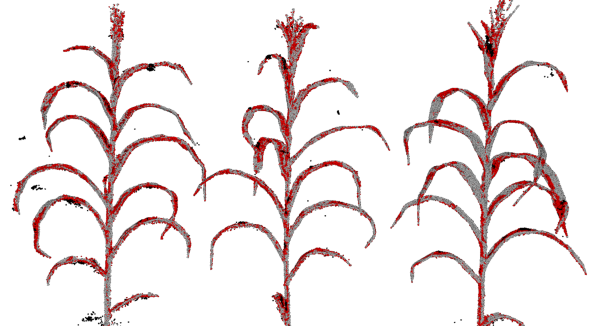}\\
    \rotatebox{90}{{Recall}} &
    \includegraphics[width=0.99\linewidth]{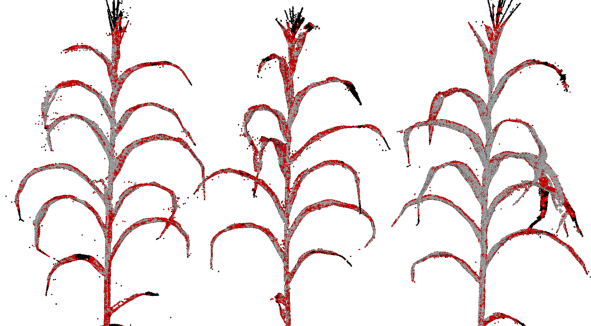} &
    \includegraphics[width=0.99\linewidth]{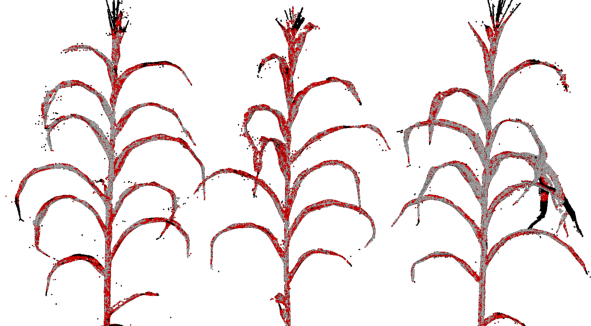} &
    \includegraphics[width=0.99\linewidth]{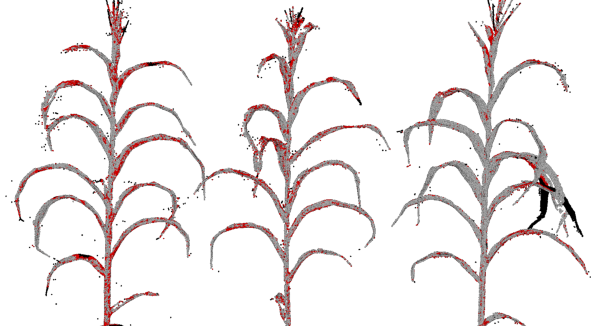} \\
     & & Scenario II &  \\
    \rotatebox{90}{{Precision}} &
    \includegraphics[width=0.99\linewidth]{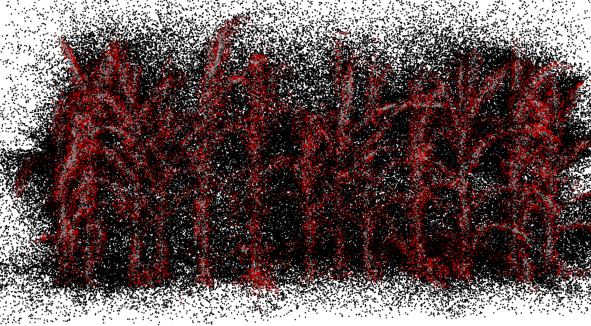} &
    \includegraphics[width=0.99\linewidth]{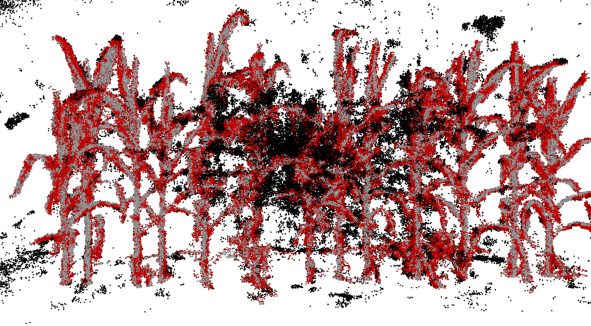} &
    \includegraphics[width=0.99\linewidth]{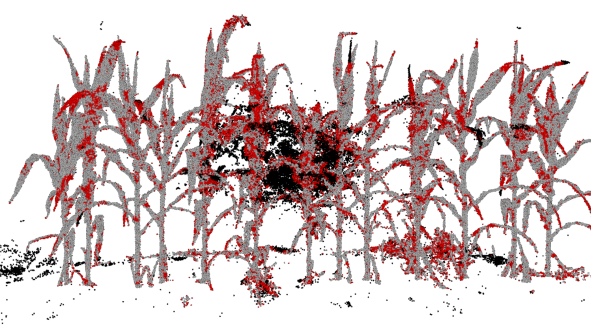}\\
    \rotatebox{90}{{Recall}} &
    \includegraphics[width=0.99\linewidth]{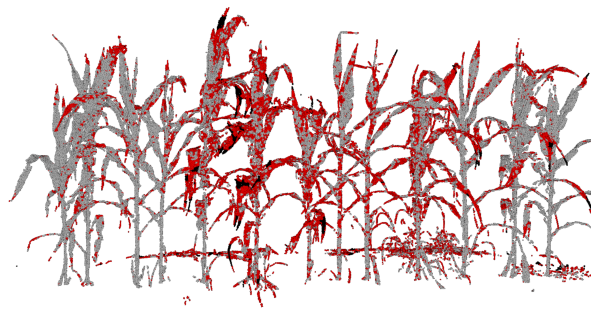} &
    \includegraphics[width=0.99\linewidth]{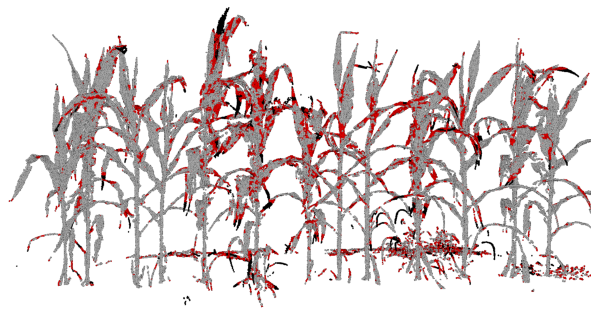} &
    \includegraphics[width=0.99\linewidth]{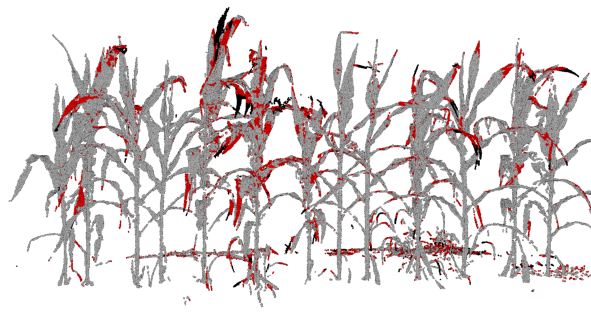} \\
    & & Scenario III &  \\
    &  Instant-NGP & TensoRF & NeRFacto \\
    \end{tblr}
    \caption{Precision and recall of 3D reconstruction using different NeRF techniques across different scenarios. Legend: $\filledsquare{gray}$~Correct, $\filledsquare{red}$~Missing, $\filledsquare{black}$~Outlier.}
    \label{fig:precision_recall}
\end{figure*}

\noindent\textbf{Precision:} For Scenario I, NeRFacto, achieved the highest precision followed by TensoRF and Instant-NGP (see \figref{fig:precision_recall}) after 30000 iterations. Across all models, precision generally increases with the number of iterations. For a detailed evaluation of the change of precision with iterations, please refer the Supplement.

\noindent\textbf{Recall:} The recall metric follows a similar trend, with Instant-NGP and NeRFacto showing increases with more iterations, indicating an enhanced ability to encompass points from the ground truth. Notably, NeRFacto achieves remarkably high recall values (over 90) at higher iterations, suggesting its superiority in the completeness of reconstruction. TensoRF's recall values are significantly lower, indicating that it may miss more details from the ground truth compared to the other models.

\noindent\textbf{F1 Score:} The F1 score, balancing precision and recall, highlights NeRFacto as the most balanced model, especially at higher iterations, with scores above 80. Instant-NGP shows a significant improvement in F1 scores as iterations increase, but it doesn't reach the same peak as NeRFacto. TensoRF lags in this metric, indicating a less balanced performance between precision and recall.

\noindent\textbf{Computation Time:} Time efficiency is a crucial factor, especially for practical applications. Instant-NGP demonstrates a relatively balanced approach between efficiency and performance, with time increments correlating reasonably with the increase in iterations. However, it becomes time-consuming at high iterations (20000 and 30000). NeRFacto, while showing better performance in many metrics, demands considerably more time, especially at higher iterations, which could be a limiting factor in time-sensitive scenarios. The evolution of precision over training time for NeRFacto is given in the supplementary material. TensoRF, despite its lower performance in other metrics, maintains a more consistent time efficiency, suggesting its suitability for applications where time is a critical constraint. Please see the Supplement for the evolution of precision metric over the training iterations.

\begin{figure*}[!t]
    \centering
    \begin{minipage}[c]{0.04\linewidth}
        \vspace{-0.4in}\rotatebox{90}{Scenario I}
    \end{minipage}%
    \begin{minipage}[c]{0.02\linewidth}
        \vspace{-0.4in}\rotatebox{90}{\fontfamily{phv}\selectfont \centering \footnotesize{\# of points (\%)}}
    \end{minipage}%
    \begin{minipage}[c]{0.92\linewidth}
        \begin{subfigure}{0.3\linewidth}
            \centering
            \includegraphics[width=\linewidth]{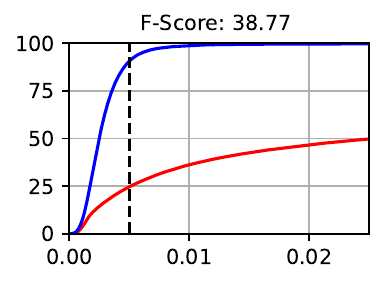}
            \raisebox{0.1in}{\fontfamily{phv}\selectfont\footnotesize{Threshold (m)}}
        \end{subfigure}
        \hfill
        \begin{subfigure}{0.3\linewidth}
            \centering
            \includegraphics[width=\linewidth]{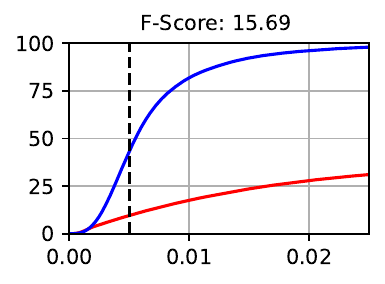}
            \raisebox{0.1in}{\fontfamily{phv}\selectfont\footnotesize{Threshold (m)}}
        \end{subfigure}
        \hfill
        \begin{subfigure}{0.3\linewidth}
            \centering
            \includegraphics[width=\linewidth]{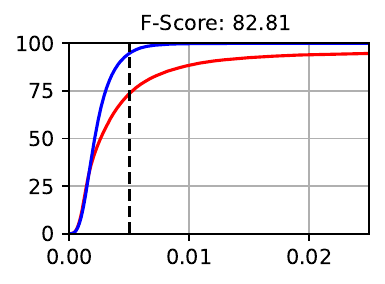}
            \raisebox{0.1in}{\fontfamily{phv}\selectfont\footnotesize{Threshold (m)}}
        \end{subfigure}
    \end{minipage}
    \begin{minipage}[c]{0.04\linewidth}
        \vspace{-0.4in}\rotatebox{90}{Scenario II}
    \end{minipage}%
    \begin{minipage}[c]{0.02\linewidth}
        \vspace{-0.4in}\rotatebox{90}{\fontfamily{phv}\selectfont \centering \footnotesize{\# of points (\%)}}
    \end{minipage}%
    \begin{minipage}[c]{0.92\linewidth}
        \begin{subfigure}{0.3\linewidth}
            \centering
            \includegraphics[width=\linewidth]{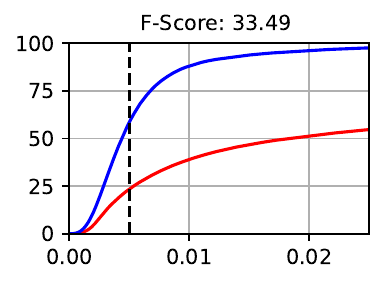}
            \raisebox{0.1in}{\fontfamily{phv}\selectfont\footnotesize{Threshold (m)}}
        \end{subfigure}
        \hfill
        \begin{subfigure}{0.3\linewidth}
            \centering
            \includegraphics[width=\linewidth]{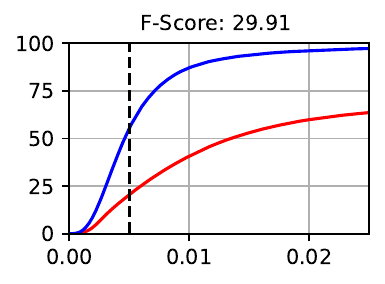}
            \raisebox{0.1in}{\fontfamily{phv}\selectfont\footnotesize{Threshold (m)}}
        \end{subfigure}
        \hfill
        \begin{subfigure}{0.3\linewidth}
            \centering
            \includegraphics[width=\linewidth]{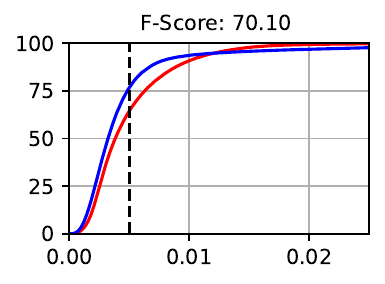}
            \raisebox{0.1in}{\fontfamily{phv}\selectfont\footnotesize{Threshold (m)}}
        \end{subfigure}
    \end{minipage}
    \begin{minipage}[c]{0.04\linewidth}
        \vspace{-0.4in}\rotatebox{90}{Scenario III}
    \end{minipage}%
    \begin{minipage}[c]{0.02\linewidth}
        \vspace{-0.4in}\rotatebox{90}{\fontfamily{phv}\selectfont \centering \footnotesize{\# of points (\%)}}
    \end{minipage}%
    \begin{minipage}[c]{0.92\linewidth}
        \begin{subfigure}{0.3\linewidth}
            \centering
            \includegraphics[width=\linewidth]{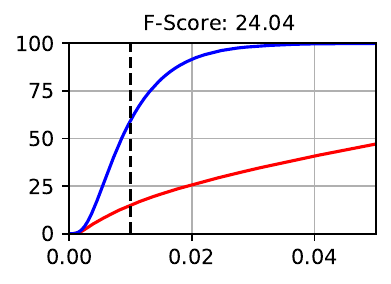}
            \raisebox{0.1in}{\fontfamily{phv}\selectfont\footnotesize{Threshold (m)}}
            \caption{Instant-NGP}
        \end{subfigure}
        \hfill
        \begin{subfigure}{0.3\linewidth}
            \centering
            \includegraphics[width=\linewidth]{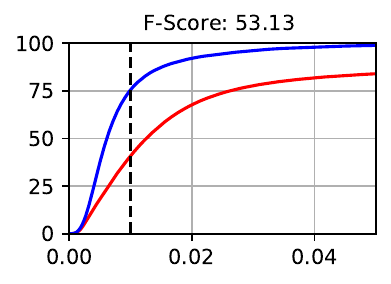}
            \raisebox{0.1in}{\fontfamily{phv}\selectfont\footnotesize{Threshold (m)}}
            \caption{TensoRF}
        \end{subfigure}
        \hfill
        \begin{subfigure}{0.3\linewidth}
            \centering
            \includegraphics[width=\linewidth]{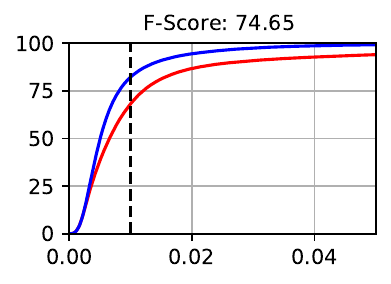}
            \raisebox{0.1in}{\fontfamily{phv}\selectfont\footnotesize{Threshold (m)}}
             \caption{NeRFacto}
        \end{subfigure}
    \end{minipage}
    \vspace{1ex}
    \vspace{1ex}
    \raisebox{0.5ex}{\textcolor{red}{\rule{1cm}{1pt}}} Precision \quad
    \raisebox{0.5ex}{\textcolor{blue}{\rule{1cm}{1pt}}} Recall
    \caption{Precision-Recall curves for the three scenarios based on varying distance threshold after 30,000 iterations (Scenario I and II) and 60,000 iterations (Scenario III).}
    \label{fig:all_curves_scenario}
\end{figure*}

\noindent\textbf{Overall Performance and Suitability:} In sum, NeRFacto emerges as the most robust model in terms of precision, recall, F1 score, and image quality metrics (PSNR, SSIM, LPIPS), making it highly suitable for applications demanding high accuracy and completeness in 3D modeling. However, its time inefficiency at higher iterations might restrict its use in time-sensitive contexts. Instant-NGP presents a good balance between performance and efficiency, making it a viable option for moderately demanding scenarios. Detailed results are given in \tabref{tab:table_scenario_all_short}, after complete training. For more granular look of each metric value over the training iteration for all the algorithms, consult the supplementary. The Precision-Recall curves based on varying distance threshold after maximum training of 30,000 iterations is given in \figref{fig:all_curves_scenario}.

\noindent\textbf{Insight 1: Computational Cost and Accuracy Trade-off in Instant-NGP and NeRFacto:} The steep increase in performance metrics with the number of iterations for both Instant-NGP and NeRFacto suggests that these models require a substantial amount of data processing to achieve high accuracy, which is critical in high-fidelity 3D modeling. However, this also implies a higher computational cost, which needs to be considered in practical applications.

\noindent\textbf{Insight 2: Model Suitability in High-Detail 3D Reconstructions:} The significant disparity in the performance of TensoRF compared to the other two models, particularly in precision and recall, indicates that not all NeRF models are equally suited for tasks requiring high-detail 3D reconstructions. This highlights the importance of model selection based on the specific requirements of the application.

\noindent\textbf{Insight 3: Divergence in 2D Image Quality and 3D Reconstruction in Instant-NGP:} A detailed examination reveals that Instant-NGP demonstrates strength in 2D image quality metrics such as PSNR, SSIM, and LPIPS, reflecting its ability to produce better rendered image quality. However, this excellence in 2D imaging does not correspondingly extend to 3D reconstruction metrics like Precision, Recall, and F1 Score. This observation highlights a significant distinction in the challenges associated with optimizing for high-quality image rendering as opposed to achieving accurate 3D representations. The model's adeptness at rendering highly detailed 2D images does not necessarily imply its effectiveness in accurately reconstructing complex 3D structures, particularly in the context of intricate plant models. This insight underscores the need for a nuanced approach in evaluating the performance of models that are tasked with both 2D image rendering and 3D spatial reconstruction.

\subsection{Scenario II - Multiple Plants Indoors}
We observe marked differences in model behaviors compared to the single plant scenario, likely attributed to the added intricacy of multiple plants in a single scene. Detailed evolution of each metric over training iterations is given in the Supplement.

\noindent\textbf{Precision:} As shown in \figref{fig:precision_recall}, Instant-NGP exhibits a steady increase in precision with more iterations, peaking at a high value. However, NeRFacto starts at a higher precision and reaches an even higher peak, indicating a more accurate reconstruction of the corn plants. TensoRF, although improving with more iterations, lags behind the others in terms of precision.

\noindent\textbf{Recall:} A similar pattern is observed for recall, with NeRFacto consistently maintaining a higher recall compared to the other methods, suggesting its ability to better encompass points in the ground truth. Both Instant-NGP and TensoRF exhibit increasing recall with more iterations, but at lower levels than NeRFacto.

\noindent\textbf{F1 Score:} The F1 Score, balancing precision and recall, follows a similar trend. NeRFacto demonstrates the best balance between precision and recall, with its F1 score peaking at 70.10, while Instant-NGP and TensoRF achieve lower peak F1 scores.

\noindent\textbf{Computation Time:} The time taken for iterations is crucial for efficiency. Instant-NGP and NeRFacto have comparable times, but TensoRF takes significantly longer at higher iterations, indicating less time efficiency.  See supplementary material for evolution of precision metric over the training iterations.

\noindent\textbf{Overall Performance and Suitability:} NeRFacto emerges as the most balanced and efficient model, exhibiting high precision, recall, and F1 scores, along with favorable PSNR, SSIM, and LPIPS values. Its efficiency in time taken is also comparable to Instant-NGP. Instant-NGP, while showing improvements, doesn’t quite match NeRFacto’s balance of precision and recall. TensoRF, despite its merits, falls behind in several key metrics, particularly in precision, recall, SSIM, and LPIPS. The results after complete training are given in \tabref{tab:table_scenario_all_short}. For more granular look of each metric value over the training iteration for all the algorithms, consult the supplementary. The Precision-Recall curves based on varying distance thresholds after maximum training of 30,000 iterations are given in \figref{fig:all_curves_scenario}.

\noindent\textbf{Insight 1: Improved Performance of TensoRF in Scenario II:} In the second scenario, TensoRF demonstrated an improvement compared to its performance in the first scenario. Specifically, its F1 score, a critical metric for 3D modeling accuracy, increased from 15.69 in the first scenario to 29.91 after 30,000 iterations in the second scenario. This improvement highlights TensoRF's potential in more complex or demanding 3D modeling tasks, especially when allowed to complete its training process.

\noindent\textbf{Insight 2: 2D Metrics Versus 3D F1 Score for Instant-NGP and NeRFacto:} While Instant-NGP and NeRFacto show comparable results in 2D image quality metrics such as PSNR and SSIM, a distinct difference is observed in their 3D modeling capabilities, as reflected in their F1 scores, as observed in last scenario. This suggests that NeRFacto might be a more reliable choice for applications requiring high accuracy in 3D reconstructions.

\subsection{Scenario III - Multiple Plants Outdoors}

Scenario III is the most complex, with multiple overlapping plants captured in field conditions. The models were also trained until 60,000 iterations, while the previous two scenarios were trained only for 30,000 iterations. Detailed evolution of each metric over training iterations is given in the Supplement.

\noindent\textbf{Precision:} As observed in \figref{fig:precision_recall}, NeRFacto consistently demonstrates the highest precision across all iterations, peaking at 68.29\%, suggesting its ability to reconstruct points close to the ground truth. Instant-NGP shows a steady increase in precision with more iterations, while TensoRF, although starting lower, reaches a comparable precision to Instant-NGP at higher iterations.

\noindent\textbf{Recall:} NeRFacto leads in recall, achieving a high of 82.32\%, indicating its effectiveness in encompassing points from the ground truth. Instant-NGP shows significant improvement in recall with increased iterations, but remains behind NeRFacto. TensoRF's recall growth positions it between Instant-NGP and NeRFacto in terms of completeness.

\noindent\textbf{F1 Score:} Reflecting the balance between precision and recall, NeRFacto emerges as the superior model, with its F1 score peaking at 74.65\%. Instant-NGP's F1 score improves with more iterations but remains significantly lower, while TensoRF's F1 score surpasses Instant-NGP, reaching 53.13\%.

\noindent\textbf{Computation Time:} In terms of efficiency, Instant-NGP and NeRFacto are the fastest, followed by TensoRF. 

\noindent\textbf{Overall Performance and Suitability:} NeRFacto again emerges as the most balanced and robust model, excelling in precision, recall, F1 score, and LPIPS. Detailed results are given in \tabref{tab:table_scenario_all_short}, after complete training. For more granular look of each metric value over the training iteration for all the algorithms, consult the supplementary. The Precision-Recall curves based on varying distance threshold after maximum training of 60,000 iterations is given in \figref{fig:all_curves_scenario}. The GPU memory usage of this scenario comes out to be approximately a constant 3GB (for the total memory of GPU being 80GB). 

\noindent\textbf{Insight 1: Enhanced Performance of TensoRF in Outdoor Settings:} TensoRF demonstrates significant improvement in its performance in the third scenario compared to the first. Specifically, its F1 score has seen a good increase; from 15.69 in the first scenario to 29.91 in the second, and reaching 53.13 after 30,000 iterations in the current outdoor scenario. This upward trajectory in F1 scores, which is a balanced measure of precision and recall, indicates TensoRF's enhanced capability in outdoor environments, potentially outperforming Instant-NGP in these settings. This suggests that TensoRF might be a more suitable choice for outdoor 3D modeling tasks where both precision and completeness are crucial. This property may have contributed in the selection of TensoRF as a building block for using multiple local radiance fields, during in-the-wild reconstruction~\citep{meuleman2023progressively}.

\noindent\textbf{Insight 2: LPIPS as a Strong Indicator of 3D Model Quality:} The LPIPS metric appears to be a more representative measure of the quality of the resulting 3D models. In the analysis, we observe that models with lower LPIPS scores consistently show better performance across other metrics. This trend indicates the relevance of LPIPS in assessing the perceptual quality of 3D models. The further investigation into how LPIPS correlates with other metrics could provide deeper insights into model performance, especially in the context of realistic and perceptually accurate 3D reconstructions.

\subsection{Early Stopping Algorithm}

The implementation of early stopping based on the LP-IPS metric yielded substantial savings in computational time across all scenarios, with a minor sacrifice in the fidelity of 3D reconstructions, as measured by the F1 score. Time savings were notable across the three tested methodologies—Instant-NGP, TensorRF, and NeRFacto—with each showing a marked decrease in training time without a commensurate loss in F1 score accuracy. For a deeper look of LPIPS, F1 Score and the recommended stopping point for each case, consult the supplementary material. 

On average, the early stopping strategy resulted in a 61.1\% reduction in training time, suggesting a significant efficiency gain in the process of 3D plant reconstruction using neural radiance fields. Concurrently, the average F1 score loss was contained to 7.4\%, indicating that the early plateau detection has a moderate impact on the quality of the 3D point cloud reconstructions. Specifically, Instant-NGP presented a more pronounced variation in F1 score loss, which was notably higher in Scene-III, thereby affecting its average loss more than TensorRF and NeRFacto. TensorRF and NeRFacto showed a remarkable consistency in time savings, which was mirrored in their comparable F1 score losses, highlighting the robustness of these methods in early stopping scenarios.

These findings articulate a compelling case for the utilization of early stopping in NeRF-based 3D reconstruction tasks, emphasizing the need to balance between computational resources and reconstruction precision. Such a balance is pivotal in scenarios where time efficiency is paramount yet a minimal compromise on reconstruction accuracy is permissible.

\subsection{Scenario IV - Validation Examples In Field Conditions}

\begin{figure*}[!t]
    \centering
    \begin{minipage}[c]{0.03\linewidth}
        \rotatebox{90}{\centering Scene-1}
    \end{minipage}%
    \begin{minipage}[c]{0.96\linewidth}
        \begin{subfigure}{0.20\linewidth}
            \centering
            \includegraphics[width=\linewidth, angle=-90]{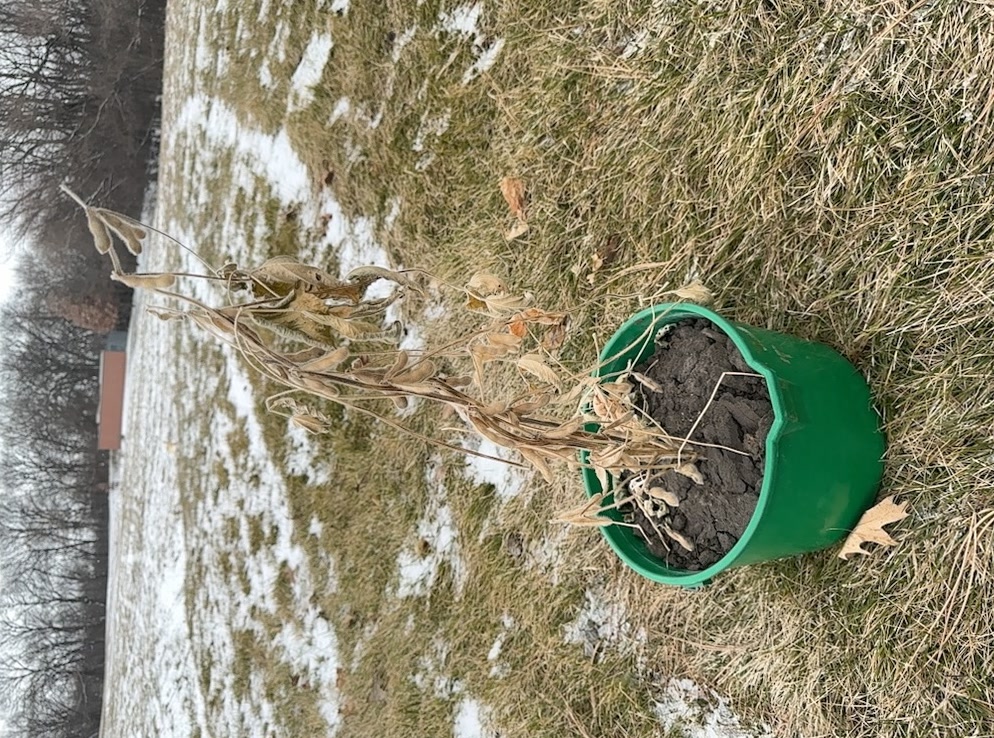}
            \caption*{Soybean}
        \end{subfigure}
        \hfill
        \begin{subfigure}{0.22\linewidth}
            \centering
            \includegraphics[width=\linewidth]{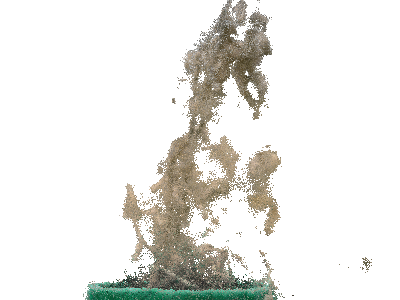}
            \caption*{1000 Iterations}
        \end{subfigure}
        \hfill
        \begin{subfigure}{0.22\linewidth}
            \centering
            \includegraphics[width=\linewidth]{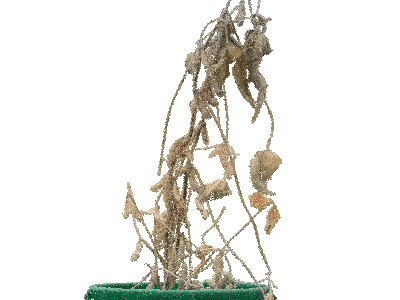}
            \caption*{20000 Iterations*}
        \end{subfigure}
        \hfill
        \begin{subfigure}{0.22\linewidth}
            \centering
            \includegraphics[width=\linewidth]{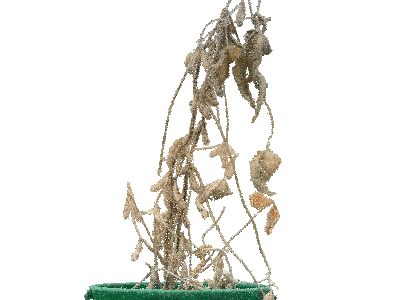}
            \caption*{60000 Iterations}
        \end{subfigure}
    \vspace{6pt}
    \end{minipage}
    \begin{minipage}[c]{0.03\linewidth}
        \rotatebox{90}{\centering Scene-2}
    \end{minipage}%
    \begin{minipage}[c]{0.96\linewidth}
        \begin{subfigure}{0.20\linewidth}
            \centering
            \includegraphics[width=\linewidth, angle=-90]{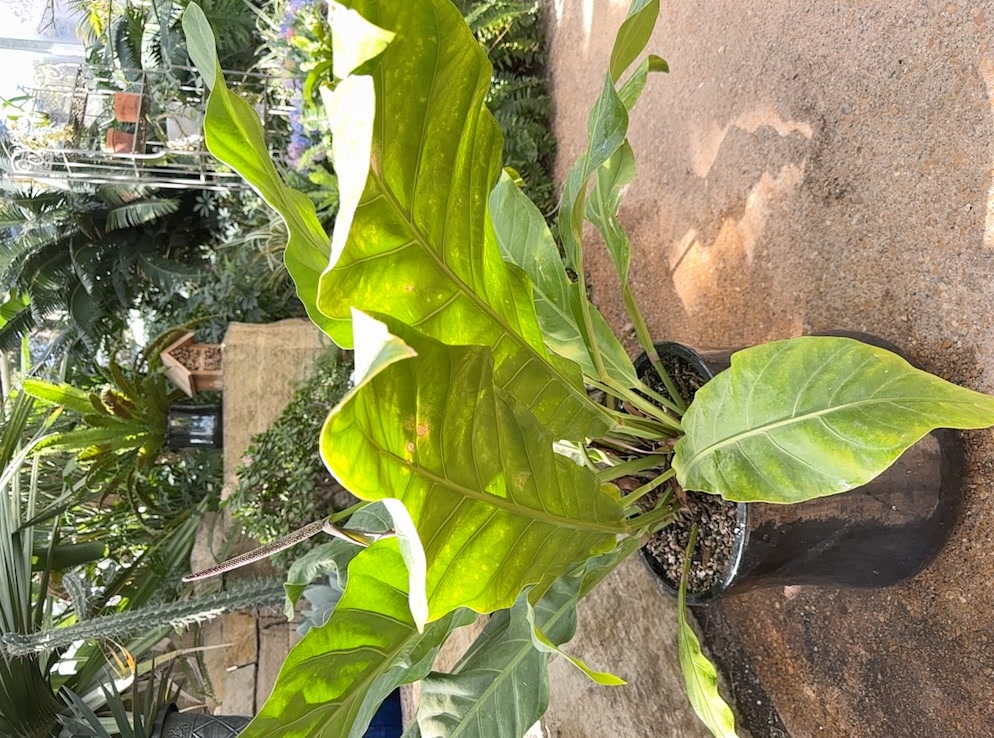}
            \caption*{Anthurium Hookeri}
        \end{subfigure}
        \hfill
        \begin{subfigure}{0.22\linewidth}
            \centering
            \includegraphics[width=\linewidth]{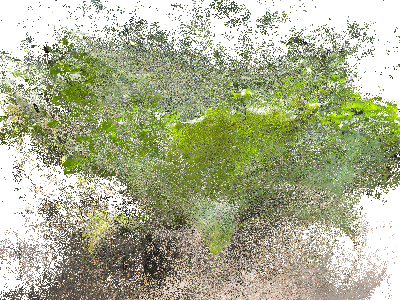}
            \caption*{ 1000 Iterations}
        \end{subfigure}
        \hfill
        \begin{subfigure}{0.22\linewidth}
            \centering
            \includegraphics[width=\linewidth]{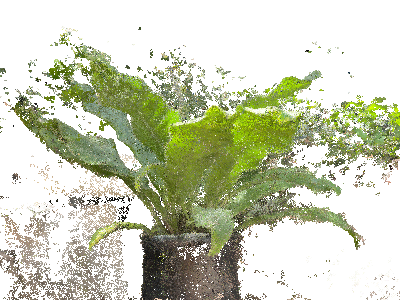}
            \caption*{ 20000 Iterations*}
        \end{subfigure}
        \hfill
        \begin{subfigure}{0.22\linewidth}
            \centering
            \includegraphics[width=\linewidth]{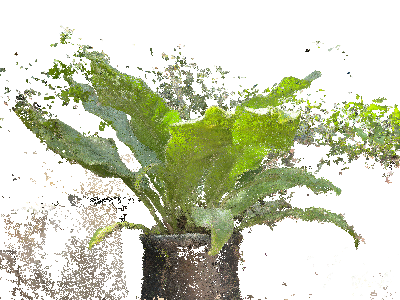}
            \caption*{ 60000 Iterations}
        \end{subfigure}
        \vspace{6pt}
    \end{minipage}
    \begin{minipage}[c]{0.03\linewidth}
        \rotatebox{90}{\centering Scene-3}
    \end{minipage}%
    \begin{minipage}[c]{0.96\linewidth}
        \begin{subfigure}{0.20\linewidth}
            \centering
            \includegraphics[width=\linewidth, angle=-90]{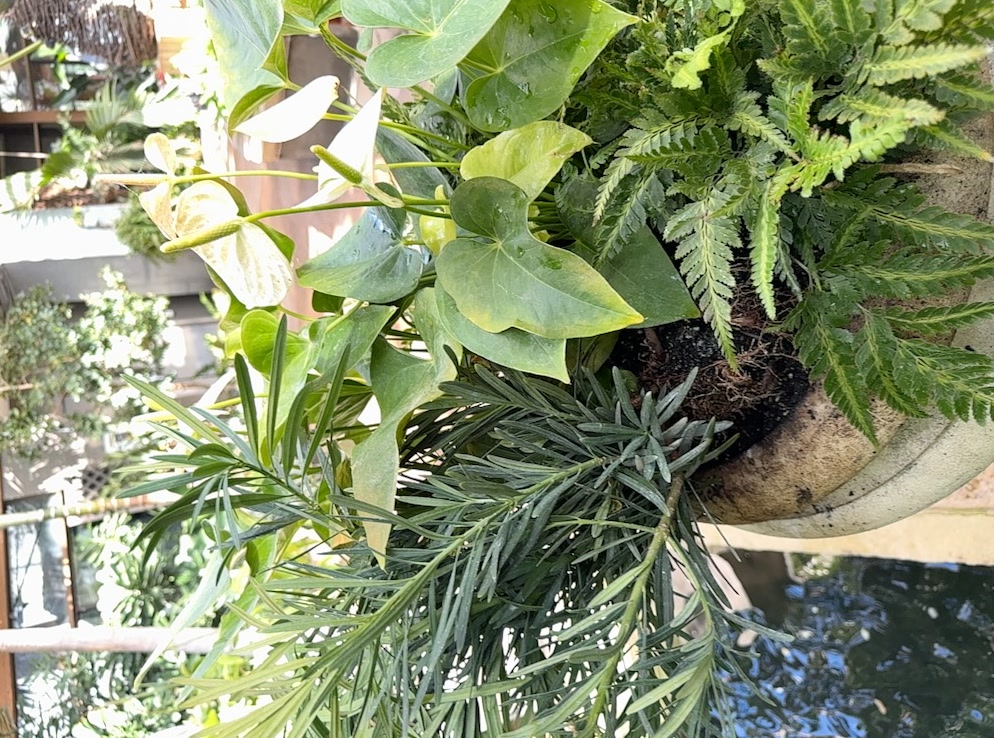}
            \caption*{Mixture of plants}
        \end{subfigure}
        \hfill
        \begin{subfigure}{0.22\linewidth}
            \centering
            \includegraphics[width=\linewidth]{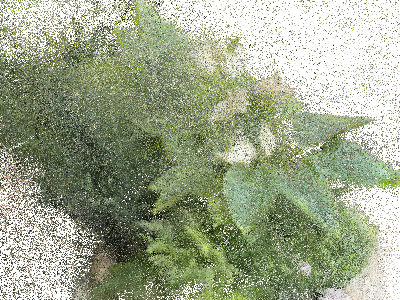}
            \caption*{ 1000 Iterations}
        \end{subfigure}
        \hfill
        \begin{subfigure}{0.22\linewidth}
            \centering
            \includegraphics[width=\linewidth]{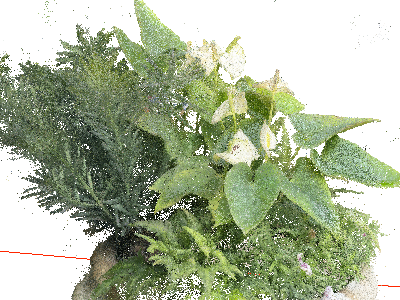}
            \caption*{ 20000 Iterations*}
        \end{subfigure}
        \hfill
        \begin{subfigure}{0.22\linewidth}
            \centering
            \includegraphics[width=\linewidth]{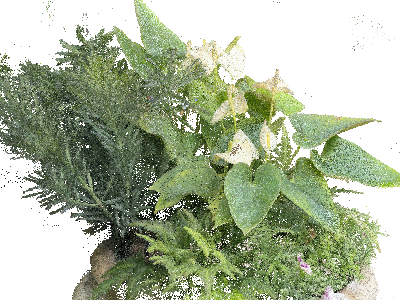}
            \caption*{ 60000 Iterations}
        \end{subfigure}
        \vspace{6pt}
    \end{minipage}
    \begin{minipage}[c]{0.03\linewidth}
        \rotatebox{90}{\centering Scene-4}
    \end{minipage}%
    \begin{minipage}[c]{0.96\linewidth}
        \begin{subfigure}{0.20\linewidth}
            \centering
            \includegraphics[width=\linewidth, angle=-90]{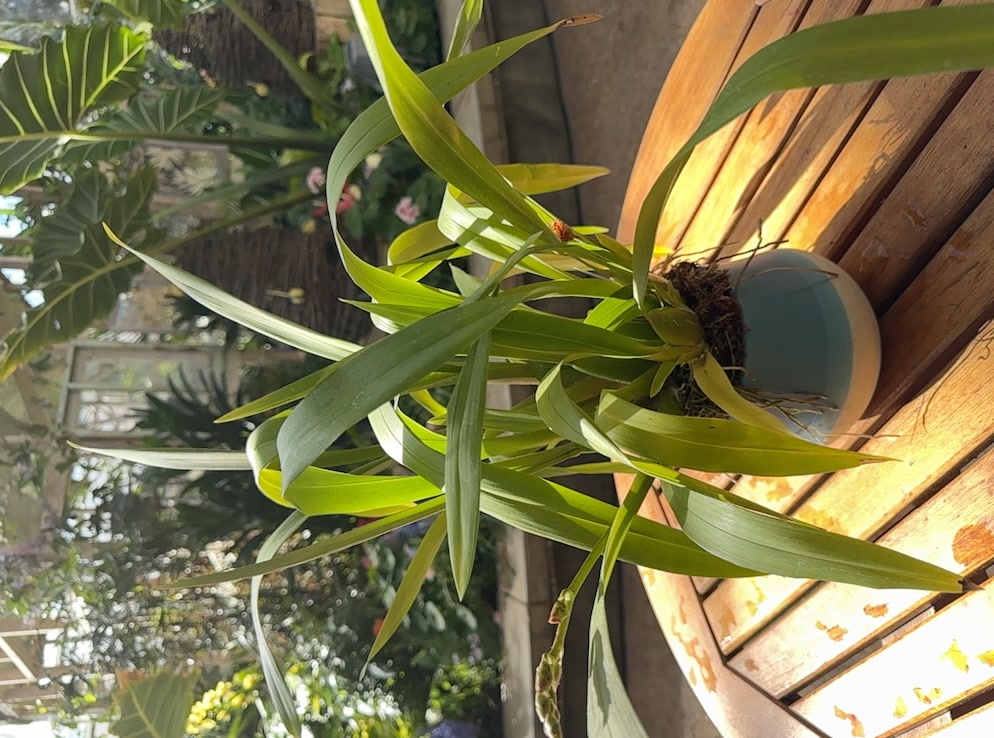}
            \caption*{C. Floribundum}
        \end{subfigure}
        \hfill
        \begin{subfigure}{0.22\linewidth}
            \centering
            \includegraphics[width=\linewidth]{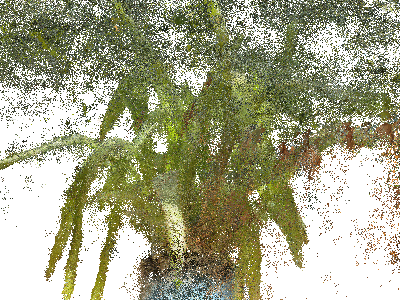}
            \caption*{ 1000 Iterations}
        \end{subfigure}
        \hfill
        \begin{subfigure}{0.22\linewidth}
            \centering
            \includegraphics[width=\linewidth]{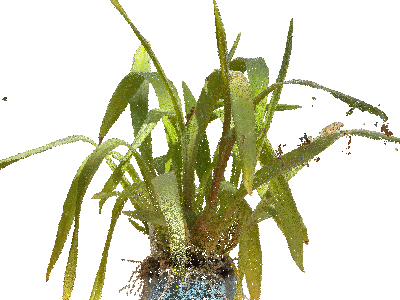}
            \caption*{ 20000 Iterations*}
        \end{subfigure}
        \hfill
        \begin{subfigure}{0.22\linewidth}
            \centering
            \includegraphics[width=\linewidth]{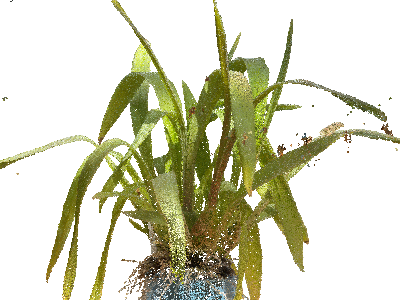}
            \caption*{ 60000 Iterations}
        \end{subfigure}
        \vspace{6pt}
    \end{minipage}
    \begin{minipage}[c]{0.03\linewidth}
        \rotatebox{90}{\centering Scene-5}
    \end{minipage}%
    \begin{minipage}[c]{0.96\linewidth}
        \begin{subfigure}{0.20\linewidth}
            \centering
            \includegraphics[width=\linewidth, angle=-90]{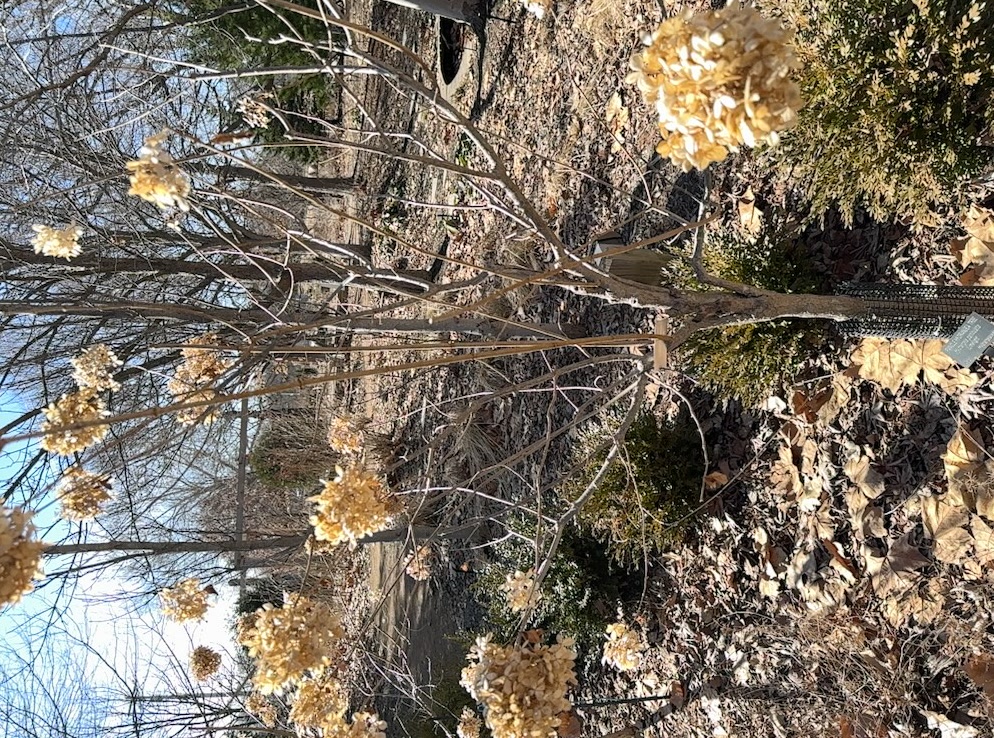}
            \caption*{H. Paniculata}
        \end{subfigure}
        \hfill
        \begin{subfigure}{0.22\linewidth}
            \centering
            \includegraphics[width=\linewidth]{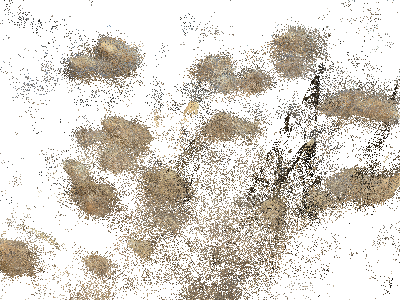}
            \caption*{ 1000 Iterations}
        \end{subfigure}
        \hfill
        \begin{subfigure}{0.22\linewidth}
            \centering
            \includegraphics[width=\linewidth]{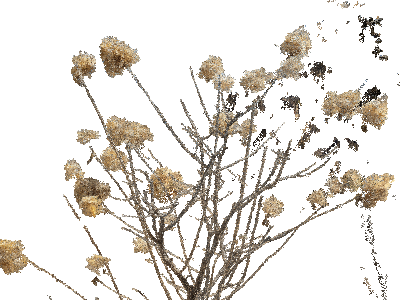}
            \caption*{ 30000 Iterations*}
        \end{subfigure}
        \hfill
        \begin{subfigure}{0.22\linewidth}
            \centering
            \includegraphics[width=\linewidth]{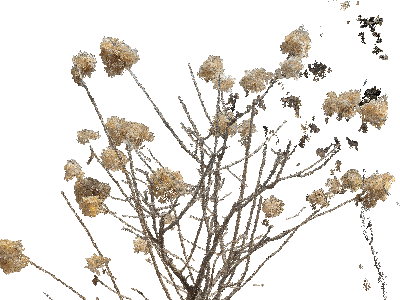}
            \caption*{ 60000 Iterations}
        \end{subfigure}
        \vspace{6pt}
    \end{minipage}
    \caption{Scenes for validating the early stopping algorithm and their 3D reconstructions: original scenes in the first column, iterative reconstructions in the right column, and optimal iterations in the third column~(*).}
    \label{fig:lpips_validation}
\end{figure*}

The efficacy of the LPIPS-based early stopping algorithm was validated using a diverse dataset comprising images from five different types of plants captured in both indoor and outdoor settings, as illustrated in \figref{fig:lpips_validation}. The validation process employed a threshold $\theta$ set to 0.005 and a consistency length $C$ of 6, with the granularity of interpolation fixed at 1000, spanning a total of 60000 training iterations. For practical application, checkpoints, inherently exponential in nature, necessitated linear interpolation to facilitate algorithm execution. \figref{fig:lpips_validation} shows the rendered point clouds at three stages: after 1000 iterations, at the recommended early stopping iteration, and upon completing the full 60000 iterations of training. Each row of the figure corresponds to one of the five validation scenes, providing a qualitative comparative analysis.

Notably, for all indoor scenes, the algorithm recommended halting training at 20000 iterations, whereas for outdoor scenes, the suggestion extended to 30000 iterations. This distinction underscores the algorithm's sensitivity to environmental variables affecting perceptual similarity metrics. The rendered point clouds, particularly at the early stopping points, exhibit minimal visual discrepancies when compared to those obtained after the full training duration. By reducing computational demands without much loss in fidelity, this approach is a cost-effective strategy for enhancing modeling throughput in precision agriculture and botanical research. We substantiate the hypothesis that LPIPS can serve as a reliable surrogate for direct F1 score estimation in the context of NeRF training. The algorithm's ability to accurately predict optimal stopping points—balancing computational efficiency with reconstruction accuracy—presents a compelling case for its adoption in scenarios where resource conservation is paramount, yet quality cannot be entirely sacrificed.

\section{Discussion}
\label{Sec:Discussion}

In this section, we discuss the findings from our comparative analysis of NeRF models for 3D plant reconstruction. The results indicate that the Nerfacto model achieved the best performance, and we explore the theoretical basis for its superiority by examining the sampling strategies employed by the different models. Understanding these strategies provides insights into why Nerfacto outperformed the other models in terms of reconstruction quality. In our experiments, we found that the Nerfacto model produced the highest quality 3D reconstructions compared to Instant-NGP and other NeRF models. To understand the theoretical basis for Nerfacto's superior performance, in this section we take a deeper look at the sampling strategies used by Nerfacto and Instant-NGP and how they influence the visual quality and level of detail in the rendered scenes.

The divergent performance of the NeRF models necessitates a deeper examination of their underlying sampling strategies and their influence on the quality of 3D reconstruction. The difference in the output quality between Instant-NGP and Nerfacto, especially concerning the density and crispness of the rendered scenes, could indeed be related to the sampling strategies used by each algorithm.

\noindent\textbf{Instant-NGP Sampling Strategy:} Instant-NGP uses an improved training and rendering algorithm that involves a ray marching scheme with an occupancy grid. This means that when the algorithm shoots rays into the scene to sample colors and densities, it uses an occupancy grid to skip over empty space, as well as areas behind high-density regions to improve efficiency.

The occupancy grid used in Instant-NGP is a multiscale grid that coarsely marks empty and non-empty space and is used to determine where to skip samples to speed up processing. This approach is quite effective in terms of speed, leading to significant improvements over naive sampling methods. However, if the occupancy grid isn't fine-grained enough or if the method for updating this grid isn't capturing the scene's density variations accurately, it could lead to a ``muddy'' or overly dense rendering because it might not be sampling the necessary areas with enough precision.

\clearpage

\noindent\textbf{NeRFacto Sampling Strategy:} Nerfacto, on the other hand, uses a combination of different sampling techniques:

\begin{itemize}
\item \textit{Camera Pose Refinement:} By refining camera poses, Nerfacto ensures that the samples taken are based on more accurate viewpoints, which directly affects the clarity of the rendered images.
\item \textit{Piecewise Sampler:} This sampler is used to produce an initial set of samples, with a distribution that allows both dense sampling near the camera and appropriate sampling further away. This could lead to clearer images since it captures details both near and far from the camera.
\item \textit{Proposal Sampler:} This is a key part of the Nerfacto method. It uses a proposal network to concentrate sample locations in regions that contribute most to the final render, usually around the first surface intersection. This targeted sampling could be a major reason why Nerfacto produces crisper images—it focuses computational resources on the most visually significant parts of the scene.
\item \textit{Density Field:} By using a density field guided by a hash encoding and a small fused MLP, Nerfacto can efficiently guide sampling even further. It doesn't require an extremely detailed density map since it is used primarily for guiding the sampling process, which means that it balances quality and speed without necessarily impacting the final image's detail.
\end{itemize}

Instant-NGP's sampling strategy is built for speed, with an occupancy grid that helps skip irrelevant samples. This approach is great for real-time applications but can potentially miss subtle density variations, leading to a denser and less clear output if the grid isn't capturing all the necessary detail.
Nerfacto’s sampling strategy is more complex and layered, with multiple mechanisms in place to ensure that sampling is done more effectively in areas that greatly affect the visual output. The combination of pose refinement, piecewise sampling, proposal sampling, and an efficient density field leads to more accurate sampling, which in turn produces crisper images.
In summary, the reason for Nerfacto's better reconstruction likely stems from its more refined and targeted approach to sampling, which concentrates computational efforts on the most visually impactful parts of the scene. In contrast, Instant-NGP's faster but less targeted sampling may result in less clarity and more visual artifacts.

Finally, to retrieve the scale of the 3D reconstruction in the absence of reference point cloud data, a known scale can be placed on the ground during data collection. The exported point cloud can then be proportionally scaled based on this reference scale, which allows the size of the reconstructed plant to be calibrated to match its real-world dimensions. In order to show the practicality of this approach, we placed a 3D printed sphere of known diameter for the plant in Scenario I and captured the images. We then go through our pipeline of NeRF reconstruction, and instead of registering and scaling the scene to the ground truth LiDAR data, we scale it to the known sphere size. We then measured the height of the plant in this scenario. We find that by using this approach, the error in the height of the plant is within 1\%. We provide additional details of this experiment in the Supplement. We note that this is a preliminary result, and more detailed studies need to be performed in the future on extracting the correct scale from NeRF reconstructions.

\section{Conclusions}
\label{Sec:Conclusion}

The findings of this research underscore the value of NeRFs as a non-destructive approach for 3D plant reconstruction in precision agriculture. Our methodology more effectively facilitates critical agricultural tasks, such as growth monitoring, yield prediction, and early disease detection from accurate reconstruction of plant structures. Our comparative analysis, which benchmarks different NeRF models against ground truth data, highlights the method's efficiency, achieving a 74.65\% F1 score within 30 minutes of GPU training. Introducing an early stopping algorithm based on LPIPS further enhances this process, reducing training time by 61.1\% while limiting the average F1 score loss to just 7.4\%. 

Additionally, our work provides a comprehensive dataset and an evaluation framework, aiding the validation of current models and serving as a foundation for developing future NeRF applications in agriculture. The detailed insights into model performance across varied scenarios, coupled with the early stopping case study, offer practical guidance for 3D reconstruction using NeRFs. This research supports the advancement of non-intrusive agricultural technologies and also sets a baseline for future work at the intersection of NeRF technologies and agriculture, aiming to improve efficiency and accuracy in plant phenotyping and breeding.

\subsection*{Acknowledgements}
This work was supported in part by the Plant Science Institute at Iowa State University, the National Science Foundation under grant number OAC:1750865, and the National Institute of Food and Agriculture (USDA-NIFA) as part of the AI Institute for Resilient Agriculture (AIIRA), grant number 2021-67021-35329.

\subsection*{Supplementary Materials}
Figures S1 to S9.\\
Tables S1 to S5.

\subsection*{Author Contributions}
\textbf{Muhammad Arbab Arshad:} Methodology, Formal analysis, Software, Writing--Original Draft, Data Curation, Visualization.\\
\textbf{Talukder Jubery:}~Data Curation, Supervision, Writing--Review \& Editing.\\
\textbf{James Afful:}~Data Curation.\\
\textbf{Anushrut Jignasu:}~Data Curation.\\
\textbf{Aditya Balu:}~Methodology, Supervision, Writing--Review \& Editing.\\
\textbf{Baskar Ganapathysubramanian:}~Conceptualization, Methodology, Supervision, Project administration.\\
\textbf{Soumik Sarkar:}~Conceptualization, Methodology, Project administration, Funding acquisition.\\
\textbf{Adarsh Krishnamurthy:}~Conceptualization, Methodology, Supervision, Writing--Review \& Editing, Project administration, Funding acquisition.

\subsection*{Data Availability}
The data for the four Scenarios including raw images and the point cloud will be made available online. The GIT repository containing the different implementations of the NeRF code will also be made public. 

\bibliographystyle{unsrtnat}
\bibliography{Refs}

\end{document}


\maketitle

\renewcommand{\thepage}{S\arabic{page}}
\renewcommand{\thesection}{S\arabic{section}}
\renewcommand{\thetable}{S\arabic{table}}
\renewcommand{\thefigure}{S\arabic{figure}}
\setcounter{figure}{0}
\setcounter{table}{0}
\setcounter{page}{1}
\setcounter{section}{0}


This Supplement provides additional information on interpreting the visualizations and detailed metrics at different points of the training process for all scenarios. We also detail additional visualizations to support the validity of the proposed LPIPS-based early-stopping algorithms.

\section{Visualization of 3D Reconstruction Metrics}

\begin{figure*}[!h]
    \centering
    \begin{subfigure}{0.23\textwidth}
        \centering
        \includegraphics[width=0.99\linewidth]{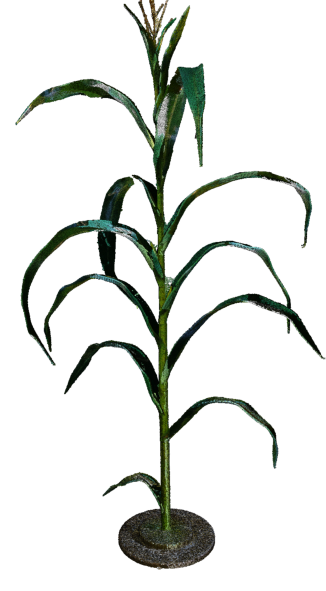}
        \caption{Ground Truth}
    \end{subfigure}
    \begin{subfigure}{0.23\textwidth}
        \centering
        \includegraphics[width=0.99\linewidth]{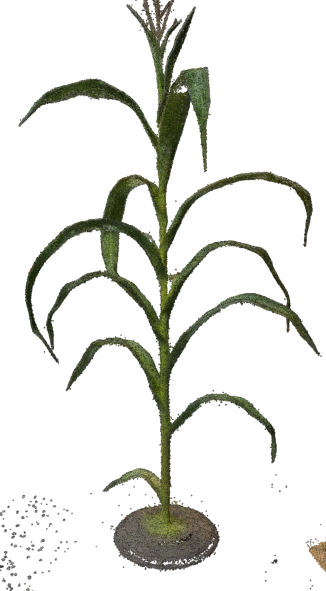}
        \caption{Reconstruction}
    \end{subfigure}    
    \begin{subfigure}{0.23\textwidth}
        \centering
        \includegraphics[width=0.99\linewidth]{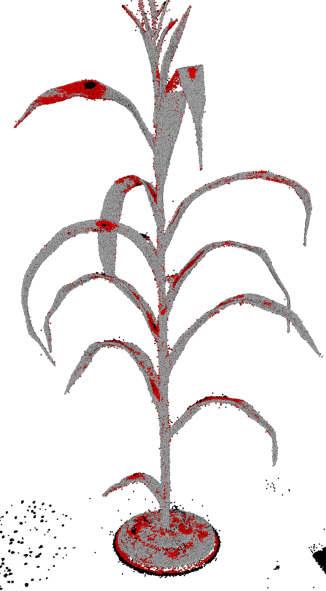}
        \caption{Precision}
    \end{subfigure}
    \begin{subfigure}{0.23\textwidth}
        \centering
        \includegraphics[width=0.99\linewidth]{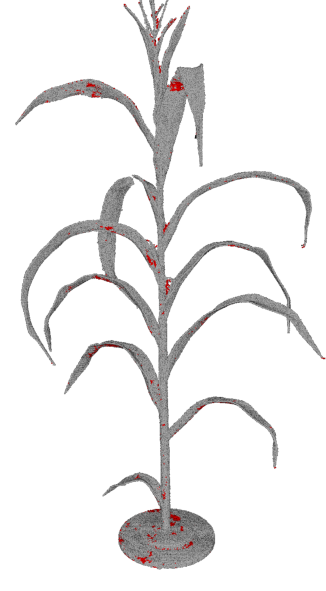}
        \caption{Recall}
    \end{subfigure}
   \caption{Point cloud 3D reconstruction metrics: (a) Original data; (b) Reconstruction; (c) Precision; (d) Recall. Legend: \textcolor{gray}{$\blacksquare$} indicates correct points, \textcolor{red}{$\blacksquare$} indicates missing points, and \textcolor{black}{$\blacksquare$} indicates outliers.}
    \label{fig:all_recision_3d}
\end{figure*}
 
 The interpretation of the colors in the precision and recall figures (see \figref{fig:all_recision_3d}) is as follows:
\begin{itemize}
\item \textbf{Grey}: (Correct) Represents points within a predefined distance threshold relative to the reference point cloud. This color indicates accurate points in precision and recall evaluations, where precision assesses the reconstruction against the ground truth, and recall evaluates the ground truth against the reconstruction.
\item \textbf{Red}: (Missing) Depicts points in the point cloud being tested that are beyond the distance threshold but within 3 standard deviations from the nearest point in the reference point cloud. These points are considered inaccuracies, showing missing details in the reconstruction when assessing precision and highlighting missing elements in the ground truth during recall analysis.
\item \textbf{Black}: (Outlier) Highlights points in the point cloud being tested that are more than 3 standard deviations away from any point in the reference point cloud. These points are extreme outliers and represent significant errors in the reconstruction relative to the ground truth for precision evaluations, and similarly significant discrepancies in the ground truth relative to the reconstruction for recall.
\end{itemize}

\begin{figure*}[!b]
    \centering
    \begin{subfigure}{0.3\textwidth}
        \centering
        \includegraphics[width=0.5\linewidth]{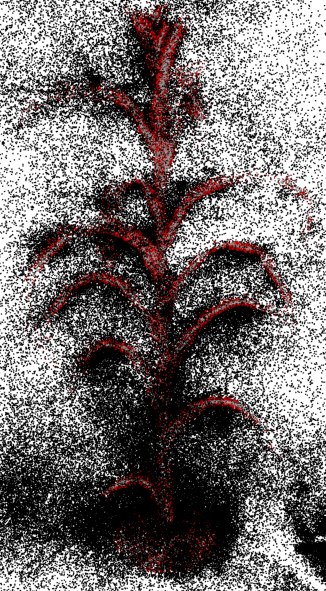}
        \caption{200 Iterations}
    \end{subfigure}
    \begin{subfigure}{0.3\textwidth}
        \centering
        \includegraphics[width=0.5\linewidth]{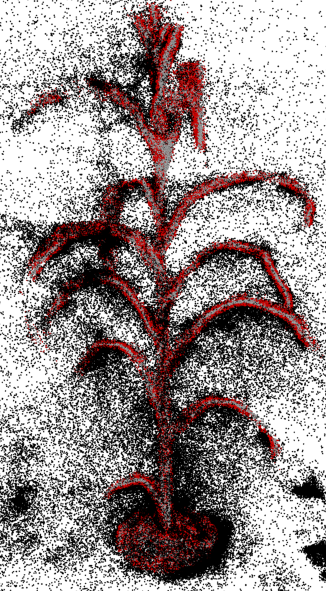}
        \caption{400 Iterations}
    \end{subfigure}
    \begin{subfigure}{0.3\textwidth}
        \centering
        \includegraphics[width=0.5\linewidth]{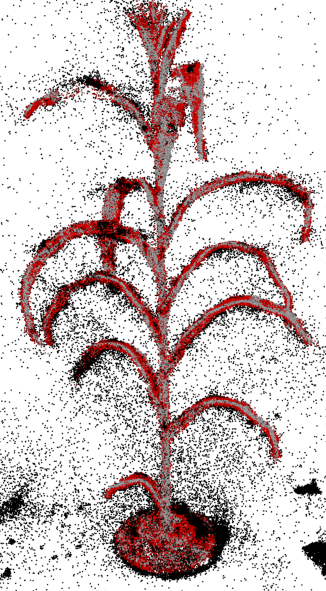}
        \caption{800 Iterations}
    \end{subfigure}\\
    \vspace{1em}
    \begin{subfigure}{0.3\textwidth}
        \centering
        \includegraphics[width=0.5\linewidth]{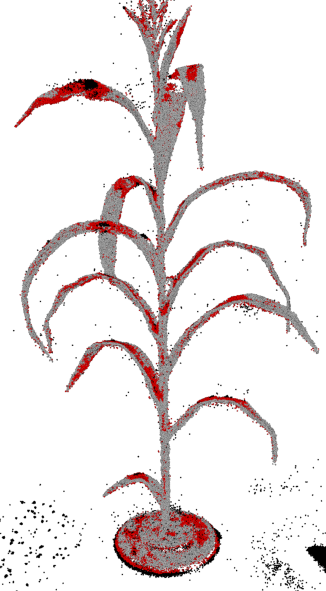}
        \caption{5000 Iterations}
    \end{subfigure}
    \begin{subfigure}{0.3\textwidth}
        \centering
        \includegraphics[width=0.5\linewidth]{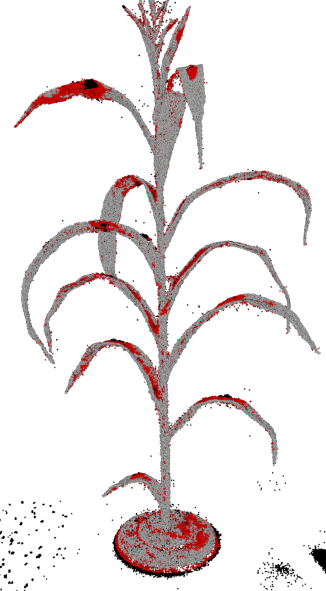}
        \caption{10000 Iterations}
    \end{subfigure}
    \begin{subfigure}{0.3\textwidth}
        \centering
        \includegraphics[width=0.5\linewidth]{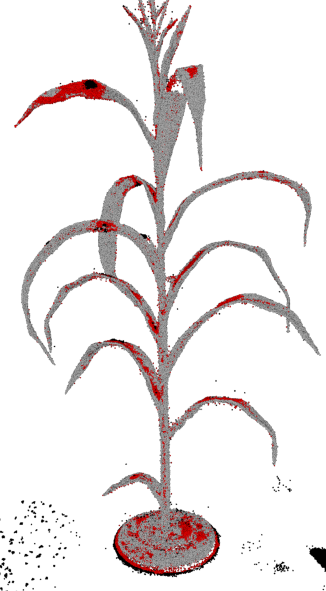}
        \caption{30000 Iterations}
    \end{subfigure}
    \caption{Precision change over iterations for NeRFacto - Scenario I. Legend: $\filledsquare{gray}$~Correct, $\filledsquare{red}$~Missing, $\filledsquare{black}$~Outlier. }
    \label{fig:evolution_of_training_scenario_1}
\end{figure*}

\section{Additional Performance Metric Results}

Granular explanation of 2D metrics (PSNR, SSIM and LPIPS), and their evolution over the training process for each scenario is given below.

\subsection{Scenario I}

\begin{figure*}[!t]
    \centering
    \begin{subfigure}{0.32\textwidth}
        \centering
        \includegraphics[width=\linewidth]{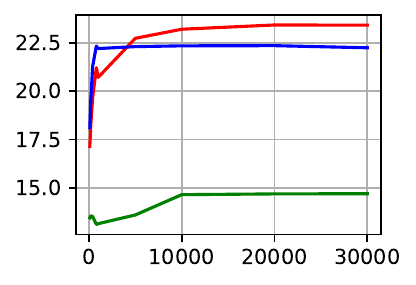}
        \raisebox{0.1in}{\small{Iterations}}
        \caption{PSNR}
    \end{subfigure}
    \hfill
    \begin{subfigure}{0.32\textwidth}
        \centering
        \includegraphics[width=\linewidth]{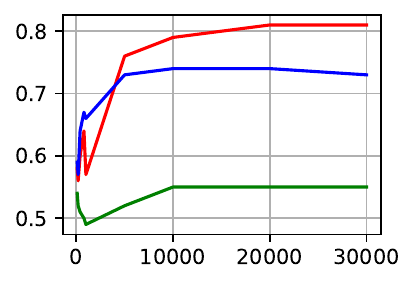}
        \raisebox{0.1in}{\small{Iterations}}
        \caption{SSIM}
    \end{subfigure}
    \hfill
    \begin{subfigure}{0.32\textwidth}
        \centering
        \includegraphics[width=\linewidth]{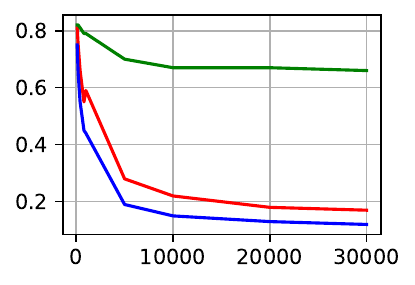}
        \raisebox{0.1in}{\small{Iterations}}
        \caption{LPIPS}
    \end{subfigure}\\
    \begin{subfigure}{0.32\textwidth}
        \centering
        \includegraphics[width=\linewidth]{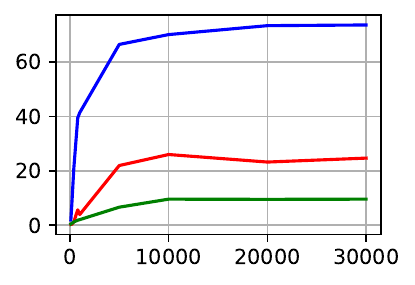}
        \raisebox{0.1in}{\small{Iterations}}
        \caption{Precision}
    \end{subfigure}
    \hfill
    \begin{subfigure}{0.32\textwidth}
        \centering
        \includegraphics[width=\linewidth]{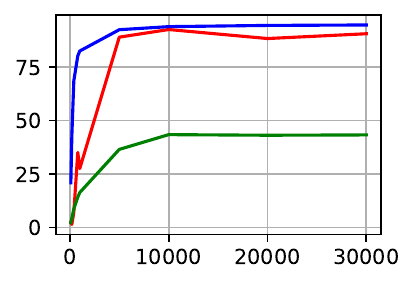}
        \raisebox{0.1in}{\small{Iterations}}
        \caption{Recall}
    \end{subfigure}
    \hfill
    \begin{subfigure}{0.32\textwidth}
        \centering
        \includegraphics[width=\linewidth]{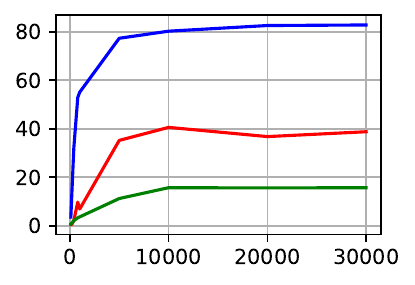}
        \raisebox{0.1in}{\small{Iterations}}
        \caption{F1 Score}
    \end{subfigure}\\
    \raisebox{0.5ex}{\textcolor{red}{\rule{1cm}{1pt}}} Instant-NGP \quad
    \raisebox{0.5ex}{\textcolor{blue}{\rule{1cm}{1pt}}} NeRFacto \quad
    \raisebox{0.5ex}{\textcolor{green}{\rule{1cm}{1pt}}} TensoRF
    \caption{Comparison of 2D quality (top) and 3D geometry (bottom) metrics for Scenario-I.}
    \label{fig:combined_iterations_score_scenario_1}
\end{figure*}

The trend of all metrics throughout the training process is depicted in \figref{fig:combined_iterations_score_scenario_1}.

\noindent\textbf{Precision:} Instant-NGP shows a significant leap in precision from 100 to 5000 iterations (0.29 to 21.93), indicating a drastic improvement in the accuracy of reconstructed points relative to the ground truth. NeRFacto demonstrates a more consistent and steep rise in precision, reaching a peak of 73.57 at 30000 iterations, which surpasses Instant-NGP's best precision. TensoRF, however, shows a relatively modest increase in precision, suggesting its limited capability in accurately capturing fine details compared to the other two models. Visuals of precision at different points of the training is shown in \figref{fig:evolution_of_training_scenario_1}.

\noindent\textbf{PSNR:} The Peak Signal-to-Noise Ratio (PSNR) reflects the quality of rendered images. In this metric, Instant-NGP and NeRFacto show a gradual increase in PSNR with more iterations, suggesting improved image quality. TensoRF's PSNR values are lower, indicating potentially lower image quality throughout its iterations.

\noindent\textbf{SSIM:} The Structural Similarity Index (SSIM) is another measure of image quality, assessing the perceived change in structural information. Here, NeRFacto and Instant-NGP both show a steady increase in SSIM with more iterations, with NeRFacto achieving slightly higher scores, suggesting better preservation of structural information in its renderings. TensoRF, again, shows relatively lower SSIM scores.

\noindent\textbf{LPIPS:} The Lower Perceptual Image Patch Similarity (LPIPS) metric indicates perceived image similarity, with lower values being better. NeRFacto and Instant-NGP both show a significant decrease in LPIPS with more iterations, indicating improved perceptual similarity to the ground truth. TensoRF's LPIPS values are consistently higher, suggesting lower perceptual similarity.

\begin{table*}[!t]
    \centering
    \caption{Detailed performance metrics of NeRFs reconstruction techniques - Scenario I}
    \begin{filecontents*}{Data/scenario1.csv}
Model Name,Number of Iterations,Precision,Recall,F1 Score,PSNR,SSIM,LPIPS,Time Taken (s)
Instant-NGP,100,0.29,2.48,0.53,17.14,0.58,0.81,13
,200,0.32,1.50,0.53,18.23,0.56,0.75,26
,400,1.25,7.14,2.13,19.64,0.60,0.66,42
,800,5.66,35.17,9.74,21.21,0.64,0.55,61
,1000,3.93,27.61,6.88,20.71,0.57,0.59,76
,5000,21.93,89.03,35.20,22.73,0.76,0.28,175
,10000,25.98,92.59,40.57,23.20,0.79,0.22,297
,20000,23.21,88.38,36.77,23.42,0.81,0.18,527
,30000,24.66,90.62,38.77,23.41,0.81,0.17,756
,,,,,,,,
TensoRF,100,0.43,2.27,0.72,13.44,0.54,0.82,13
,200,0.65,4.47,1.13,13.55,0.52,0.82,25
,400,1.18,8.76,2.07,13.51,0.51,0.81,40
,800,1.86,14.22,3.29,13.10,0.50,0.79,61
,1000,2.05,16.40,3.65,13.14,0.49,0.79,77
,5000,6.63,36.57,11.22,13.59,0.52,0.70,420
,10000,9.58,43.47,15.69,14.64,0.55,0.67,859
,20000,9.51,43.19,15.59,14.68,0.55,0.67,1651
,30000,9.58,43.34,15.69,14.69,0.55,0.66,1973
,,,,,,,,
NeRFacto,100,1.94,20.98,3.55,18.11,0.59,0.75,14
,200,7.72,42.84,13.08,19.50,0.57,0.65,27
,400,20.86,68.64,32.00,21.27,0.64,0.55,43
,800,39.48,80.26,52.92,22.33,0.67,0.45,64
,1000,41.35,82.54,55.09,22.20,0.66,0.44,79
,5000,66.43,92.51,77.33,22.30,0.73,0.19,430
,10000,70.04,93.94,80.25,22.34,0.74,0.15,564
,20000,73.32,94.51,82.58,22.35,0.74,0.13,1068
,30000,73.57,94.72,82.81,22.24,0.73,0.12,1938
\end{filecontents*}
\pgfplotstabletypeset[
        col sep=comma,
        string type,
        every head row/.style={
            before row={\toprule},
            after row={\midrule}
        },
        every last row/.style={after row={\bottomrule}},
        columns/Model Name/.style={column name=Model Name, column type=l},
        columns/Number of Iterations/.style={column name=Iters , column type=r},
        columns/Precision/.style={column name=Precision $\uparrow$, column type=r},
        columns/Recall/.style={column name=Recall $\uparrow$, column type=r},
        columns/F1 Score/.style={column name=F1 $\uparrow$, column type=r},
        columns/PSNR/.style={column name=PSNR $\uparrow$, column type=r},
        columns/SSIM/.style={column name=SSIM $\uparrow$, column type=r},
        columns/LPIPS/.style={column name=LPIPS $\downarrow$, column type=r},
        columns/Time Taken (s)/.style={column name=T (s) $\downarrow$, column type=r}
    ]{Data/scenario1.csv}
    \label{fig:table_scenario1}
\end{table*}

\clearpage

\subsection{Scenario II}
\begin{figure*}[!b]
    \centering
    \begin{subfigure}{0.32\textwidth}
        \centering
        \includegraphics[width=0.95\linewidth]{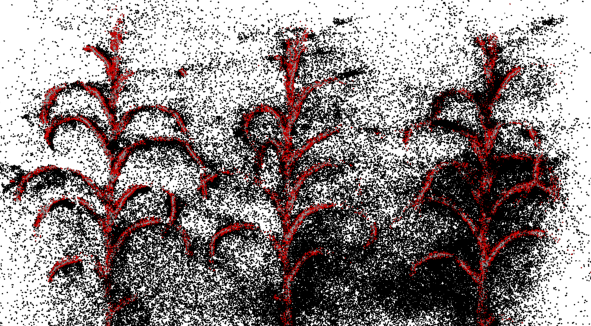}
        \caption{200 Iterations}
    \end{subfigure}
    \begin{subfigure}{0.32\textwidth}
        \centering
        \includegraphics[width=0.95\linewidth]{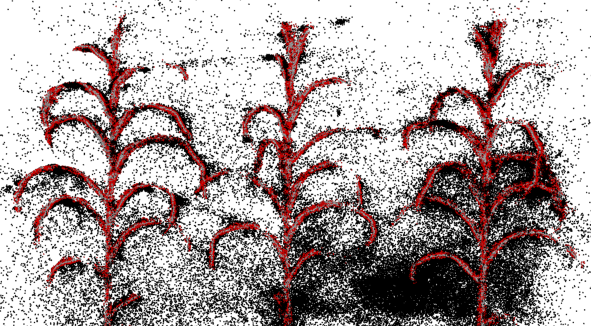}
        \caption{400 Iterations}
    \end{subfigure}
    \begin{subfigure}{0.32\textwidth}
        \centering
        \includegraphics[width=0.95\linewidth]{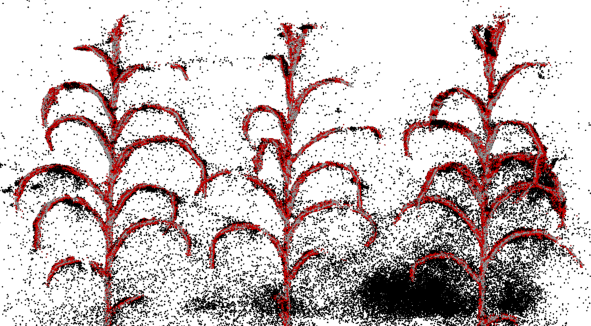}
        \caption{800 Iterations}
    \end{subfigure}\\
    \vspace{3em}
    \begin{subfigure}{0.32\textwidth}
        \centering
        \includegraphics[width=0.95\linewidth]{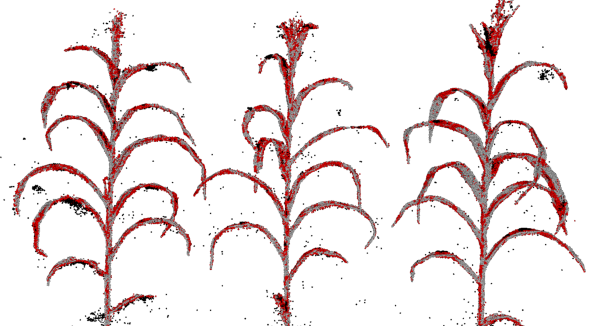}
        \caption{5000 Iterations}
    \end{subfigure}
    \begin{subfigure}{0.32\textwidth}
        \centering
        \includegraphics[width=0.95\linewidth]{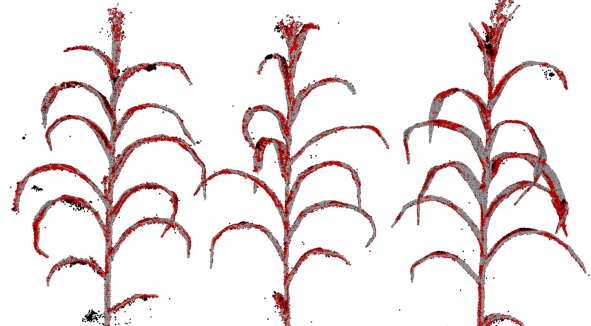}
        \caption{10000 Iterations}
    \end{subfigure}
    \begin{subfigure}{0.32\textwidth}
        \centering
        \includegraphics[width=0.95\linewidth]{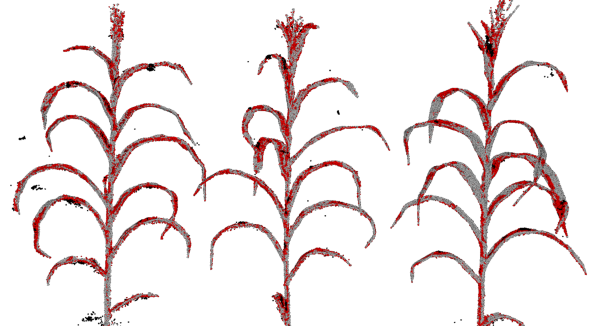}
        \caption{30000 Iterations}
    \end{subfigure}
    \caption{Precision change as a function of different training lengths for NeRFacto - Scenario II. Legend: $\filledsquare{gray}$~Correct, $\filledsquare{red}$~Missing, $\filledsquare{black}$~Outlier.}
    \label{fig:evolution_of_training_scenario_2}
\end{figure*}

\begin{figure*}[!t]
    \centering
    \begin{subfigure}{0.32\textwidth}
        \centering
        \includegraphics[width=\linewidth]{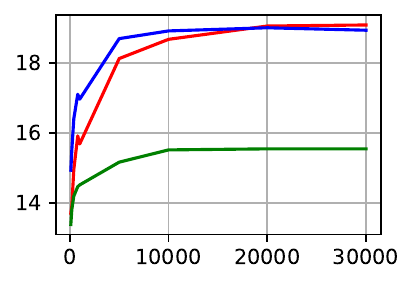}
        \raisebox{0.1in}{\small{Iterations}}
        \caption{PSNR}
    \end{subfigure}
    \hfill
    \begin{subfigure}{0.32\textwidth}
        \centering   
        \includegraphics[width=\linewidth]{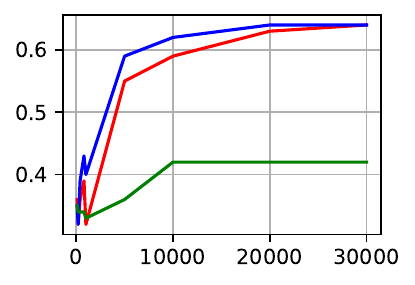}
        \raisebox{0.1in}{\small{Iterations}}
        \caption{SSIM}
    \end{subfigure}
    \hfill
    \begin{subfigure}{0.32\textwidth}
        \centering           
        \includegraphics[width=\linewidth]{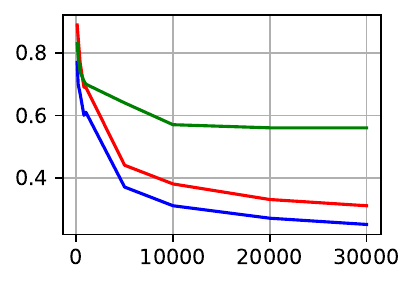}
        \raisebox{0.1in}{\small{Iterations}}
        \caption{LPIPS}
    \end{subfigure}
    \begin{subfigure}{0.32\textwidth}
        \centering
        \includegraphics[width=\linewidth]{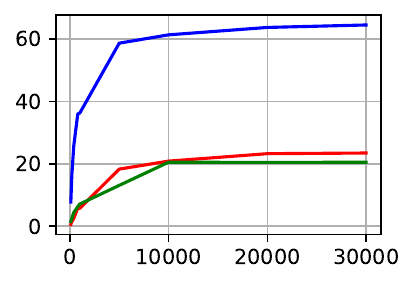}
        \raisebox{0.1in}{\small{Iterations}}
        \caption{Precision}
    \end{subfigure}
    \hfill
    \begin{subfigure}{0.32\textwidth}
        \centering
        
        \includegraphics[width=\linewidth]{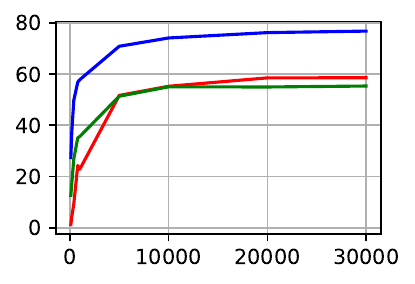}
        \raisebox{0.1in}{\small{Iterations}}
        \caption{Recall}
    \end{subfigure}
    \hfill
    \begin{subfigure}{0.32\textwidth}
        \centering
        \includegraphics[width=\linewidth]{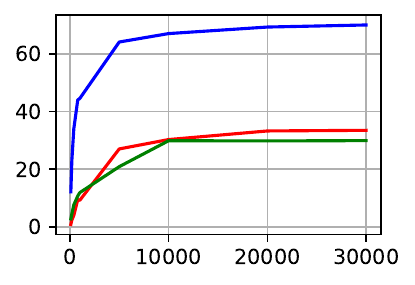}
        \raisebox{0.1in}{\small{Iterations}}
         \caption{F1 Score}
    \end{subfigure}\\
    \vspace{1ex}
    \raisebox{0.5ex}{\textcolor{red}{\rule{1cm}{1pt}}} Instant-NGP \quad
    \raisebox{0.5ex}{\textcolor{blue}{\rule{1cm}{1pt}}} NeRFacto \quad
    \raisebox{0.5ex}{\textcolor{green}{\rule{1cm}{1pt}}} TensoRF
    \caption{Comparison of 2D quality (top) and 3D geometry (bottom) metrics for Scenario-II}
    \label{fig:combined_iterations_score_scenario_2}
\end{figure*}

\figref{fig:evolution_of_training_scenario_2} illustrates the change of precision over the course of the training iterations. The trend of all metrics throughout the training process is depicted in \figref{fig:combined_iterations_score_scenario_2}. For the same scenario, we show that  Mip-NeRF fails to produce a reasonable reconstruction (\tabref{tab:table_scenario_2_mipnerf}).

\noindent\textbf{PSNR:} In terms of PSNR, which evaluates the quality of rendered images, all models show improvement with more iterations. Instant-NGP goes from 13.70 to 19.08, NeRFacto from 14.93 to 18.93, and TensoRF from 13.38 to 15.54. Instant-NGP achieves the highest PSNR, suggesting better image quality.

\noindent\textbf{SSIM:}  For SSIM, higher values indicate better image structure similarity. Instant-NGP progresses from 0.36 to 0.64, NeRFacto from 0.35 to 0.64, and TensoRF from 0.35 to 0.42. Both Instant-NGP and NeRFacto perform similarly and better than TensoRF in this aspect.

\noindent\textbf{LPIPS:} Lower LPIPS values signify higher perceptual similarity to the ground truth. Instant-NGP decreases from 0.89 to 0.31, NeRFacto from 0.77 to 0.25, and TensoRF from 0.83 to 0.56, with NeRFacto showing the best perceptual image quality.

\begin{table*}[!b]
  \centering
  \caption{Performance metrics of MipNeRF reconstruction - Scenario II (failed).}
  \begin{filecontents*}{Data/scenario2-mipnerf.csv}
Number of Iterations,Precision,Recall,F1 Score,PSNR,SSIM,LPIPS,Time Taken (s)
100,0.47,7.57,0.89,11.01,0.34,0.91,26
200,0.63,10.67,1.19,11.24,0.34,0.93,52
400,0.69,12.1,1.3,11.07,0.34,0.92,86
800,0.48,7.68,0.91,9.88,0.33,0.94,158
1000,0.44,7.19,0.82,9.65,0.32,0.94,192
5000,0.35,5.41,0.66,9.31,0.31,0.94,687
10000,0.42,6.29,0.79,8.88,0.29,0.9,1303
20000,0.34,5.3,0.65,8.78,0.28,0.89,2508
30000,0.35,5.71,0.67,9,0.27,0.88,3856
\end{filecontents*}
\pgfplotstabletypeset[
    col sep=comma,
    string type,
    every head row/.style={
      before row={\toprule},
      after row={\midrule}
    },
    every last row/.style={after row={\bottomrule}},
    columns/Number of Iterations/.style={column name=Iters , column type=r},
    columns/Precision/.style={column name=Precision $\uparrow$, column type=r},
    columns/Recall/.style={column name=Recall $\uparrow$, column type=r},
    columns/F1 Score/.style={column name=F1 $\uparrow$, column type=r},
    columns/PSNR/.style={column name=PSNR $\uparrow$, column type=r},
    columns/SSIM/.style={column name=SSIM $\uparrow$, column type=r},
    columns/LPIPS/.style={column name=LPIPS $\downarrow$, column type=r},
    columns/Time Taken (s)/.style={column name=T (s) $\downarrow$, column type=r}
  ]{Data/scenario2-mipnerf.csv}
  \label{tab:table_scenario_2_mipnerf}
\end{table*}

\begin{table*}[!t]
    \centering
    \caption{Detailed performance metrics of NeRFs reconstruction techniques - Scenario II}
    \begin{filecontents*}{Data/scenario2.csv}
Model Name,Number of Iterations,Precision,Recall,F1 Score,PSNR,SSIM,LPIPS,Time Taken (s)
Instant-NGP,100,0.57,1.23,0.78,13.70,0.36,0.89,13
,200,1.52,4.41,2.27,13.91,0.32,0.85,27
,400,2.49,9.35,3.93,14.95,0.36,0.77,43
,800,5.83,24.29,9.41,15.91,0.39,0.69,63
,1000,5.72,22.80,9.14,15.68,0.32,0.69,78
,5000,18.30,51.68,27.02,18.12,0.55,0.44,176
,10000,20.86,55.25,30.28,18.67,0.59,0.38,296
,20000,23.24,58.53,33.27,19.05,0.63,0.33,1023
,30000,23.45,58.57,33.49,19.08,0.64,0.31,1886
,,,,,,,,
TensoRF,100,1.51,12.62,2.70,13.38,0.35,0.83,14
,200,2.53,17.53,4.43,13.81,0.34,0.79,27
,400,4.41,26.94,7.58,14.19,0.34,0.74,44
,800,6.28,35.10,10.66,14.46,0.34,0.71,66
,1000,7.09,35.61,11.82,14.51,0.33,0.70,82
,5000,13.09,51.34,20.86,15.16,0.36,0.64,399
,10000,20.51,55.03,29.88,15.51,0.42,0.57,840
,20000,20.44,54.97,29.80,15.54,0.42,0.56,1709
,30000,20.50,55.34,29.91,15.54,0.42,0.56,2607
,,,,,,,,
NeRFacto,100,7.77,27.44,12.12,14.93,0.35,0.77,15
,200,16.42,37.77,22.89,15.51,0.32,0.70,28
,400,25.83,49.81,34.02,16.40,0.39,0.67,44
,800,35.97,56.86,44.07,17.10,0.43,0.60,64
,1000,36.22,57.80,44.53,16.96,0.40,0.61,79
,5000,58.64,70.87,64.18,18.69,0.59,0.37,404
,10000,61.31,74.12,67.11,18.91,0.62,0.31,749
,20000,63.68,76.21,69.38,19.00,0.64,0.27,988
,30000,64.47,76.80,70.10,18.93,0.64,0.25,1226
\end{filecontents*}
\pgfplotstabletypeset[
        col sep=comma,
        string type,
        every head row/.style={
            before row={\toprule},
            after row={\midrule}
        },
        every last row/.style={after row={\bottomrule}},
        columns/Model Name/.style={column name=Model Name, column type=l},
        columns/Number of Iterations/.style={column name=Iters , column type=r},
        columns/Precision/.style={column name=Precision $\uparrow$, column type=r},
        columns/Recall/.style={column name=Recall $\uparrow$, column type=r},
        columns/F1 Score/.style={column name=F1 $\uparrow$, column type=r},
        columns/PSNR/.style={column name=PSNR $\uparrow$, column type=r},
        columns/SSIM/.style={column name=SSIM $\uparrow$, column type=r},
        columns/LPIPS/.style={column name=LPIPS $\downarrow$, column type=r},
        columns/Time Taken (s)/.style={column name=T (s) $\downarrow$, column type=r}
    ]{Data/scenario2.csv}
    \label{fig:table_scenario2}
\end{table*}

\clearpage

\subsection{Scenario III}

\begin{figure*}[!h]
    \centering
    \begin{minipage}[c]{0.03\textwidth}
    \end{minipage}%
    \begin{minipage}[c]{0.96\textwidth}
        \begin{subfigure}{0.3\textwidth}
            \centering
            \includegraphics[width=\linewidth]{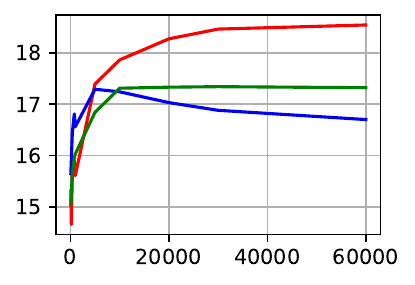}
            \raisebox{0.1in}{\small{Iterations}}
            \caption{PSNR}
        \end{subfigure}
        \hfill
        \begin{subfigure}{0.3\textwidth}
            \centering
            \includegraphics[width=\linewidth]{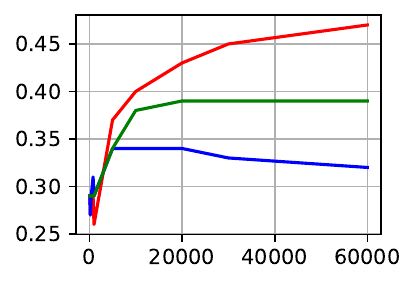}
            \raisebox{0.1in}{\small{Iterations}}
            \caption{SSIM}
        \end{subfigure}
        \hfill
        \begin{subfigure}{0.3\textwidth}
            \centering
            \includegraphics[width=\linewidth]{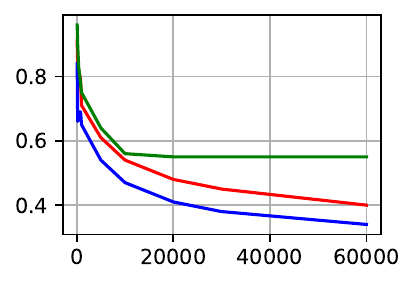}
            \raisebox{0.1in}{\small{Iterations}}
            \caption{LPIPS}
        \end{subfigure}
    \end{minipage}
    \begin{minipage}[c]{0.03\textwidth}
    \end{minipage}%
    \begin{minipage}[c]{0.96\textwidth}
        \begin{subfigure}{0.3\textwidth}
            \centering
            \includegraphics[width=\linewidth]{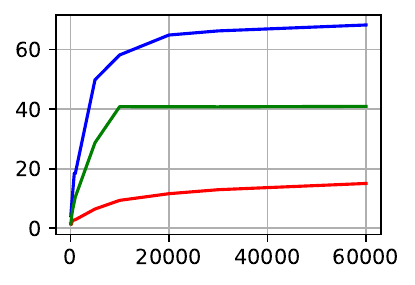}
            \raisebox{0.1in}{\small{Iterations}}
            \caption{Precision}
        \end{subfigure}
        \hfill
        \begin{subfigure}{0.3\textwidth}
            \centering
            \includegraphics[width=\linewidth]{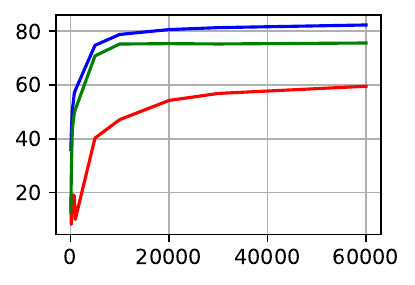}
            \raisebox{0.1in}{\small{Iterations}}
            \caption{Recall}
        \end{subfigure}
        \hfill
        \begin{subfigure}{0.3\textwidth}
            \centering
            \includegraphics[width=\linewidth]{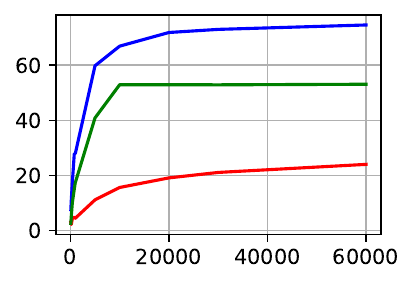}
            \raisebox{0.1in}{\small{Iterations}}
            \caption{F1 Score}
        \end{subfigure}
    \end{minipage}
    \vspace{1ex}
    \vspace{1ex}
    \raisebox{0.5ex}{\textcolor{red}{\rule{1cm}{1pt}}} Instant-NGP \quad
    \raisebox{0.5ex}{\textcolor{blue}{\rule{1cm}{1pt}}} NeRFacto \quad
    \raisebox{0.5ex}{\textcolor{green}{\rule{1cm}{1pt}}} TensoRF
    \caption{Comparison of 2D quality (top) and 3D geometry (bottom) metrics for Scenario-III.}
    \label{fig:combined_iterations_score_scenario_3}
\end{figure*}

The trend of all metrics throughout the training process is depicted in \figref{fig:combined_iterations_score_scenario_3}.

\noindent\textbf{PSNR:} In terms of image quality, as measured by PSNR, all models show incremental improvements with more iterations. Instant-NGP and NeRFacto display similar trends, with NeRFacto slightly leading, peaking at 16.70 at 60000 iterations. TensoRF shows a comparable maximum PSNR of 17.32, suggesting its slight edge in rendering higher-quality images.

\noindent\textbf{SSIM:} For the SSIM metric, all three models show improvements with increased iterations. NeRFacto maintains a slight advantage over the others, peaking at 0.32 at 60000 iterations, indicating its better performance in maintaining structural integrity in the rendered images. Instant-NGP and TensoRF show similar SSIM scores, with TensoRF slightly leading at higher iterations.

\noindent\textbf{LPIPS:} The LPIPS scores, which assess perceptual similarity, decrease for all models with more iterations, indicating improved performance. NeRFacto and TensoRF show similar trends, with NeRFacto having a slight edge, achieving a score of 0.34 at 60000 iterations compared to TensoRF's 0.55. Instant-NGP's performance is consistently lower in this metric.

\begin{table*}[!t]
    \centering
    \caption{Detailed performance metrics of NeRFs reconstruction techniques - Scenario III}
    \begin{filecontents*}{Data/scenario3.csv}
Model Name,Number of Iterations,Precision,Recall,F1 Score,PSNR,SSIM,LPIPS,Time Taken (s)
Instant-NGP,100,1.61,17.61,2.95,15.31,0.29,0.91,16
,200,1.29,8.19,2.23,14.66,0.27,0.84,30
,400,2.45,19.22,4.34,15.65,0.29,0.82,46
,800,2.81,18.90,4.89,15.91,0.29,0.79,67
,1000,2.88,10.08,4.48,15.61,0.26,0.71,82
,5000,6.47,40.28,11.15,17.39,0.37,0.61,186
,10000,9.38,47.12,15.65,17.86,0.40,0.54,311
,20000,11.62,54.26,19.13,18.27,0.43,0.48,548
,30000,12.96,56.88,21.11,18.46,0.45,0.45,783
,60000,15.06,59.55,24.04,18.54,0.47,0.40,1466
,,,,,,,,
TensoRF,100,1.48,12.65,2.65,15.05,0.29,0.96,17
,200,3.05,27.40,5.48,15.26,0.29,0.90,31
,400,5.69,43.45,10.06,15.63,0.29,0.83,47
,800,8.92,49.59,15.13,15.96,0.29,0.78,68
,1000,10.48,50.82,17.37,16.03,0.29,0.75,84
,5000,28.74,70.83,40.89,16.84,0.34,0.64,208
,10000,40.85,75.24,52.95,17.31,0.38,0.56,374
,20000,40.82,75.38,52.96,17.33,0.39,0.55,697
,30000,40.80,75.26,52.92,17.34,0.39,0.55,1018
,60000,40.95,75.62,53.13,17.32,0.39,0.55,1965
,,,,,,,,
NeRFacto,100,4.26,35.99,7.61,15.65,0.29,0.84,13
,200,6.53,42.49,11.31,15.91,0.27,0.66,27
,400,10.48,50.45,17.35,16.44,0.29,0.69,43
,800,18.49,57.01,27.93,16.81,0.31,0.69,64
,1000,18.39,58.02,27.93,16.56,0.29,0.65,80
,5000,49.87,74.75,59.83,17.29,0.34,0.54,189
,10000,58.22,78.80,66.96,17.24,0.34,0.47,318
,20000,64.91,80.61,71.92,17.03,0.34,0.41,561
,30000,66.30,81.33,73.05,16.88,0.33,0.38,803
,60000,68.29,82.32,74.65,16.70,0.32,0.34,1499
\end{filecontents*}
\pgfplotstabletypeset[
        col sep=comma,
        string type,
        every head row/.style={
            before row={\toprule},
            after row={\midrule}
        },
        every last row/.style={after row={\bottomrule}},
        columns/Model Name/.style={column name=Model Name, column type=l},
        columns/Number of Iterations/.style={column name=Iters , column type=r},
        columns/Precision/.style={column name=Precision $\uparrow$, column type=r},
        columns/Recall/.style={column name=Recall $\uparrow$, column type=r},
        columns/F1 Score/.style={column name=F1 $\uparrow$, column type=r},
        columns/PSNR/.style={column name=PSNR $\uparrow$, column type=r},
        columns/SSIM/.style={column name=SSIM $\uparrow$, column type=r},
        columns/LPIPS/.style={column name=LPIPS $\downarrow$, column type=r},
        columns/Time Taken (s)/.style={column name=T (s) $\downarrow$, column type=r}
    ]{Data/scenario3.csv}
    \label{fig:table_scenario3}
\end{table*}

\begin{figure*}[!t]
    \centering
    \begin{subfigure}{0.3\linewidth}
        \centering
        \includegraphics[width=\linewidth]{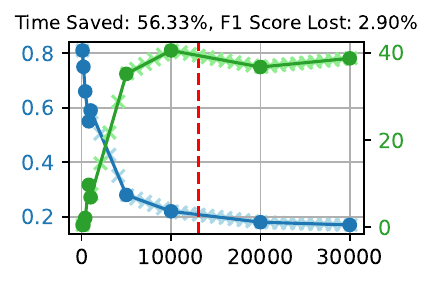}
        \raisebox{0.1in}{\small{Iterations}}
        \caption{Instant-NGP (Scene-I)}
    \end{subfigure}
    \hfill
    \begin{subfigure}{0.3\linewidth}
        \centering
        \includegraphics[width=\linewidth]{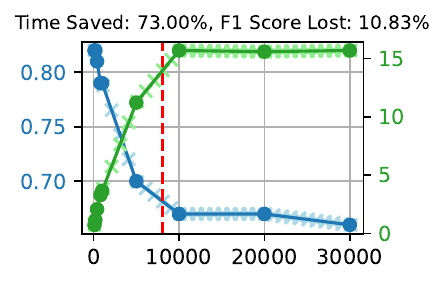}
        \raisebox{0.1in}{\small{Iterations}}
        \caption{TensoRF (Scene-I)}
    \end{subfigure}
    \hfill
    \begin{subfigure}{0.3\linewidth}
        \centering
        \includegraphics[width=\linewidth]{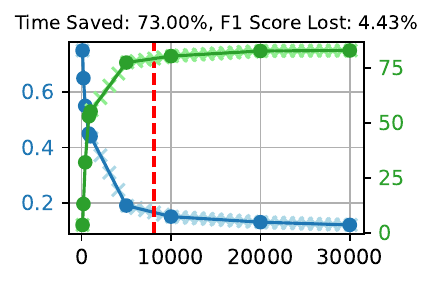}
        \raisebox{0.1in}{\small{Iterations}}
        \caption{NeRFacto (Scene-I)}
    \end{subfigure}
    \begin{subfigure}{0.3\linewidth}
        \centering
        \includegraphics[width=\linewidth]{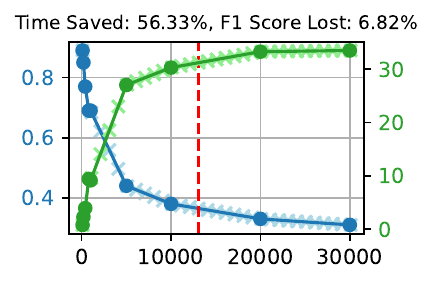}
        \raisebox{0.1in}{\small{Iterations}}
        \caption{Instant-NGP (Scene-II)}
    \end{subfigure}
    \hfill
    \begin{subfigure}{0.3\linewidth}
        \centering
        \includegraphics[width=\linewidth]{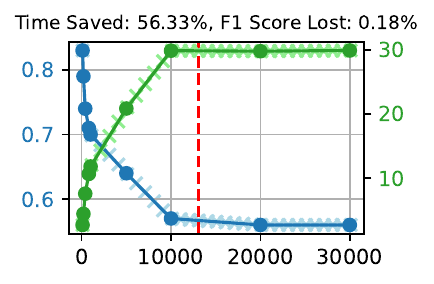}
        \raisebox{0.1in}{\small{Iterations}}
        \caption{TensoRF (Scene-II)}
    \end{subfigure}
    \hfill
    \begin{subfigure}{0.3\linewidth}
        \centering
        \includegraphics[width=\linewidth]{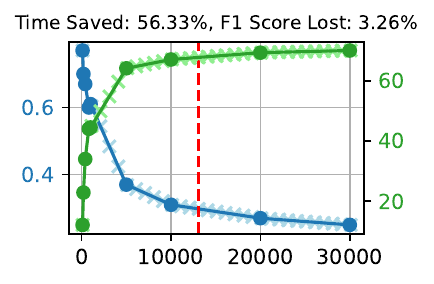}
        \raisebox{0.1in}{\small{Iterations}}
        \caption{NeRFacto (Scene-II)}
    \end{subfigure}
    \begin{subfigure}{0.3\linewidth}
        \centering
        \includegraphics[width=\linewidth]{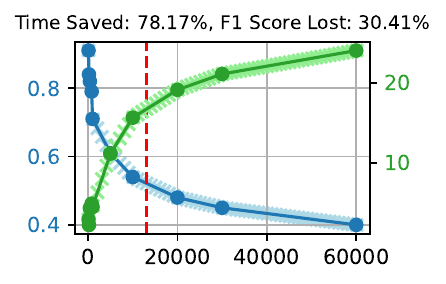}
        \raisebox{0.1in}{\small{Iterations}}
        \caption{Instant-NGP (Scene-III)}
    \end{subfigure}
    \hfill
    \begin{subfigure}{0.3\linewidth}
        \centering
        \includegraphics[width=\linewidth]{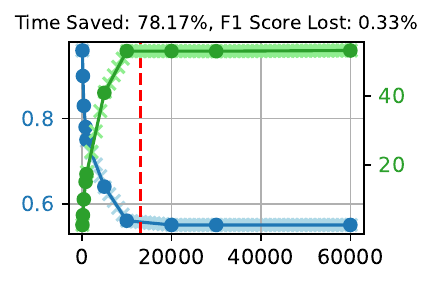}
        \raisebox{0.1in}{\small{Iterations}}
        \caption{TensoRF (Scene-III)}
    \end{subfigure}
    \hfill
    \begin{subfigure}{0.3\linewidth}
        \centering
        \includegraphics[width=\linewidth]{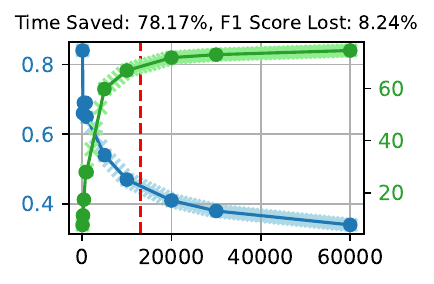}
        \raisebox{0.1in}{\small{Iterations}}
        \caption{NeRFacto (Scene-III)}
    \end{subfigure}\\
    \raisebox{0ex}{\textcolor{red}{\rule[0.5ex]{0.1cm}{1pt}\hspace{0.1cm}\rule[0.5ex]{0.1cm}{1pt}\hspace{0.1cm}\rule[0.5ex]{0.1cm}{1pt}}}~Detected Plateau Point \quad
    \raisebox{0.5ex}{\textcolor{blue}{\rule{0.5cm}{1pt}}} LPIPS \quad
    \raisebox{0.5ex}{\textcolor{green}{\rule{0.5cm}{1pt}}} F1 Score \quad
    \textcolor{black}{\textbullet} Original \quad
    \textcolor{black}{$\times$} Interpolated \quad
    \caption{Performance of early stopping algorithm based on LPIPS on scenes with ground truth.}
    \label{fig:lpips_vs_f1}
\end{figure*}

\noindent Detailed metrics for all the three scenarios are given in the tables: \tabref{fig:table_scenario1}, \tabref{fig:table_scenario2}, and \tabref{fig:table_scenario3}.

\subsection{Validation of Early-Stopping Algorithm}
\begin{figure*}[!t]
    \centering
    \begin{subfigure}{0.19\linewidth}
        \centering
        \includegraphics[width=\linewidth]{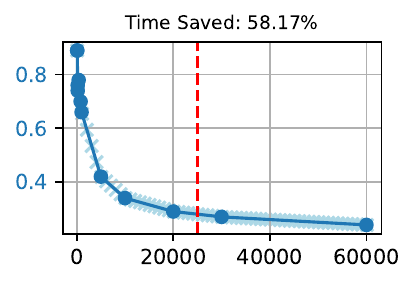}
        \raisebox{0.1in}{\small{Iterations}}
        \caption{ Scene-1}
    \end{subfigure}
    \hfill
    \begin{subfigure}{0.19\linewidth}
        \centering
        \includegraphics[width=\linewidth]{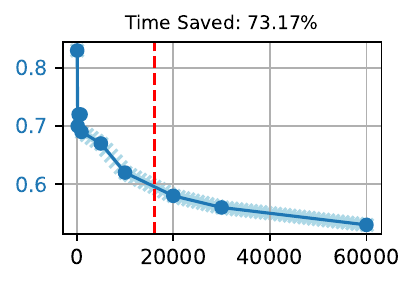}
        \raisebox{0.1in}{\small{Iterations}}
        \caption{ Scene-2}
    \end{subfigure}
    \hfill
    \begin{subfigure}{0.19\linewidth}
        \centering
        \includegraphics[width=\linewidth]{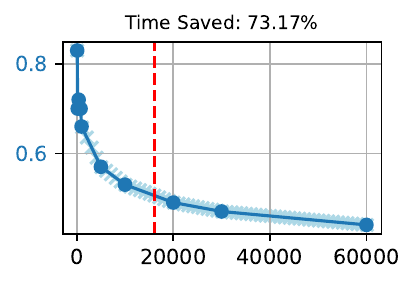}
        \raisebox{0.1in}{\small{Iterations}}
        \caption{Scene-3}
    \end{subfigure}
    \hfill
    \begin{subfigure}{0.19\linewidth}
        \centering
        \includegraphics[width=\linewidth]{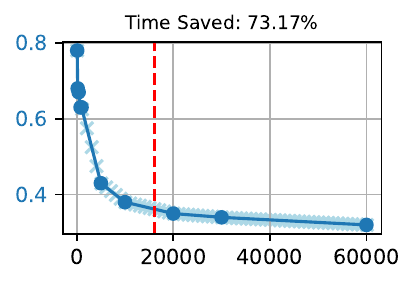}
        \raisebox{0.1in}{\small{Iterations}}
        \caption{Scene-4}
    \end{subfigure}
    \hfill
    \begin{subfigure}{0.19\linewidth}
        \centering
        \includegraphics[width=\linewidth]{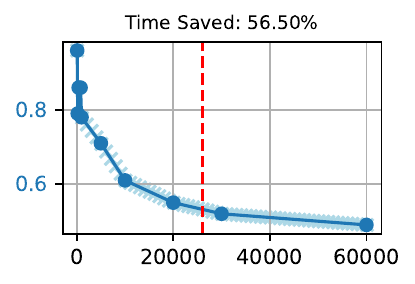}
        \raisebox{0.1in}{\small{Iterations}}
        \caption{Scene-5}
    \end{subfigure}\\
    \raisebox{0ex}{\textcolor{red}{\rule[0.5ex]{0.1cm}{1pt}\hspace{0.1cm}\rule[0.5ex]{0.1cm}{1pt}\hspace{0.1cm}\rule[0.5ex]{0.1cm}{1pt}}}~Detected Plateau Point \quad
    \raisebox{0.5ex}{\textcolor{blue}{\rule{0.5cm}{1pt}}} LPIPS \quad
    \textcolor{black}{\textbullet} Original \quad
    \textcolor{black}{$\times$} Interpolated \quad
    \caption{LPIPS on validation scenes and proposed early stoping of training.}
    \label{fig:lpips_plots}
\end{figure*}

Detailed look of LPIPS metric plots across various validation scenes alongside the algorithmically proposed early stopping points, which does not significantly compromise the reconstruction quality, is also provided. For a deeper look of LPIPS, F1 Score and the recommended stopping point for each case, consult \figref{fig:lpips_vs_f1} and \figref{fig:lpips_plots}. The different number of images used for the different scenarios is provided in \tabref{tab:data_distribution}.

\begin{table*}[!t]
  \centering
  \caption{Overview of Data Distribution Across Different Scenarios.}
  \begin{tabular}{lcrr}
    \toprule
    Category              & Scenario & Training Images & Validation Images \\
    \midrule
    \multirow{3}{*}{Main Scenario} & I      & 45           & 5                 \\
                                   & II     & 63           & 7                 \\
                                   & III    & 63           & 7                 \\
    \midrule
    \multirow{5}{*}{LPIPS Validation Scenario} & I      & 77           & 8                 \\
                                                & II     & 147          & 16                \\
                                                & III    & 103          & 11                \\
                                                & IV     & 115          & 12                \\
                                                & V      & 203          & 22                \\
    \bottomrule
  \end{tabular}
  \label{tab:data_distribution}
\end{table*}

\clearpage
\pagebreak

\subsection{Accurate Scaling of NeRF Reconstructions}
\begin{figure*}[!h]
    \centering
    \begin{subfigure}{0.45\linewidth}
        \centering
        \includegraphics[width=\linewidth]{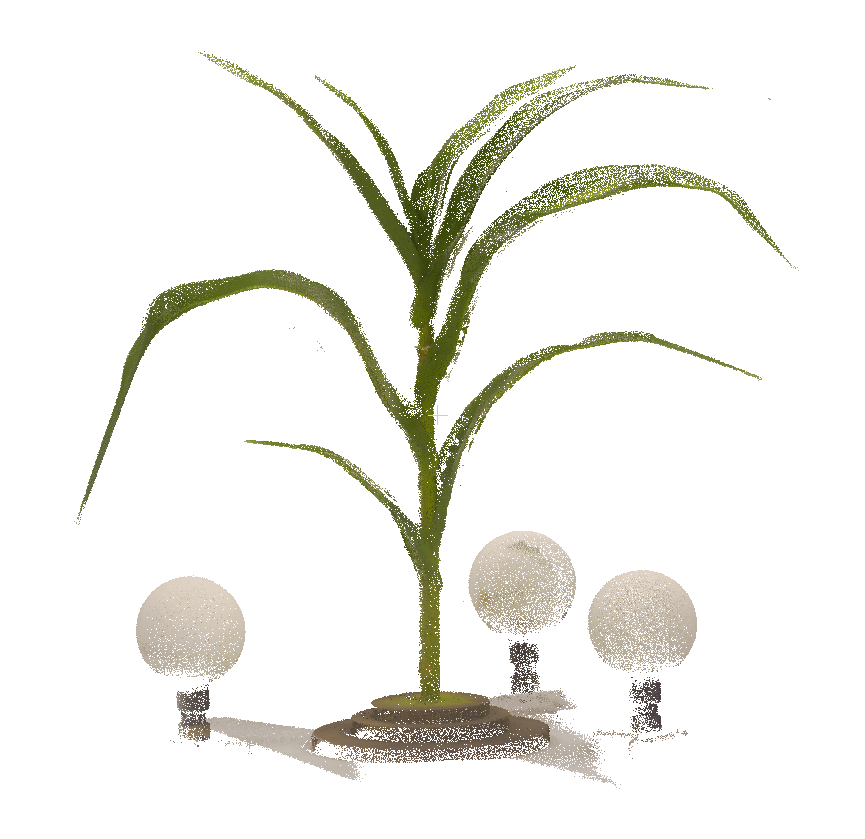}
        \caption{Original TLS}
    \end{subfigure}
    \hfill
    \begin{subfigure}{0.45\linewidth}
        \centering
        \includegraphics[width=\linewidth]{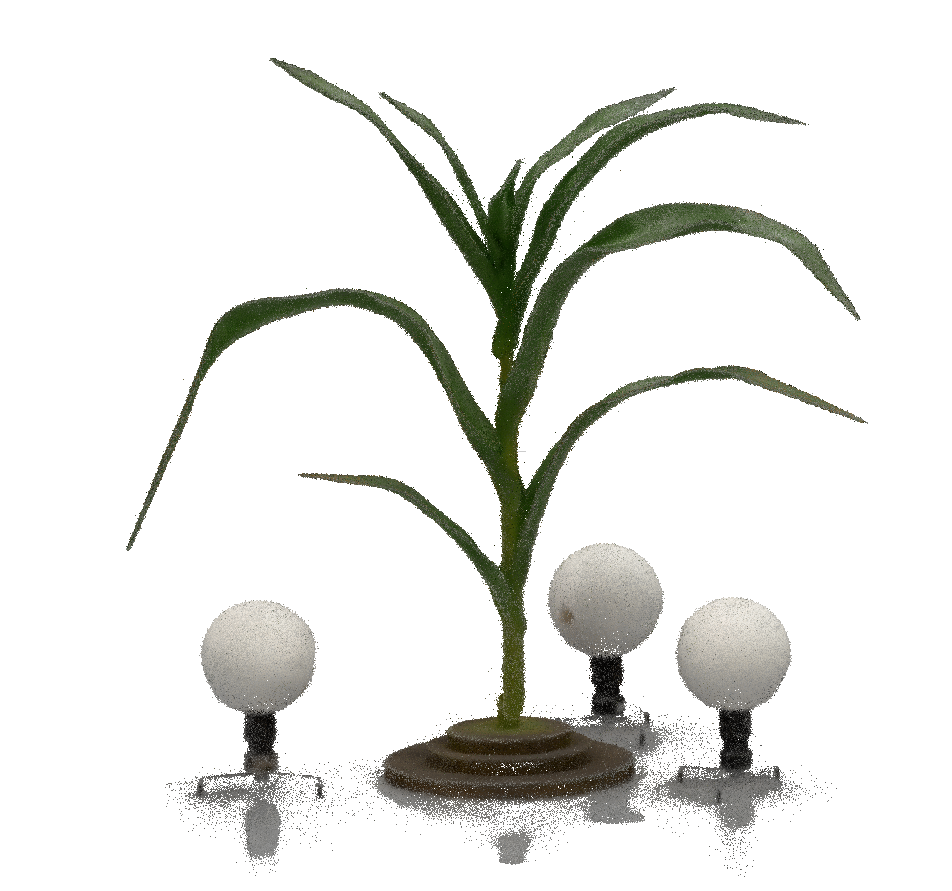}
        \caption{Scaling using NeRF.}
    \end{subfigure}
    \caption{Scaling comparison of 3D scene reconstruction. (a) Scene captured using Terrestrial Laser Scanning (TLS). (b) Scene reconstructed using Neural Radiance Fields (NeRF) and scaled using the calibration spheres. The plant height from the scaled NeRF aligns well with manual measurements, demonstrating NeRF's accuracy in scene reconstruction and scaling using reference objects. The plant height is measured from the stalk base to the highest point.}
    \label{fig:scaling}
\end{figure*}

The effectiveness of using known reference objects for scaling NeRF reconstructions is demonstrated in \figref{fig:scaling}. We placed three calibration spheres of known dimensions alongside a plant specimen. The scene was captured using both TLS and reconstructed using NeRF. Sphere measurements from TLS served as the ground truth, with radii of approximately 69.0, 68.6, and 69.2 mm. Using the average sphere size from TLS as a reference, we scaled the NeRF reconstruction. The NeRF-derived sphere measurements were 68.6, 68.8, and 69.2 mm, showing close agreement with the TLS data. We measured the plant height to validate the scaling accuracy. The NeRF-based measurement yielded a height of 772 mm, which aligns well with manual measurements of 770 ± 5 mm (averaged over three repetitions). This example demonstrates that NeRF reconstruction, when appropriately scaled using known reference objects, can achieve dimensional accuracy within 1\% of the physical measurements.

